\documentclass{article}

\usepackage{microtype}
\usepackage{booktabs} 

\usepackage[accepted]{icml2025}

\usepackage{amsmath}
\usepackage{amssymb}
\usepackage{mathtools}
\usepackage{amsthm}

\usepackage{tabularx}
\usepackage{makecell}
\usepackage{pifont}
\usepackage{xcolor}
\usepackage{rotating}

\usepackage{algorithm}
\usepackage{algorithmic}
\usepackage{amsmath}
\DeclareMathOperator*{\argmax}{arg\,max}

\usepackage{graphicx, subfig}
\usepackage{caption}
\usepackage{multicol}

\usepackage{amssymb}
\usepackage{setspace}
\usepackage{hyperref}

\usepackage[skins,breakable]{tcolorbox}
\tcbuselibrary{listings}

\usepackage{listings}
\usepackage{tcolorbox}
\usepackage{mdframed}
\usepackage{marvosym}

\usepackage{xcolor}
\usepackage{colortbl}

\usepackage{enumitem}
\setenumerate[1]{itemsep=0pt,partopsep=0pt,parsep=\parskip,topsep=5pt}
\setitemize[1]{itemsep=0pt,partopsep=0pt,parsep=\parskip,topsep=5pt}
\setdescription{itemsep=0pt,partopsep=0pt,parsep=\parskip,topsep=5pt}

\usepackage[capitalize,noabbrev]{cleveref}

\theoremstyle{plain}

\theoremstyle{definition}

\theoremstyle{remark}

\usepackage[textsize=tiny]{todonotes}

\usepackage{xurl}

\icmltitlerunning{\textbf{A Step-Wise, Multi-Dimensional, and Generalist Reward Model with Benchmark for Virtual Agents}}

\begin{document}

\twocolumn[
\icmltitle{Boosting Virtual Agent Learning and Reasoning: A Step-Wise, Multi-Dimensional, and Generalist Reward Model with Benchmark}



\icmlsetsymbol{equal}{*}
\icmlsetsymbol{corresponding}{\Letter}

\begin{icmlauthorlist}
\icmlauthor{Bingchen Miao}{zju,ant,equal}
\icmlauthor{Yang Wu}{ant,equal}
\icmlauthor{Minghe Gao}{zju,equal}
\icmlauthor{Qifan Yu}{zju}
\icmlauthor{Wendong Bu}{zju,ant}
\icmlauthor{Wenqiao Zhang}{zju}
\icmlauthor{Yunfei Li}{ant}
\icmlauthor{Siliang Tang}{zju}
\icmlauthor{Tat-Seng Chua}{nus}
\icmlauthor{Juncheng Li}{zju,corresponding}
\end{icmlauthorlist}

\icmlaffiliation{zju}{Zhejiang University, Hangzhou, China}
\icmlaffiliation{ant}{Ant Group, Hangzhou, China}
\icmlaffiliation{nus}{National University of Singapore, Kent Ridge, Singapore}

\icmlcorrespondingauthor{Juncheng Li}{junchengli@zju.edu.cn}

\icmlkeywords{Machine Learning, ICML}

\vskip 0.3in
]



\printAffiliationsAndNotice{\icmlEqualContribution} 

\begin{abstract}
\label{sec:abstract} 

\vspace{-2pt}
The development of Generalist Virtual Agents (GVAs) has shown significant promise in autonomous task execution. However, current training paradigms face critical limitations, including reliance on outcome supervision and labor-intensive human annotations. To address these challenges, we propose \textbf{\texttt{Similar}}, a \underline{\textbf{s}}tep-w\underline{\textbf{i}}se \underline{\textbf{m}}ult\underline{\textbf{i}}-dimensiona\underline{\textbf{l}} gener\underline{\textbf{a}}list \underline{\textbf{r}}eward model, which offers fine-grained signals for agent training and can choose better actions for inference-time scaling. Specifically, we begin by systematically defining five dimensions for evaluating agent actions. Building on this framework, we design an MCTS-P algorithm to automatically collect and annotate step-wise, five-dimensional agent execution data. Using this data, we train \textbf{\texttt{Similar}} with our crafted Triple-M strategy. Furthermore, we introduce the first benchmark in the virtual agent domain for step-wise, multi-dimensional reward model training and evaluation, named \textbf{\textit{SRM}}. This benchmark consists of two components: \textbf{\textit{SRMTrain}}, which serves as the training set for \textbf{\texttt{Similar}}, and \textbf{\textit{SRMEval}}, a manually selected test set for evaluating the reward model. Experimental results demonstrate that \textbf{\texttt{Similar}}, through its step-wise, multi-dimensional assessment and synergistic gain, provides GVAs with effective intermediate signals during both training and inference-time scaling. The code is available in \href{https://github.com/antgroup/Similar}{https://github.com/antgroup/Similar}.

\end{abstract}
\vspace{-25pt}

\section{Introduction}
\label{sec:introduction}

\vspace{-5pt}

Generalist Virtual Agents (GVAs~\cite{gao2024generalistvirtualagentssurvey, bu2025limitsvirtualagentapplication}) powered by Multimodal Large Language Models (MLLMs~\cite{li2023fine, 10121664, pan2025generative, pan2024auto}) process multimodal inputs (UI elements~\cite{zhang2024ufouifocusedagentwindows}, text~\cite{DBLP:conf/nips/0001ST00Z23}, visuals~\cite{yan2023gpt}) to navigate digital environments, performing tasks and generating outputs that manipulate interfaces or provide responses. The training of GVAs relies on outcome-based rewards from human-annotated trajectories, where task completion serves as the primary supervision signal~\cite{he2024webvoyager}.

\begin{figure*}
    \begin{minipage}[b]{0.5\linewidth}
    \centering
    \includegraphics[width=0.9\textwidth]{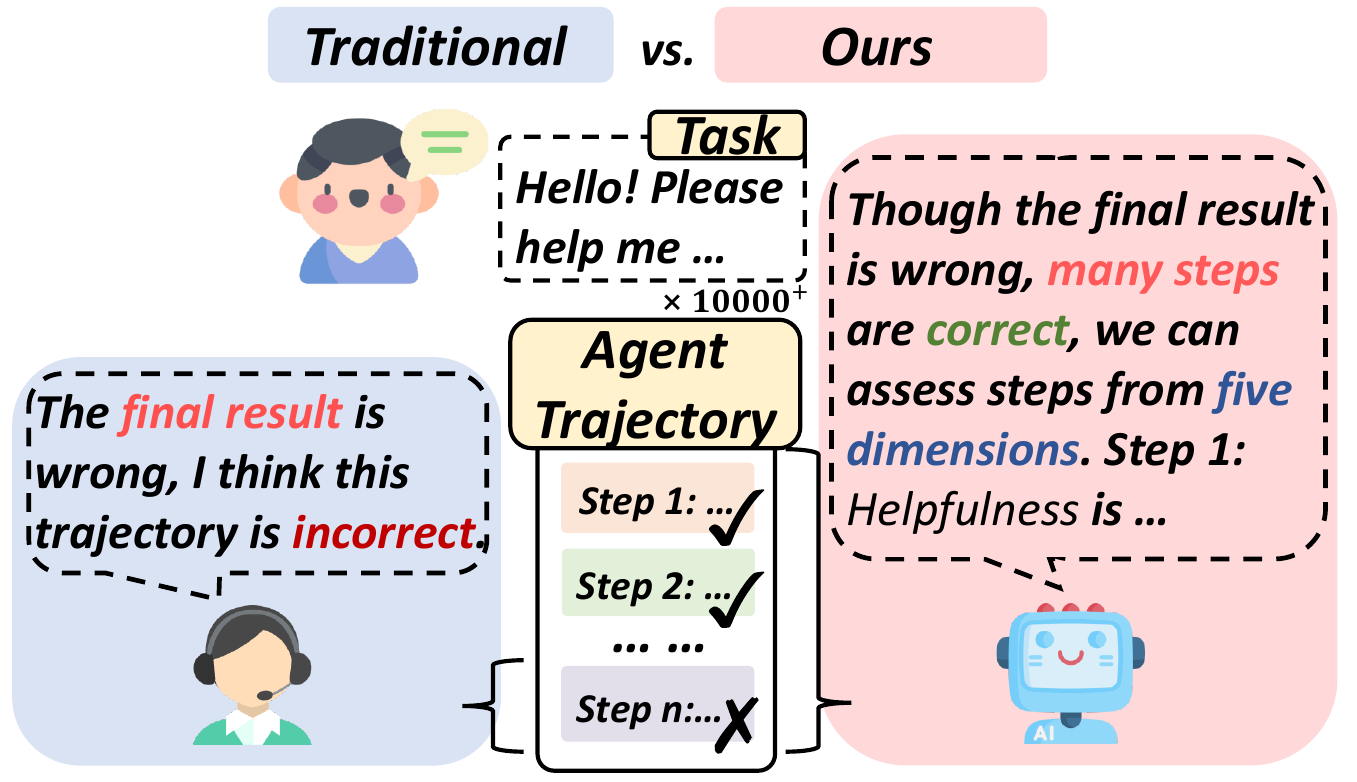}
    \vspace{-5pt}
    \caption*{(a)}
    \end{minipage}
    \begin{minipage}[b]{.5\linewidth}
    \centering
        \vskip 0.15in
        \begin{center}
        \scriptsize
        \begin{sc}
        \begin{tabular}{lccccc}
        \toprule
        \textbf{Work} & \textbf{\makecell{PB}} & \textbf{\makecell{AA}} & \textbf{\makecell{MD}} & \textbf{\makecell{Platforms}} & \textbf{Size} \\
        \midrule
        WebArena & \textcolor{red}{\ding{55}} & \textcolor{red}{\ding{55}} & \textcolor{red}{\ding{55}} & Web & 812 \\
        VisualWebArena & \textcolor{red}{\ding{55}} & \textcolor{red}{\ding{55}} & \textcolor{red}{\ding{55}} & Web & 910 \\
        WorkArena & \textcolor{red}{\ding{55}} & \textcolor{red}{\ding{55}} & \textcolor{red}{\ding{55}} & \makecell{Web, Windows} & 19.9k \\
        Android World & \textcolor{red}{\ding{55}} & \textcolor{red}{\ding{55}} & \textcolor{red}{\ding{55}} & \makecell{Android} & 116 \\
        WebVoyager & \textcolor{red}{\ding{55}} & \textcolor{green}{\ding{51}} & \textcolor{red}{\ding{55}} & \makecell{Web} & 300 \\
        OSWorld & \textcolor{red}{\ding{55}} & \textcolor{green}{\ding{51}} & \textcolor{red}{\ding{55}} & \makecell{Linux, Windows} & 369 \\
        GUI Odyssey & \textcolor{red}{\ding{55}} & \textcolor{green}{\ding{51}} & \textcolor{red}{\ding{55}} & Android & 7.7K \\
        Mobile-Agent & \textcolor{green}{\ding{51}} & \textcolor{red}{\ding{55}} & \textcolor{red}{\ding{55}} & Android & 30 \\
        OmniACT & \textcolor{green}{\ding{51}} & \textcolor{red}{\ding{55}} & \textcolor{red}{\ding{55}} & \makecell{Web, Windows} & 9.8K \\
        \midrule
        \textbf{\textit{SRM}} (Ours) & \textcolor{green}{\ding{51}} & \textcolor{green}{\ding{51}} & \textcolor{green}{\ding{51}} & \textbf{\makecell{Web, Linux,\\Windows, Android}} & \textbf{110k} \\
        \bottomrule
        \end{tabular}
        \end{sc}
        \end{center}
        \vskip -0.1in
        \vspace{-2pt}
        \caption*{(b)}
    \end{minipage}
    \vspace{-20pt}
    \caption{(a) Traditional coarse-grained outcome-based labor-intensive paradigm vs. Our fine-grained process-based autonomous paradigm. (b) Comparison of benchmark platforms. Previous works focused on virtual agent benchmarks, while ours is the first benchmark specifically for virtual agent reward models. The PB, AA, and MD in the list represent Process-based, Automatic Annotation, and Multi-Dimension, respectively.}
    \label{fig:1}
    \vspace{-15pt}
\end{figure*}

\vspace{-3pt}

However, this paradigm with the outcome reward for GVAs has significant limitations. \textbf{1) Lack of multi-dimensional fine-grained process supervision:} Existing methods typically focus on global task success or the final state of the task, overlooking intermediate steps in execution~\cite{yu-etal-2024-ovm}. This oversight makes it impossible to pinpoint failures in unsuccessful trajectories or identify errors in successful ones, resulting in inefficient learning and reasoning processes~\cite{uesato2022solvingmathwordproblems, lightman2023lets, gao2025benchmarkingmultimodalcotreward}. In contrast, a Process Reward Model (PRM) offers a better alternative by providing fine-grained supervision signals to guide agent behavior. \textbf{2) Reliance on human-annotated trajectories with reward signals:} Domain experts need to meticulously annotate trajectories consisting of dozens of steps with accurate outcome-based rewards to train GVAs~\cite{he2024webvoyager}. Furthermore, obtaining step-wise fine-grained process-based rewards makes the process labor-intensive, time-consuming, and nearly infeasible at scale~\cite{NEURIPS2023_5950bf29, burns2022motifvln}. \textbf{3) Difficulty in scaling inference-time}. Recent outstanding work has demonstrated that inference-time scaling can significantly enhance agent performance~\cite{deepseekai2025deepseekr1incentivizingreasoningcapability, wu2024comparative}. However, relying on result-based training with extensive human annotation limits the ability to handle complex tasks by selecting the best action or step-by-step assessments.~\cite{snell2024scalingllmtesttimecompute, NEURIPS2022_639a9a17}. Therefore, our focus has shifted to breaking result-oriented manual annotation-dependent training methods through step-wise automatic reward model.

\vspace{-3pt}

To address these challenges, we propose \textbf{\texttt{Similar}}, a \underline{\textbf{s}}tep-w\underline{\textbf{i}}se \underline{\textbf{m}}ult\underline{\textbf{i}}-dimensiona\underline{\textbf{l}} gener\underline{\textbf{a}}list \underline{\textbf{r}}eward model. It provides fine-grained supervision signals for agent training and inference-time scaling, enabling automated, multi-faceted assessment without relying on labor-intensive human annotations. Specifically, \textbf{1)} we introduce a step-wise, multi-dimensional assessment system for GVA actions, \textbf{defining five key dimensions of process supervision signals}: \textit{Helpfulness}, \textit{Odds of Success}, \textit{Efficiency}, \textit{Task Relevance}, and \textit{Coherence}. These dimensions are designed to minimize overlap while collectively providing a comprehensive assessment of each action's quality. \textbf{2)} Then we design an \textbf{MCTS-P} algorithm to \textbf{automatically collect and annotate tens of thousands of step-wise actions} based on the five dimensions. This approach is applied across four distinct environment domains: Web, Android, Linux, and Windows. Unlike existing methods that rely on labor-intensive human annotations, this automated framework ensures scalability across diverse environments and generates a unified, fine-grained dataset that captures universal reasoning patterns, significantly reducing the cost and time required for data collection. \textbf{3)} Finally, using this dataset, we employ a \textbf{Triple-M} (multi-step, multi-objective, and multi-modal) \textbf{strategy to train a reward model}. This strategy integrates multiple dimensions of assessment and generates a synergistic gain by combining the strengths of five dimensions. As illustrated in Figure~\ref{fig:1} (a), traditional methods focus solely on outcomes, require significant manual effort, and are coarse-grained, outcome-based, and labor-intensive. In contrast, our approach enables \textbf{\texttt{Similar}} to perform step-wise, multi-dimensional automatic assessment of agent trajectories, making it fine-grained, process-based, and autonomous. Building on this, \textbf{\texttt{Similar}} provides fine-grained rewards for GVA during the training phase, while seamlessly guiding GVA and robustly optimizing performance by scaling inference-time during the inference phase.

\vspace{-2pt}

Since reward models are crucial for GVAs, and prior research has not focused on evaluating reward models, we propose \textbf{\textit{SRM}}, the first benchmark in the GVA domain for step-wise, multi-dimensional reward model training and evaluation. Figure~\ref{fig:1} (b) illustrates that it consists of~$110$k automatically annotated data points, divided into the scalable \textbf{\textit{SRMTrain}} ($78$k) for training \textbf{\texttt{Similar}} and the curated \textbf{\textit{SRMEval}} ($32$k) for evaluating reward models.

\vspace{-2pt}

Our reward model, \textbf{\texttt{Similar}}, can enhance the learning and reasoning of GVAs. \textbf{For training}, it serves as a reward model in a reinforcement learning framework, guiding GVAs to optimize its behavior based on action quality. By providing fine-grained feedback, it effectively guides the agents' learning process and enhances their performance. \textbf{For inference-time scaling}, it can be integrated with search algorithms such as Monte Carlo Tree Search (MCTS) to leverage reward signals for filtering candidate actions, and improve model performance~\cite{deepseekai2025deepseekr1incentivizingreasoningcapability, zang2025internlmxcomposer25rewardsimpleeffectivemultimodal}. By selecting actions that are more likely to complete the task, it enhances accuracy and reduces time.

\vspace{-2pt}

Extensive experiments demonstrate the superiority of our approach: \textbf{1)} Effectiveness of step-wise, multi-dimensional data: Using our collected data for reward modeling, \textbf{\texttt{Similar}}-RL-Llama achieves a $13.2\%$ improvement over the baseline Llama-3.2-11B-Vision model on the \textbf{\textit{SRMEval}} benchmark, demonstrating the effectiveness of our automated framework in enabling fine-grained assessment of GVA actions. \textbf{2)} Synergistic gain from the Triple-M strategy: The Triple-M strategy integrates multiple dimensions by leveraging the strengths of five dimensions, enabling \textbf{\texttt{Similar}}-TM-Llama to achieve an Avg score of $61.2$ on \textbf{\textit{SRMEval}}, significantly outperforming \textbf{\texttt{Similar}}-RL-Llama ($53.9$, a $13.5\%$ improvement). This highlights the synergistic gain of our training strategy. \textbf{3)} Effective guidance in training and inference: \textbf{\texttt{Similar}} provides fine-grained, multi-dimensional feedback during training and integrates with search algorithms like MCTS to scale inference-time during inference to improve reasoning accuracy. Its strong performance across multiple benchmarks underscores its versatility and practical applicability.

\vspace{-2pt}

Our contributions can be summarized as follows:

\begin{itemize}
    \vspace{-8pt}
    \item We define five dimensions for step-wise GVA assessment and an MCTS-P algorithm to collect fine-grained, cross-platform reward model data annotations.
    \vspace{-3pt}
    \item Based on these data, we propose a Triple-M strategy to train a reward model, called \textbf{\texttt{Similar}}, integrating multiple dimensions and generating synergistic gains for robust, fine-grained feedback.
    \vspace{-3pt}
    \item Moreover, we introduce \textbf{\textit{SRMEval}}, a multi-step, multi-dimensional, and multi-platform benchmark for evaluating reward models, which is a set of \textbf{\textit{SRM}} to advance research in reward model performance assessment.
    \vspace{-3pt}
    \item Experiments demonstrate that our approach, through step-wise multi-dimensional assessment, provides GVAs with superior intermediate signals during both training and inference-time scaling.
    \vspace{-3pt}
\end{itemize}

\vspace{-10pt}
\section{Related Work}
\label{sec:related_work}



\vspace{-3pt}

\subsection{Fine-Tuning Virtual Agent}

Fine-tuning Virtual Agents traditionally relies on human-annotated datasets, which are labor-intensive and time-consuming~\cite{wang2023survey}. Methods such as imitation learning~\cite{pmlr-v162-humphreys22a} and reinforcement learning~\cite{branavan-etal-2009-reinforcement, branavan-etal-2010-reading} have been employed to fine-tune agents based on curated expert trajectories or outcome rewards, but these approaches often suffer from compounding errors and limited exploration~\cite{christianos2023panguagentfinetunablegeneralistagent, xi2024agentgym, song-etal-2024-trial}. Recent advancements, such as reject sampling fine-tuning (RFT)~\cite{yuan2023scaling} and direct policy optimization (DPO)~\cite{NEURIPS2023_a85b405e}, have sought to reduce reliance on human annotations by leveraging both successful and failure trajectories~\cite{lai2024stepdpo, zhang2024chain}. However, these methods face significant challenges, including the lack of process supervision and reliance on human-annotated data, which limit their scalability and adaptability~\cite{xu-etal-2021-grounding, he2024webvoyager, 10.5555/3666122.3668731}. In contrast, our work addresses these limitations by introducing a novel training paradigm that leverages multi-dimensional process supervision and automated annotation to enhance the learning and reasoning capabilities of GVAs.

\vspace{-5pt}

\subsection{Reward Models for Virtual Agent}

Reward Models (RMs) are critical for guiding virtual agents by evaluating action quality~\cite{zhai2024enhancingdecisionmakingllmagents, zhou2024trad}. While Outcome Reward Models (ORMs) focus on task success~\cite{yu-etal-2024-ovm, yu2023metamath}, Process Reward Models (PRMs) provide feedback on intermediate steps, offering an evaluation of agent performance in complex reasoning tasks~\cite{uesato2022solvingmathwordproblems}. Recent studies show that PRMs outperform ORMs in tasks like math reasoning, where process supervision is essential~\cite{lightman2023lets, li-etal-2023-making}. However, generating high-quality process supervision data remains challenging, as human annotation is expensive. To address this, methods like step-level Q-value~\cite{zhai2024enhancingdecisionmakingllmagents} and ReST-MCTS$^*$~\cite{zhang2024rest} have explored MCTS to automate data collection, achieving significant gains. Building on these insights, our work introduces a step-wise, multi-dimensional system leveraging MCTS to collect fine-grained annotations, enabling a robust reward model to guide GVAs.

\vspace{-10pt}
\section{Method}
\label{sec:method}

\begin{figure*}[!t]
\includegraphics[width=1\textwidth]{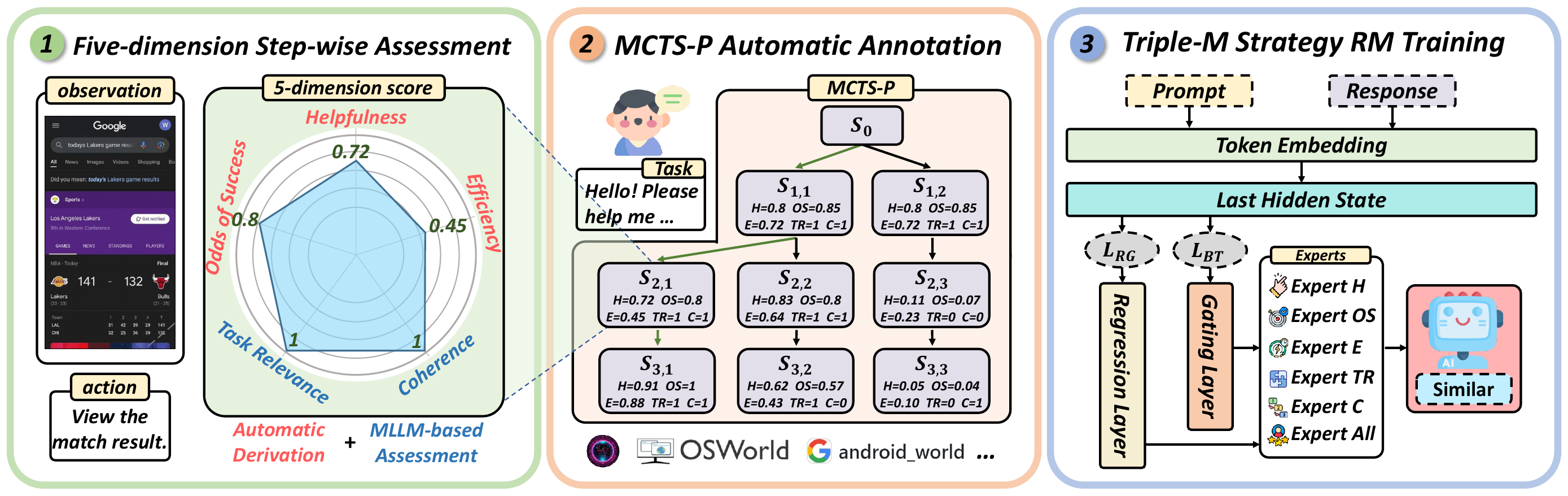}
\centering\caption{\textbf{\texttt{Similar} model training pipeline.} First, we systematically define five dimensions to describe the quality of an agent's step. Next, we propose an MCTS-P algorithm to automatically collect annotated step-wise data. Finally, we design the Triple-M strategy to train the \textbf{\texttt{Similar}} model, which can guide the agent during both the training and inference phases.}
\label{fig:model_structure}
\vspace{-15pt}
\end{figure*}

\vspace{-5pt}

In this section, we present the pipeline for training our proposed \textbf{\texttt{Similar}} model. The \textbf{\textit{SRM}} benchmark will be introduced in Section~\ref{sec:SRM_benchmark}. As shown in Figure~\ref{fig:model_structure}, to evaluate agent steps multi-dimensionally, we first define five-dimension process supervision (Section~\ref{3.1}). Next, we introduce an MCTS-P algorithm to automatically collect step-wise annotations (Section~\ref{3.2}). Finally, we design the Triple-M strategy to train \textbf{\texttt{Similar}}, achieving synergistic gains across five dimensions (Section~\ref{3.3}).

\vspace{-10pt}

\subsection{Five-Dimensional Process Supervision Framework}
\label{3.1}

\vspace{-5pt}

To assess the quality of an agent's steps, we systematically define a five-dimensional process supervision framework. Given task complexity and interdependencies, a single metric is insufficient for assessing step quality~\cite{zhai2024enhancingdecisionmakingllmagents}. Our framework addresses this limitation by covering the multi-faceted nature of step assessment. The first three dimensions—\textit{Helpfulness}, \textit{Odds of Success}, and \textit{Efficiency}—are computed automatically, while the remaining two—\textit{Task Relevance} and \textit{Coherence}—are assessed using MLLMs. These dimensions are independent and interpretable, ensuring broad applicability across tasks.

\vspace{-3pt}

The current step is denoted as $S_i$, where $i$ is the step index. The three automatic metrics are derived through MCTS simulations~\cite{luo2024improvemathematicalreasoninglanguage}. For $S_i$, we simulate $N$ subsequent trajectories until a termination condition is met (i.e., the agent completes the task or reaches the maximum step length). We define the basic reward $r_{i}$ as:
\(
r_{i}=\begin{cases} 
1, & \exists a_{i,j} \in A, \, a_{i,j} = a^* \\ 
0, & \text{otherwise} 
\end{cases},\, j \in N,
\)
where $a^*$ represents the ground truth, $a_{i,j}$ denotes the final action of the $j$-th trajectory in step $i$, and $A$ is the set of all actions. The following sections detail how each dimension assesses $S_i$.

\textbf{\textit{Helpfulness (H)}.} It quantifies whether a given step contributes positively or negatively to task completion, assigning values inversely proportional to the trajectory length. This dimension is designed to assess the impact of each step on the overall task trajectory. Steps that facilitate task completion are considered helpful, while those that hinder progress are assigned negative values. For example, each step in a $3$-step successful trajectory is worth $1/3$, while steps hindering progress (those failing to lead to success) receive corresponding negative values. And in two successful trajectories of the same task, the steps in the trajectory with fewer steps will have higher \textit{Helpfulness} value.


The \textit{Helpfulness} can be calculated as the following formula:
\[ H_{i} = \frac{1 - AC_{i-1}}{M - i + 1} (2r_{i} - 1) ,\]
where \( AC_{i} = \begin{cases} 0, & i = 0 \\ \max(AC_{i-1} + H_{i}, 0), & \text{otherwise} \end{cases} \), which is a mathematical placeholder to recursively track cumulative \textit{Helpfulness} scores during MCTS rollouts. And \( M \) is the total number of reasoning steps.

\begin{figure}[t]
\vspace{0.2cm}
\includegraphics[width=0.35\textwidth]{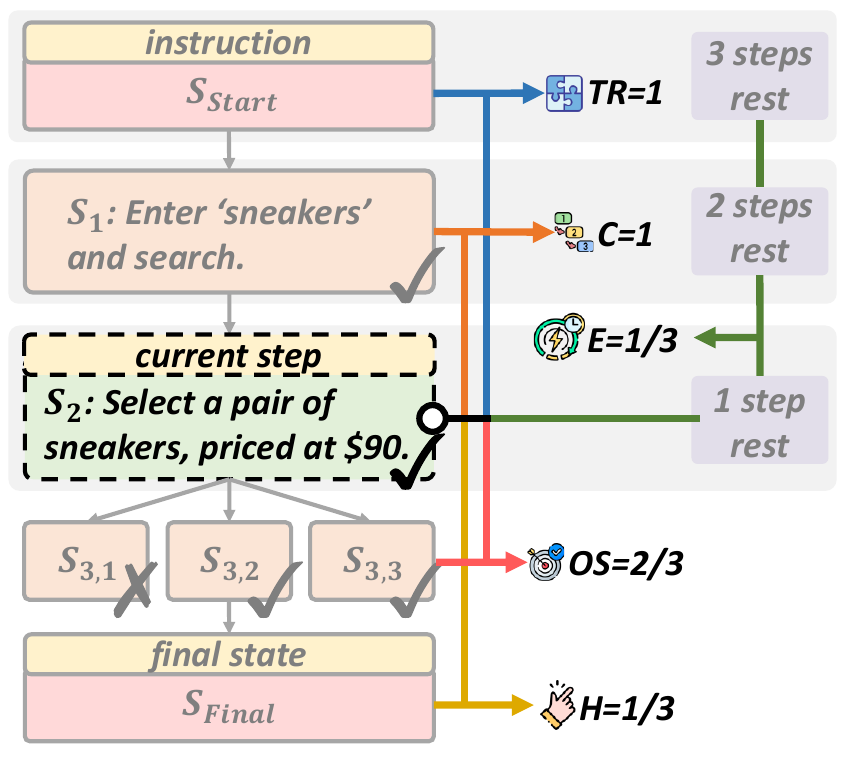}
\centering\caption{\textbf{An example describing the $5$ dimensions}. A single step by an agent relates to $5$ task elements: instruction, last step, next step, final state, and number of steps. Our $5$ dimensions align with these: \textbf{\textit{H}} (final state), \textbf{\textit{OS}} (next step), \textbf{\textit{E}} (number of steps), \textbf{\textit{TR}} (instruction), and \textbf{\textit{C}} (last step).}  
\label{fig:dimension_relation}
\vspace{-15pt}
\end{figure}

\textbf{\textit{Odds of Success (OS)}.} It measures the probability that a given step will lead to the successful completion of the task. This dimension is crucial for identifying steps that are more likely to result in a successful outcome. Steps with higher values are more likely to lead to success, while those with lower values are less likely to succeed. Conversely, incorrect steps lead to failure. The \textit{Odds of Success} is calculated by evaluating the proportion of successful paths among simulated results from a given step.

The formula for \textit{Odds of Success} is defined as:
\[ OS_{i} = \frac{\sum^N_{j=1} \mathbb{I}(a_{i,j} = a^*)}{N} ,\]
where \( \mathbb{I}(\cdot) \) is the indicator function.

\textbf{\textit{Efficiency (E)}.} It evaluates whether a given step is operationally efficient in terms of resource consumption, such as time or computational effort. A fundamental assumption is that fewer steps equate to higher efficiency, as shorter paths imply lower resource usage. Steps that reduce the total number of steps required to complete a task are considered efficient, as they enable the agent to accomplish the task more quickly and with fewer resources.

The \textit{Efficiency} metric is calculated as the following formula:
\[ E_i = \frac{Len_{i-1} - Len_i}{len_0} ,\]
where \( Len_i = \text{avg}(Len(S_{i,j})) \), and \( Len(S_{i,j}) \) represents the number of steps remaining to complete the task after executing action \( a_{i,j} \).


\textbf{\textit{Task Relevance (TR)}.} It assesses whether a step is related to the task instruction. Some steps may be task-relevant but still fail (e.g., recording incorrect notes), while others may be irrelevant yet contribute to success (e.g., clicking on a blank screen). These distinctions cannot be captured through automated calculations. However, MLLMs with advanced image understanding can evaluate this dimension. \textit{Task Relevance} is binary, with values $\{0, 1\}$.

\textbf{\textit{Coherence (C)}.} It: measures the continuity and logical flow between consecutive steps. Some operations, although task-relevant, efficient, and likely to lead to success, may lack coherence with the previous step. For example, in a task such as ``Query the Lakers' game result and record it in a Note," opening a browser and a Note simultaneously may lack coherence compared to directly searching for the game result after opening the browser. \textit{Coherence} is also evaluated using MLLMs and is a binary classification dimension, with possible values of $\{0, 1\}$.

The prompts of two MLLM-based assessment dimensions is detailed in Appendix~\ref{sec:prompt}. To better understand the five dimensions, we use Figure~\ref{fig:dimension_relation}. Each agent step relates to five task elements: instruction, last step, next step, final state, and step count. In the figure, step $S_2$ is assessed by these dimensions. The task requires $3$ steps: $S_1$, $S_2$, and $S_{3,2}$, with all sharing an \textit{H} value of $\frac{1}{3}$. Among the next steps from $S_2$, $S_{3,2}$ and $S_{3,3}$ are correct, yielding an \textit{OS} value of $\frac{2}{3}$. The \textit{E} value is $\frac{1}{3}$, calculated as $\frac{2-1}{3}$. Since $S_2$ aligns with $S_1$ and the instruction, \textit{TR} and \textit{C} values are $1$.

\vspace{-8pt}

\subsection{Automatic Generalist Dataset Collecting}
\label{3.2}

\vspace{-4pt}

MCTS (Monte Carlo Tree Search) is a heuristic search algorithm used in decision-making, combining random sampling and tree-based search to find the optimal option. Based on its advantages, such as its scalability and efficient exploration-exploitation balance~\cite{luo2024improvemathematicalreasoninglanguage, wang-etal-2024-multi-step}, we propose a modified version, MCTS-P, to automatically collect annotated data. MCTS-P leverages the five dimensions introduced in Section~\ref{3.1} to comprehensively assess each step taken by the virtual agent.

In MCTS-P, the five-dimensional scores are used as the basis for node selection and backpropagation. Specifically, the algorithm computes a weighted sum of the five dimensions to obtain a composite score for each step. This composite score serves as the value \( v_{i,j} \) for each node \( S_{i,j} \) in the search tree. The tree structure in MCTS-P is similar to traditional MCTS, with each node \( S_{i,j} \) storing the action \( a_{i,j} \), visit count \( n_{i,j} \), and value \( v_{i,j} \). The detailed pseudo-code for MCTS-P is provided in Algorithm~\ref{alg:mcts-p}.

To build a comprehensive and generalist dataset for training and testing reward models, we collect a large number of task trajectories from agents across four different platforms: Web, Android, Linux, and Windows. Using the MCTS-P algorithm, we perform automatic data annotation to collect process supervision signals. The annotation process involves the following steps: \textbf{1)} For each node \( S_{i,j} \) in the search tree \( T_q \), we calculate the minimum number of steps \( M \) required to reach a correct answer. \textbf{2)} During the expansion phase, the algorithm simulates \( N \) possible outcomes for each step to obtain the basic reward \( r_i \). \textbf{3)} Based on \( M \) and \( r_i \), we compute the three automatically calculated dimensions: \textit{Helpfulness}, \textit{Odds of Success}, and \textit{Efficiency}. \textbf{4)} We then use a MLLM (GPT-4o~\cite{hurst2024gpt}) to evaluate the \textit{Task Relevance} and \textit{Coherence} of each step. \textbf{5)} We prune all incomplete branches (those that do not reach a final answer) and verify the correctness of the remaining paths using the evaluation methods provided by the four benchmark environments. The obtained trajectories are selected as the final dataset for training and evaluation.

\setlength{\textfloatsep}{5pt}
\begin{algorithm}[tb]
   \caption{MCTS-P Algorithm}
   \label{alg:mcts-p}
   \scriptsize
\textbf{Input:} Initial state $s_0$
\begin{algorithmic}[1]
   \STATE Create root node $S_0$ with state $s_0$
   \WHILE{within computational budget}
      \STATE $S_i \leftarrow \text{TreePolicy}(S_0)$
      \STATE $\Delta \leftarrow \text{DefaultPolicy}(s(S_i))$ \quad// Simulate a random playout to estimate
      \STATE \text{Backup}($S_i$, $\Delta$) \quad// Backpropagate the value to update parent nodes
   \ENDWHILE
   \STATE \textbf{return} $a(\text{BestChild}(S_0, 0))$

   \STATE
   \FUNCTION{TreePolicy($S$)}
      \WHILE{$S$ is nonterminal}
         \IF{$S$ not fully expanded}
            \STATE \textbf{return} \text{Expand}($S$) \quad// Expand the tree by adding a new child node
         \ELSE
            \STATE $S \leftarrow \text{BestChild}(S, C)$
         \ENDIF
      \ENDWHILE
      \STATE \textbf{return} $S$
   \ENDFUNCTION

   \STATE
   \FUNCTION{BestChild($S, c$)}
      \STATE $v(S) = H(S)+OS(S)+E(S)+TR(S)+C(S)$
      \STATE \textbf{return} $\argmax_{S' \in \text{children of } S} \left( \frac{v(S')}{n(S')} + c\sqrt{\frac{2 \ln n(S)}{n(S')}} \right)$
   \ENDFUNCTION
\end{algorithmic}
\textbf{Output:} Action $a$
\end{algorithm}

\vspace{-6pt}

\subsection{Triple-M Strategy for RM Training}
\label{3.3}

\vspace{-4pt}

Traditional reward modeling tasks typically rely on human-annotated data~\cite{ArmoRM}, whereas our approach generates step-wise annotations across multiple dimensions. To address the challenge of integrating \underline{M}ulti-step, \underline{M}ulti-dimensional, and \underline{M}ulti-modal data, we propose a novel \textbf{Triple-M strategy} tailored for virtual agents.

Our Triple-M strategy leverages a pre-trained decoder-only MLLM as the backbone feature extractor $f_\theta$, divided into two stages. The first stage trains a regression layer for five-dimensional score prediction. For each input sequence $x \oplus y$ (where $x$ represents the prompt and $y$ represents the response), we extract the last hidden state $h \in \mathbb{R}^d$ with $d$-dimensional features and map it to a five-dimensional reward score through a linear regression layer $W \in \mathbb{R}^{d \times 5}$. The model is optimized using a regression loss:
\[
L_{RG} = \min_{\theta, W} \mathbb{E}_{x, y, r \in \mathcal{D}} \| W^\top h - r \|_2^2,
\]
where $r \in \mathbb{R}^5$ is the ground-truth reward vector, and $\mathcal{D}$ is the training dataset.

\vspace{-3pt}
In the second stage, we train a gating network to dynamically balance the five-dimensional scores, addressing the multi-objective optimization problem. We introduce a prompt-aware gating network $g_\phi$, implemented as a shallow multi-layer perceptron (MLP). This network dynamically adjusts the model's focus based on the input prompt $x$. The gating network computes non-negative coefficients $w \in \mathbb{R}^5$ for the five reward dimensions. These coefficients are derived from the last hidden state corresponding to the prompt $x$ and normalized via a softmax function.

The gating network is trained using the Bradley-Terry (BT) loss ~\cite{Bradley1952RankAO} function, which aligns the model's predictions with human preferences. The BT loss is formulated as:
\[
L_{BT} = \min_{\phi} \mathbb{E} \left[ -\log \frac{\exp(R_{\text{chosen}})}{\exp(R_{\text{chosen}}) + \exp(R_{\text{rejected}})} \right],
\]
where $R_{\text{chosen}}$ and $R_{\text{rejected}}$ represent the preference scores for the chosen and rejected responses. During training, only the gating network parameters are updated, while the backbone network and regression layer remain frozen.

Finally, the scalarized reward $R$ of the trained \textbf{\texttt{Similar}} model is computed as $R = g_\phi(f_\theta(x))^\top r$. Through this design, our \textbf{\texttt{Similar}} model can not only output five-dimensional scores but also a comprehensive score that balances the five dimensions, just like an all-around expert combining the capabilities of six experts.


\vspace{-7pt}

\section{\textbf{\textit{SRM}} Benchmark}
\label{sec:SRM_benchmark}

\begin{figure}[t]
\includegraphics[width=0.5\textwidth]{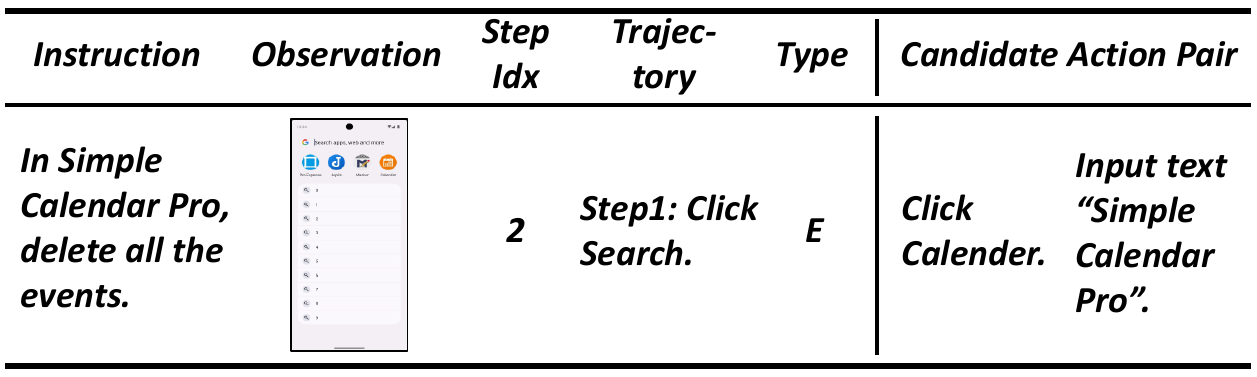}
\vspace{-20pt}
\centering\caption{A data point of \textbf{\textit{SRMEval}}.}
\label{fig:case_of_SRMEval}
\vspace{-2pt}
\end{figure}


We introduce the \textbf{\textit{SRM}} benchmark, built from multi-modal, multi-dimensional, and multi-platform annotated data.  

\textbf{Data Collection.} We used GPT-4o-1120~\cite{hurst2024gpt} as the agent to collect agent action trajectories across four benchmarks—WebArena (WA)~\cite{zhou2024webarena}, VisualWebArena (VWA)~\cite{koh2024visualwebarena}, Android World (AW)~\cite{rawles2024androidworlddynamicbenchmarkingenvironment}, and OSWorld (OW)~\cite{OSWorld}. Since these environments do not provide dedicated training and test sets, to ensure fairness and prevent data leakage, we rigorously used $70\%$ data provided by these benchmarks for agent trajectory data collection, while the remaining $30\%$ were reserved for evaluation experiments to ensure no data overlap between \textbf{\textit{SRMTrain}} and evaluation sets. Ultimately, we collected $10k$ agent trajectories by generating multiple distinct actions per step through task-specific prompt injection and stochastic exploration. And we constructed $110k$ preference pairs for the \textbf{\textit{SRM}} benchmark based on the scores of the designed dimensions. We also sampled some data for human experts to verify the accuracy of the pairs, as detailed in Appendix~\ref{sec:human_acceptance_of_SRM}.

\textbf{Dataset Construction.} We carefully selected $32$k data points for manual annotation as the test set \textbf{\textit{SRMEval}}, while the remaining $78$k data points were used as the training set \textbf{\textit{SRMTrain}} to train the \textbf{\texttt{Similar}} model. The test data tasks are distinct from those in the training data. As shown in Figure~\ref{fig:case_of_SRMEval}, each data point in \textbf{\textit{SRMEval}} includes instruction, observation screenshot, step index, trajectory, evaluation type, and candidate action pair. The evaluation types include our proposed five key dimensions—\textit{Helpfulness} (H), \textit{Odds of Success} (OS), \textit{Efficiency} (E), \textit{Task Relevance} (TR), and \textit{Coherence} (C)—as well as a total dimension that integrates the five dimensions (Tot, weighted sum of the five dimensions) and a trajectory-level dimension (Traj, average Tot score of all steps in trajectory). More visualization cases of \textbf{\textit{SRMEval}} are detailed in Appendix~\ref{sec:visualization_of_srmeval}.

\textbf{New Task and Evaluation Metric.} Based on \textbf{\textit{SRMEval}}, we proposed a new task for reward models in the virtual agent domain: \textit{Selecting the better action from candidate action pair at step $i$ in a specific dimension}. The evaluation metric is Accuracy, measuring the reward model's ability to select the better action. Accuracy is calculated under each evaluation type. For clarity, we use abbreviations such as H to represent each metric in our experiments.

\begin{table}[t]
\caption{Performance comparison of common MLLMs, \textbf{\texttt{Similar}}-RL, and \textbf{\texttt{Similar}}-TM on \textbf{\textit{SRMEval}}.}
\label{tab:model_performance_on_SRM}
\vspace{-0.4cm}
\begin{center}
\scriptsize
\setlength{\tabcolsep}{4pt}
\begin{sc}
\begin{tabular}{lccccccc|c}
\toprule
Reward Model & H & OS & E & TR & C & Tot & Traj & Avg \\
\midrule
GPT-4-Turbo & 44.7 & 46.3 & 44.8 & 48.7 & 42.3 & 46.5 & 44.5 & 46.6 \\
GPT-4o & 49.9 & 50.1 & 47.9 & 51.4 & 43.8 & 51.1 & 49.8 & 51.4 \\
InternVL-2.5 & 38.9 & 43.1 & 44.3 & 41.4 & 41.8 & 40.7 & 39.0 & 40.9 \\
\midrule
Qwen2-VL & 45.7 & 42.1 & 41.7 & 44.5 & 42.1 & 43.2 & 41.6 & 42.9 \\
\quad + \textbf{\texttt{Similar}}-RL & 52.4 & 49.6 & 48.3 & 50.2 & 47.9 & 51.0 & 45.4 & 49.2 \\
\cellcolor{gray!30}\quad + \textbf{\texttt{Similar}}-TM & \cellcolor{gray!30}\textbf{60.5} & \cellcolor{gray!30}\textbf{57.8} & \cellcolor{gray!30}\textbf{56.6} & \cellcolor{gray!30}\textbf{59.7} & \cellcolor{gray!30}\textbf{56.2} & \cellcolor{gray!30}\textbf{58.4} & \cellcolor{gray!30}\textbf{53.9} & \cellcolor{gray!30}\textbf{57.3} \\
\midrule
Llama-3.2-V & 48.2 & 47.6 & 47.1 & 51.1 & 42.6 & 49.5 & 44.5 & 47.6 \\
\quad + \textbf{\texttt{Similar}}-RL & 55.1 & 52.1 & 52.7 & 55.3 & 47.8 & 54.6 & 49.5 & 53.9 \\
\cellcolor{gray!30}\quad + \textbf{\texttt{Similar}}-TM & \cellcolor{gray!30}\cellcolor{gray!30}\textbf{63.8} & \cellcolor{gray!30}\textbf{60.5} & \cellcolor{gray!30}\textbf{59.2} & \cellcolor{gray!30}\textbf{62.7} & \cellcolor{gray!30}\textbf{56.8} & \cellcolor{gray!30}\textbf{61.4} & \cellcolor{gray!30}\textbf{58.7} & \cellcolor{gray!30}\textbf{61.2} \\

\bottomrule
\end{tabular}
\end{sc}
\end{center}
\vskip -0.1in
\vspace{5pt}
\end{table}

\begin{figure*}[t]
\vspace{0.2cm}
\includegraphics[width=0.9\textwidth]{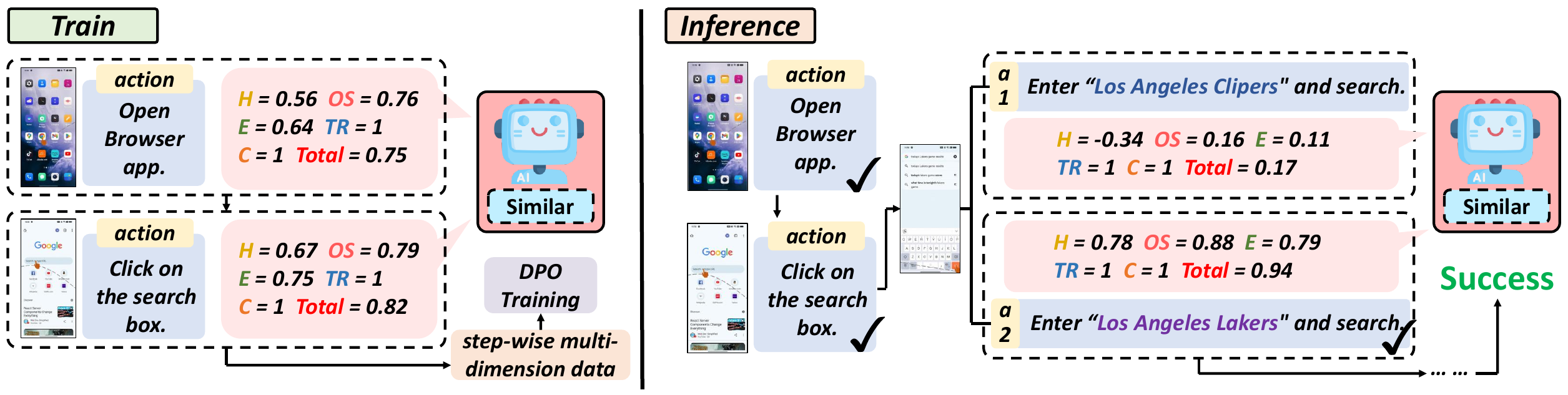}
\centering\caption{A case of \textbf{\texttt{Similar}} provides guidance for GVA training and inference.}  
\label{fig:guidance}
\vspace{-15pt}
\end{figure*}

\vspace{-8pt}

\section{Experiments}
\label{sec:experiments}

\vspace{-3pt}

\subsection{Experimental Setup}


\textbf{Baselines.} We selected two baseline methods: \textbf{1)} Qwen2-VL-7B-Instruct~\cite{Qwen2VL} and Llama-3.2-11B-Vision-Instruct were directly used as reward models, with prompts provided (detailed in in Appendix~\ref{sec:prompt}) to score agent steps. \textbf{2)} \textbf{\texttt{Similar}}-RL-Qwen and \textbf{\texttt{Similar}}-RL-Llama, whose backbones match the aforementioned models, were trained using reinforcement learning~\cite{8103164} on our \textbf{\textit{SRMTrain}} dataset to score agent steps.

\vspace{-3pt}

To benchmark against these baselines, we introduce \textbf{\texttt{Similar}}-TM-Qwen and \textbf{\texttt{Similar}}-TM-Llama, which are trained on the \textbf{\textit{SRMTrain}} dataset using the Triple-M strategy with Qwen2-VL-7B-Instruct and Llama-3.2-11B-Vision-Instruct as backbones, respectively.


\textbf{Evaluation Benchmarks.} We first tested the preference alignment capability of \textbf{\texttt{Similar}} on our \textbf{\textit{SRMEval}}, compared with GPT-4o-1120, GPT-4-Turbo, and InternVL-2.5-8B. Additionally, we evaluated our model's effectiveness as a reward model for virtual agents during both the training and inference phases. \textbf{1)~Training Phase.} Using WebArena and Android World as benchmarks, we employed our model and other reward models to annotate GPT-4o-collected data from these environments, generating preference data. This preference data was then used to perform DPO training on the open-source agents OS-Atlas~\cite{wu2024atlas} and UGround~\cite{gou2024uground}, validating our model's ability to guide agents in the training phase. \textbf{2)~Inference Phase.} With Android World and OSWorld as benchmarks, we used OS-Atlas as the open-source agent and GPT-4o-1120 and GPT-4-Turbo as the closed-source agents. During inference, \textbf{\texttt{Similar}} and other reward models evaluated the agent's simulated $N$ actions, providing rewards and updating the states of nodes in MCTS. Notably, $30\%$ of the examples partitioned from these benchmarks mentioned earlier were used as the evaluation data.

\begin{table}[t]
\caption{Task Success Rates (SR) on Android World and WebArena in training setting.}
\label{tab:sr_on_android_webarena_in_training}
\vspace{-10pt}
\begin{center}
\scriptsize
\begin{sc}
\begin{tabular}{llccc}
\toprule
\textbf{Agent} & \textbf{\makecell{Reward\\Model}} & \textbf{\makecell{Android World\\SR}} & \textbf{\makecell{WebArena\\SR}} \\
\midrule
GPT-4-Turbo & \makecell[c]{/} & 24.1 & 11.2 \\
GPT-4o & \makecell[c]{/} & 25.4 & 12.7 \\
\midrule
UGround & \makecell[c]{/} & 32.4 & 19.6 \\
UGround & Qwen2-VL & 32.6 & 20.2 \\
UGround & \quad + \textbf{\texttt{Similar}}-RL & 33.1 & 26.5 \\
UGround & \quad + \textbf{\texttt{Similar}}-TM & 33.9 & 35.9 \\
UGround & Llama-3.2-V & 33.0 & 23.4 \\
UGround & \quad + \textbf{\texttt{Similar}}-RL & 33.8 & 29.6 \\
\cellcolor{gray!30}UGround & \cellcolor{gray!30}\quad + \textbf{\texttt{Similar}}-TM & \cellcolor{gray!30}\textbf{34.6} & \cellcolor{gray!30}\textbf{36.7} \\
\midrule
OS-Atlas & \makecell[c]{/} & 30.4 & 20.2 \\
OS-Atlas & Qwen2-VL & 30.8 & 20.8 \\
OS-Atlas & \quad + \textbf{\texttt{Similar}}-RL & 32.1 & 25.9 \\
OS-Atlas & \quad + \textbf{\texttt{Similar}}-TM & 34.2 & 34.5 \\
OS-Atlas & Llama-3.2-V & 31.3 & 22.4 \\
OS-Atlas & \quad + \textbf{\texttt{Similar}}-RL & 33.6 & 27.4 \\
\cellcolor{gray!30}OS-Atlas & \cellcolor{gray!30}\quad + \textbf{\texttt{Similar}}-TM & \cellcolor{gray!30}\textbf{34.9} & \cellcolor{gray!30}\textbf{35.6} \\
\bottomrule
\end{tabular}
\end{sc}
\end{center}
\vskip -0.1in
\end{table}


\vspace{-5pt}

\subsection{Effective Alignment of Preference}

\vspace{-3pt}

We first report the performance of the models on \textbf{\textit{SRMEval}} in Table~\ref{tab:model_performance_on_SRM}. The main findings are as follows: \textbf{1)}~Effectiveness of step-wise, multi-dimensional, cross-platform data: Using our collected data for reward modeling, \textbf{\texttt{Similar}}-RL-Llama achieved an Avg score of~$53.9$, remarkably outperforming the baseline Llama-3.2-11B-Vision-Instruct with $47.6$ ($\uparrow13.2\%$) and surpassing closed-source models GPT-4o ($51.4$) and GPT-4-Turbo ($46.6$). It demonstrates that training reward models with our data enables fine-grained, step-based evaluation, providing a more comprehensive and accurate assessment of GVA action quality. \textbf{2)}~Synergistic gain from the Triple-M strategy: \textbf{\texttt{Similar}}-TM-Llama achieved an Avg score of~$61.2$, significantly outperforming \textbf{\texttt{Similar}}-RL-Llama with $53.9$ ($\uparrow13.5\%$). And it achieved higher scores across all dimensions, with improvements such as H increasing from~$48.2$ to~$63.8$ ($\uparrow32.3\%$) and E increasing from~$47.1$ to~$59.2$ ($\uparrow25.6\%$). The \textbf{\texttt{Similar}}-TM-Qwen model showed similar performance. This highlights the effectiveness of our Triple-M strategy, leveraging the complementary expertise of each component to achieve synergistic gain. The experiments demonstrate our model's ability to align preferences.

\begin{table}[t]
\caption{Task Success Rates (SR) on Android World and OSWorld in inference setting.}
\label{tab:sr_on_android_osworld_in_inference}
\vspace{-10pt}
\begin{center}
\scriptsize
\begin{sc}
\begin{tabular}{llcc}
\toprule
\textbf{Agent} & \textbf{\makecell{Reward\\Model}} & \textbf{\makecell{Android World\\SR}} & \textbf{\makecell{OSWorld\\SR}} \\
\midrule
GPT-4-Turbo & \makecell[c]{/} & 24.1 & 8.4 \\
GPT-4-Turbo & Qwen2-VL & 24.9 & 8.9 \\
GPT-4-Turbo & \quad + \textbf{\texttt{Similar}}-RL & 25.9 & 10.5 \\
GPT-4-Turbo & \quad + \textbf{\texttt{Similar}}-TM & 28.3 & 13.4 \\
GPT-4-Turbo & Llama-3.2-V & 25.3 & 8.8 \\
GPT-4-Turbo & \quad + \textbf{\texttt{Similar}}-RL & 26.5 & 10.8 \\
\cellcolor{gray!30}GPT-4-Turbo & \cellcolor{gray!30}\quad + \textbf{\texttt{Similar}}-TM & \cellcolor{gray!30}\textbf{30.4} & \cellcolor{gray!30}\textbf{13.9} \\
\midrule
GPT-4o & \makecell[c]{/} & 25.4 & 10.8 \\
GPT-4o & Qwen2-VL & 26.0 & 11.3 \\
GPT-4o & \quad + \textbf{\texttt{Similar}}-RL & 27.1 & 12.9 \\
GPT-4o & \quad + \textbf{\texttt{Similar}}-TM & 32.9 & 14.3 \\
GPT-4o & Llama-3.2-V & 26.2 & 11.7 \\
GPT-4o & \quad + \textbf{\texttt{Similar}}-RL & 29.6 & 13.1 \\
\cellcolor{gray!30}GPT-4o & \cellcolor{gray!30}\quad + \textbf{\texttt{Similar}}-TM & \cellcolor{gray!30}\textbf{34.6} & \cellcolor{gray!30}\textbf{16.5} \\
\midrule
OS-Atlas & \makecell[c]{/} & 30.4 & 14.3 \\
OS-Atlas & Qwen2-VL & 30.9 & 14.8 \\
OS-Atlas & \quad + \textbf{\texttt{Similar}}-RL & 32.0 & 15.4 \\
OS-Atlas & \quad + \textbf{\texttt{Similar}}-TM & 34.5 & 16.4 \\
OS-Atlas & Llama-3.2-V & 31.5 & 14.8 \\
OS-Atlas & \quad + \textbf{\texttt{Similar}}-RL & 32.9 & 15.7 \\
\cellcolor{gray!30}OS-Atlas & \cellcolor{gray!30}\quad + \textbf{\texttt{Similar}}-TM & \cellcolor{gray!30}\textbf{35.4} & \cellcolor{gray!30}\textbf{17.8} \\
\bottomrule
\end{tabular}
\end{sc}
\end{center}
\vskip -0.1in
\end{table}


\begin{table}[t]
\caption{Abaltion study (inference experiments). \textbf{\texttt{Similar}} in table represents \textbf{\texttt{Similar}}-TM-Llama. }
\label{tab:ablation_study}
\renewcommand{\arraystretch}{1.1}
\vskip 0.15in
\vspace{-0.7cm}
\begin{center}
\scriptsize
\begin{sc}
\begin{tabular}{lccccc|cc}
\toprule
& \multicolumn{5}{c}{\textbf{Dimension}} & \multicolumn{2}{c}{\textbf{Success Rate}} \\
\cmidrule(lr){2-6} \cmidrule(lr){7-8}
Model & H & OS & E & TR & C & \makecell{AW} & \makecell{WA} \\
\midrule
Backbone & & & & & & 30.4 & 20.6 \\
\midrule
+H & $\checkmark$ & & & & & 32.5 & 26.1 \\
+OS & & $\checkmark$ & & & & 31.9 & 24.7 \\
+E & & & $\checkmark$ & & & 31.6 & 23.3 \\
+TR & & & & $\checkmark$ & & 31.1 & 21.6 \\
+C & & & & & $\checkmark$ & 30.9 & 21.0 \\
\midrule
+OS,E & & $\checkmark$ & $\checkmark$ & & & 32.7 & 27.5 \\
+H,E & $\checkmark$ & & $\checkmark$ & & & 33.1 & 29.8 \\
+H,OS & $\checkmark$ & $\checkmark$ & & & & 33.4 & 31.4 \\
+TR,C & & & & $\checkmark$ & $\checkmark$ & 31.5 & 22.5 \\
+H,OS,E & $\checkmark$ & $\checkmark$ & $\checkmark$ & & & 34.3 & 35.9 \\
\midrule
+OS,E,TR,C & & $\checkmark$ & $\checkmark$ & $\checkmark$ & $\checkmark$ & 33.1 & 33.9 \\
+H,E,TR,C & $\checkmark$ & & $\checkmark$ & $\checkmark$ & $\checkmark$ & 33.9 & 35.7 \\
+H,OS,TR,C & $\checkmark$ & $\checkmark$ & & $\checkmark$ & $\checkmark$ & 34.2 & 36.5 \\
+H,OS,E,C & $\checkmark$ & $\checkmark$ & $\checkmark$ & & $\checkmark$ & 34.7 & 37.2 \\
+H,OS,E,TR & $\checkmark$ & $\checkmark$ & $\checkmark$ & $\checkmark$ & & 35.1 & 37.7 \\
\midrule
\textbf{\texttt{Similar}} & $\checkmark$ & $\checkmark$ & $\checkmark$ & $\checkmark$ & $\checkmark$ & \textbf{35.4} & \textbf{38.2} \\
\bottomrule
\end{tabular}
\end{sc}
\end{center}
\vskip -0.1in
\vspace{3pt}
\end{table}




\vspace{-5pt}
\subsection{\textbf{\texttt{Similar}} for RL Training}

\vspace{-5pt}

We used GPT-4o and multiple reward models to annotate reward data across benchmark environments. The annotated data was then used to train the final agent via DPO. The results, shown in Table~\ref{tab:sr_on_android_webarena_in_training}, demonstrate that our model significantly improves agent learning: \textbf{1)} The \textbf{\texttt{Similar}}-RL model derived through Reward Modeling on the \textbf{\textit{SRMTrain}} dataset outperforms the baseline. When using OS-Atlas as the agent, \textbf{\texttt{Similar}}-RL-Llama achieves improvements of $10.5\%$ ($30.4\rightarrow33.6$) and $7.3\%$ ($31.3\rightarrow33.6$) over the original OS-Atlas model and the setting using Llama-3.2V as the reward model, respectively, on Android World. On WebArena, the improvements are $35.6\%$ ($20.2\rightarrow27.4$) and $22.3\%$ ($22.4\rightarrow27.4$), respectively. \textbf{2)} The \textbf{\texttt{Similar}}-TM model performed best. With OS-Atlas, \textbf{\texttt{Similar}}-TM-Llama achieved improvements of~$3.8\%$ ($33.6\rightarrow34.9$) and~$29.9\%$($27.4\rightarrow35.6$) on Android World and WebArena, respectively, compared to \textbf{\texttt{Similar}}-RL-Llama. \textbf{3)} When using UGround as the agent or adopting Qwen2-VL as the baseline reward model, comparable performance can be observed. The consistent performance improvements across different models and environments demonstrate that our method enhances virtual agents' learning capabilities.


\vspace{-4pt}
\subsection{\textbf{\texttt{Similar}} for Inference-Time Scaling}

\vspace{-3pt}

During inference, we used various reward models to evaluate the agent's $N$ simulated actions, providing rewards and updating MCTS node states. Table~\ref{tab:sr_on_android_osworld_in_inference} shows that our model effectively guides the agent: \textbf{1)} Consistent with the training setup, the \textbf{\texttt{Similar}}-RL model outperformed both the original agent without a reward model and the setting using MLLM as the reward model. With GPT-4o, \textbf{\texttt{Similar}}-RL-Llama achieved improvements of~$16.5\%$ ($25.4\rightarrow29.6$) and~$12.9\%$ ($26.2\rightarrow29.6$) on Android World for these two settings, respectively. A similar performance is observed on OSWorld. \textbf{2)} The \textbf{\texttt{Similar}}-TM model performed best. With GPT-4o, \textbf{\texttt{Similar}}-TM-Llama achieved improvements of~$16.8\%$ ($29.6\rightarrow34.6$) and~$25.9\%$ ($13.1\rightarrow16.5$) on Android World and OSWorld, respectively, compared to \textbf{\texttt{Similar}}-RL-Llama. \textbf{3)} When employing GPT-4-Turbo, GPT-4o, or OS-Atlas as the agent, or when using Qwen2-VL as the baseline reward model, we consistently observe similar model performance. It can be concluded that our method is generalizable and effectively enhances the virtual agent's inference ability.

\begin{figure}[t]
    \centering
    \begin{minipage}[t]{0.48\linewidth}
        \centering
        \includegraphics[width=\linewidth]{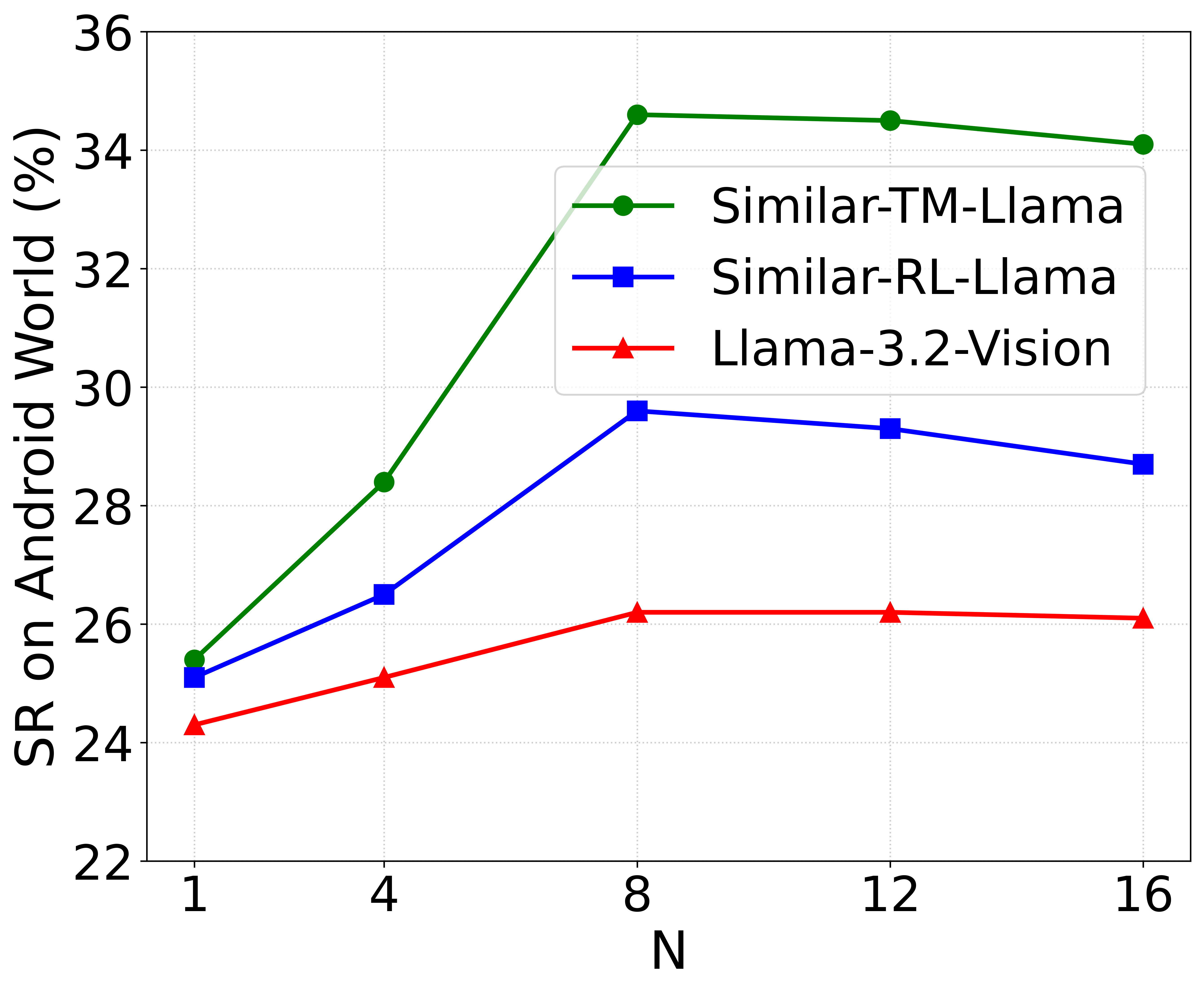}
        \vspace{-20pt}
        \caption*{(a)}
    \end{minipage}%
    \hfill%
    \begin{minipage}[t]{0.45\linewidth}
        \centering
        \includegraphics[width=\linewidth]{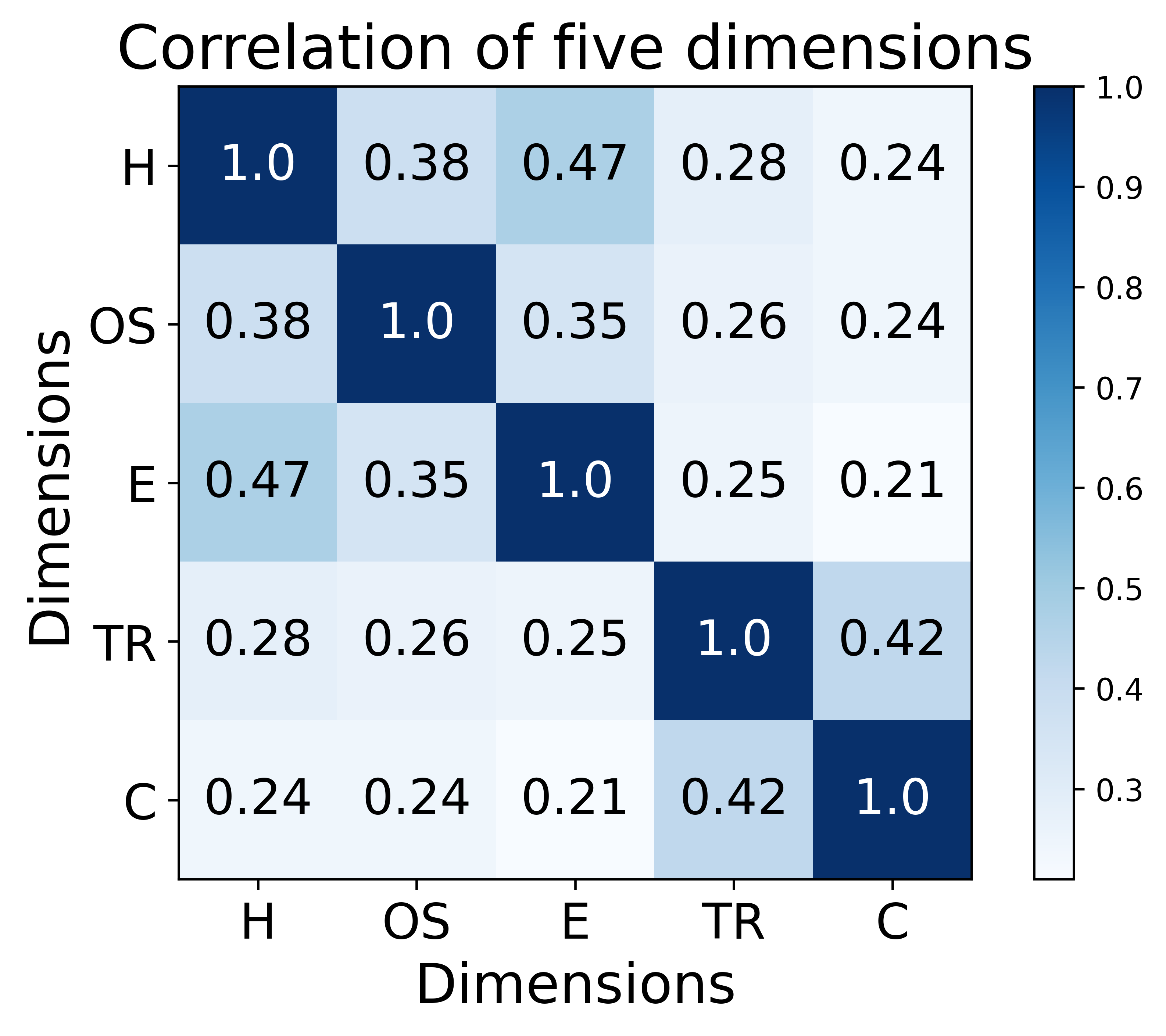}
        \vspace{-20pt}
        \caption*{(b)}
    \end{minipage}
    \vspace{-5pt}
    \caption{(a) Inference-time scaling research. The agent is GPT-4o.  (b) Correlation research of five dimensions.}
    \label{fig:6}
\end{figure}

\vspace{-3pt}
We further demonstrate that our model is essential
for scaling the inference-time capabilities of agents by varying the number of child nodes $N$ in MCTS. As shown in Figure~\ref{fig:6}~(a): 1) When $N \leq 8$, agent performance improves. However, when $N > 8$, performance plateaus or declines, likely due to limitations in the agent model, as simulating more actions fails to identify viable paths. 2) \textbf{\texttt{Similar}}-RL and \textbf{\texttt{Similar}}-TM outperform other settings, with \textbf{Similar}-RL surpassing MLLM-based reward models and \textbf{\texttt{Similar}}-TM exceeding \textbf{\texttt{Similar}}-RL. These results demonstrate the superiority of our models while highlighting the challenges of scaling inference-time in agent systems.


\vspace{-6pt}

\subsection{Indepth Analysis}

\vspace{-5pt}

\textbf{Ablation study.} The results, shown in Table~\ref{tab:ablation_study}, show that models with partial-dimensional rewards underperformed compared to \textbf{\texttt{Similar}}. For example, on Android World, models excluding H, OS, E, TR, and C rewards showed declines of $6.9\%$, $4.4\%$, $3.5\%$, $2.0\%$, and $0.8\%$, respectively, with similar trends on WebArena. Analysis reveals that the H dimension has the most significant impact, as \textit{Helpfulness} captures a step's contribution to task completion. The OS dimension follows closely, reflecting the influence of the current step on the next step. The C dimension has the least impact, as agent actions are often inherently coherent and contextually aligned. These results confirm that fine-grained rewards outperform coarse-grained ones and that our five dimensions comprehensively assess agent actions. More comprehensive results can be found in Appendix~\ref{sec:ablation_more}.




\vspace{-3pt}

\textbf{Case Study.} To demonstrate the role of our model in training and inference, we included visual cases, as shown in Figure~\ref{fig:guidance}. During training, \textbf{\texttt{Similar}} annotates the agent's trajectory with multi-dimensional scores, used for DPO training. During inference, the agent simulates multiple actions for a single step, and \textbf{\texttt{Similar}} evaluates these actions. In the figure, our model assigns high scores to action~$2$ at the third step, with a total score of~$0.94$, while action~$1$ receives lower scores. Therefore, action~$2$, the highest-scoring action, is easily selected for the current step. More case studies are detailed in Appendix~\ref{sec:case_study_of_similar}.

\vspace{-3pt}

\textbf{Correlation Study.} The Pearson correlation coefficients among the five dimensions are calculated to analyze their independence, as shown in Figure~\ref{fig:6} (b). The results show that while some correlation exists among the five dimensions, the values are all below~$0.47$, indicating independence.
\vspace{-10pt}

\section{Conclusion}
\label{conslusion}

\vspace{-5pt}

In this work, we introduce a novel reward model-based paradigm for training GVAs. Our reward model, \textbf{\texttt{Similar}}, provides step-wise, multi-dimensional feedback during GVAs' training and inference, enabling fine-grained assessment. Additionally, we build the first reward model evaluation benchmark called \textbf{\textit{SRM}}. Extensive experiments demonstrate our model’s superior performance on \textbf{\textit{SRMEval} }and its effectiveness in guiding GVAs across diverse tasks.

\vspace{-14pt}
\paragraph{Acknowledgment.} This work was supported by the NSFC (62272411), the Fundamental Research Funds for the Central Universities (226-2025-00017), the Key R\&D Projects in Zhejiang Province (No. 2024C01106, 2025C01030), Ningbo Yongjiang Talent Introduction Programme(2024A-401-G),the Zhejiang NSF (LRG25F020001),  Ant Group.


\vspace{-6pt}
\section*{Impact Statement}

\vspace{-2pt}
\subsection*{Limitations and Future Investigation}

\vspace{-3pt}
While \textbf{\texttt{Similar}} demonstrates significant advancements in the training and inference of GVAs, several limitations remain. First, the current framework relies heavily on the quality of the automatically annotated data generated by the MCTS-P algorithm. Although this approach reduces human annotation efforts, it may introduce biases or inaccuracies in the reward signals, particularly in complex or ambiguous task scenarios. Future work should focus on improving the robustness of the automatic annotation process, potentially by incorporating more sophisticated error-correction mechanisms or hybrid human-AI annotation strategies.

\vspace{-3pt}
Moreover, the scalability of \textbf{\texttt{Similar}} across diverse environments and tasks is promising but not yet fully explored. While the current experiments cover four major platforms (Web, Android, Linux, and Windows), the model's performance in more niche or specialized domains remains untested. Future investigations should extend the evaluation to a broader range of environments, including those with less structured or more dynamic interfaces, to ensure the generalizability of the approach.



\vspace{-8pt}
\subsection*{Impact on RL Training}

\vspace{-3pt}
\textbf{\texttt{Similar}} has a transformative impact on reinforcement learning (RL) training for GVAs. By providing fine-grained, multi-dimensional feedback, \textbf{\texttt{Similar}} enables more efficient and effective learning compared to traditional outcome-based reward models. The step-wise, multi-dimensional assessment allows the agent to identify and correct errors at intermediate stages, leading to faster convergence and improved task performance. 


The Triple-M strategy further enhances the RL training process by integrating multiple dimensions of assessment and leveraging the strengths of different experts. This synergistic approach not only improves the accuracy of the reward signals but also ensures that the agent learns robust policies that generalize well across diverse tasks and environments. As a result, \textbf{\texttt{Similar}} significantly reduces the reliance on labor-intensive human annotations, making RL training more scalable and cost-effective.

\vspace{-5pt}
\subsection*{Impact on Inference-Time Scaling}

\vspace{-5pt}
Recent outstanding work has demonstrated that inference-time scaling can significantly enhance agent performance~\cite{deepseekai2025deepseekr1incentivizingreasoningcapability, wu2024comparative}. \textbf{\texttt{Similar}} significantly enhances the inference-time capabilities of GVAs by integrating with search algorithms like MCTS. During inference, our model provides fine-grained, multi-dimensional rewards to evaluate and filter candidate actions, ensuring the agent selects the most promising paths. This not only improves task completion accuracy but also reduces computational overhead, making it highly efficient for real-time applications.

Experiments demonstrate that our model consistently outperforms baseline models and MLLM-based reward settings across diverse environments, such as Android World and OSWorld. The \textbf{\texttt{Similar}}-TM model, leveraging the Triple-M strategy, achieves the best performance, highlighting the synergistic gains from integrating multiple dimensions of assessment. Furthermore, our model effectively scales inference-time computations, with performance improvements observed when expanding the number of child nodes in MCTS, though performance plateaus beyond a certain threshold due to inherent agent limitations.

\vspace{-5pt}
\subsection*{Impact on Data Cleaning}

\vspace{-5pt}
\textbf{\texttt{Similar}} also has significant implications for data cleaning in the context of GVA training. The model's ability to provide fine-grained, multi-dimensional feedback allows for the precise identification and systematic removal of low-quality or noisy data points. This is particularly useful in large-scale datasets where manual inspection is impractical. By filtering out irrelevant or incoherent actions and prioritizing consistent, task-aligned examples, our model ensures that the training data is of high quality, leading to more robust and reliable agent performance.

Moreover, the automatic annotation process introduced by \textbf{\texttt{Similar}} reduces the need for human intervention in data cleaning, further enhancing the scalability of GVA training. This is especially beneficial in domains where data is abundant but of varying quality, such as web navigation or mobile app interaction. By improving the quality of the training data, our model contributes to the overall efficiency and effectiveness of GVA training pipelines.



\bibliography{main}

\begin{thebibliography}{60}
\providecommand{\natexlab}[1]{#1}
\providecommand{\url}[1]{\texttt{#1}}
\expandafter\ifx\csname urlstyle\endcsname\relax
  \providecommand{\doi}[1]{doi: #1}\else
  \providecommand{\doi}{doi: \begingroup \urlstyle{rm}\Url}\fi

\bibitem[Arulkumaran et~al.(2017)Arulkumaran, Deisenroth, Brundage, and Bharath]{8103164}
Arulkumaran, K., Deisenroth, M.~P., Brundage, M., and Bharath, A.~A.
\newblock Deep reinforcement learning: A brief survey.
\newblock \emph{IEEE Signal Processing Magazine}, 34\penalty0 (6):\penalty0 26--38, 2017.
\newblock \doi{10.1109/MSP.2017.2743240}.

\bibitem[Bradley \& Terry(1952)Bradley and Terry]{Bradley1952RankAO}
Bradley, R.~A. and Terry, M.~E.
\newblock Rank analysis of incomplete block designs: I. the method of paired comparisons.
\newblock \emph{Biometrika}, 39:\penalty0 324, 1952.
\newblock URL \url{https://api.semanticscholar.org/CorpusID:125209808}.

\bibitem[Branavan et~al.(2009)Branavan, Chen, Zettlemoyer, and Barzilay]{branavan-etal-2009-reinforcement}
Branavan, S., Chen, H., Zettlemoyer, L., and Barzilay, R.
\newblock Reinforcement learning for mapping instructions to actions.
\newblock In Su, K.-Y., Su, J., Wiebe, J., and Li, H. (eds.), \emph{Proceedings of the Joint Conference of the 47th Annual Meeting of the {ACL} and the 4th International Joint Conference on Natural Language Processing of the {AFNLP}}, pp.\  82--90, Suntec, Singapore, August 2009. Association for Computational Linguistics.
\newblock URL \url{https://aclanthology.org/P09-1010/}.

\bibitem[Branavan et~al.(2010)Branavan, Zettlemoyer, and Barzilay]{branavan-etal-2010-reading}
Branavan, S., Zettlemoyer, L., and Barzilay, R.
\newblock Reading between the lines: Learning to map high-level instructions to commands.
\newblock In Haji{\v{c}}, J., Carberry, S., Clark, S., and Nivre, J. (eds.), \emph{Proceedings of the 48th Annual Meeting of the Association for Computational Linguistics}, pp.\  1268--1277, Uppsala, Sweden, July 2010. Association for Computational Linguistics.
\newblock URL \url{https://aclanthology.org/P10-1129/}.

\bibitem[Bu et~al.(2025)Bu, Wu, Yu, Gao, Miao, Zhang, Pan, Li, Li, Ji, Li, Tang, and Zhuang]{bu2025limitsvirtualagentapplication}
Bu, W., Wu, Y., Yu, Q., Gao, M., Miao, B., Zhang, Z., Pan, K., Li, Y., Li, M., Ji, W., Li, J., Tang, S., and Zhuang, Y.
\newblock What limits virtual agent application? omnibench: A scalable multi-dimensional benchmark for essential virtual agent capabilities, 2025.
\newblock URL \url{https://arxiv.org/abs/2506.08933}.

\bibitem[Burns et~al.(2022)Burns, Arsan, Agrawal, Kumar, Saenko, and Plummer]{burns2022motifvln}
Burns, A., Arsan, D., Agrawal, S., Kumar, R., Saenko, K., and Plummer, B.~A.
\newblock A dataset for interactive vision language navigation with unknown command feasibility.
\newblock In \emph{European Conference on Computer Vision (ECCV)}, 2022.

\bibitem[Christianos et~al.(2023)Christianos, Papoudakis, Zimmer, Coste, Wu, Chen, Khandelwal, Doran, Feng, Liu, Xiong, Luo, Hao, Shao, Bou-Ammar, and Wang]{christianos2023panguagentfinetunablegeneralistagent}
Christianos, F., Papoudakis, G., Zimmer, M., Coste, T., Wu, Z., Chen, J., Khandelwal, K., Doran, J., Feng, X., Liu, J., Xiong, Z., Luo, Y., Hao, J., Shao, K., Bou-Ammar, H., and Wang, J.
\newblock Pangu-agent: A fine-tunable generalist agent with structured reasoning, 2023.
\newblock URL \url{https://arxiv.org/abs/2312.14878}.

\bibitem[DeepSeek-AI et~al.(2025)DeepSeek-AI, Guo, Yang, Zhang, Song, Zhang, Xu, Zhu, Ma, Wang, Bi, Zhang, Yu, Wu, Wu, Gou, Shao, Li, Gao, Liu, Xue, Wang, Wu, Feng, Lu, Zhao, Deng, Zhang, Ruan, Dai, Chen, Ji, Li, Lin, Dai, Luo, Hao, Chen, Li, Zhang, Bao, Xu, Wang, Ding, Xin, Gao, Qu, Li, Guo, Li, Wang, Chen, Yuan, Qiu, Li, Cai, Ni, Liang, Chen, Dong, Hu, Gao, Guan, Huang, Yu, Wang, Zhang, Zhao, Wang, Zhang, Xu, Xia, Zhang, Zhang, Tang, Li, Wang, Li, Tian, Huang, Zhang, Wang, Chen, Du, Ge, Zhang, Pan, Wang, Chen, Jin, Chen, Lu, Zhou, Chen, Ye, Wang, Yu, Zhou, Pan, Li, Zhou, Wu, Ye, Yun, Pei, Sun, Wang, Zeng, Zhao, Liu, Liang, Gao, Yu, Zhang, Xiao, An, Liu, Wang, Chen, Nie, Cheng, Liu, Xie, Liu, Yang, Li, Su, Lin, Li, Jin, Shen, Chen, Sun, Wang, Song, Zhou, Wang, Shan, Li, Wang, Wei, Zhang, Xu, Li, Zhao, Sun, Wang, Yu, Zhang, Shi, Xiong, He, Piao, Wang, Tan, Ma, Liu, Guo, Ou, Wang, Gong, Zou, He, Xiong, Luo, You, Liu, Zhou, Zhu, Xu, Huang, Li, Zheng, Zhu, Ma, Tang, Zha, Yan, Ren, Ren, Sha, Fu, Xu, Xie, Zhang,
  Hao, Ma, Yan, Wu, Gu, Zhu, Liu, Li, Xie, Song, Pan, Huang, Xu, Zhang, and Zhang]{deepseekai2025deepseekr1incentivizingreasoningcapability}
DeepSeek-AI, Guo, D., Yang, D., Zhang, H., Song, J., Zhang, R., Xu, R., Zhu, Q., Ma, S., Wang, P., Bi, X., Zhang, X., Yu, X., Wu, Y., Wu, Z.~F., Gou, Z., Shao, Z., Li, Z., Gao, Z., Liu, A., Xue, B., Wang, B., Wu, B., Feng, B., Lu, C., Zhao, C., Deng, C., Zhang, C., Ruan, C., Dai, D., Chen, D., Ji, D., Li, E., Lin, F., Dai, F., Luo, F., Hao, G., Chen, G., Li, G., Zhang, H., Bao, H., Xu, H., Wang, H., Ding, H., Xin, H., Gao, H., Qu, H., Li, H., Guo, J., Li, J., Wang, J., Chen, J., Yuan, J., Qiu, J., Li, J., Cai, J.~L., Ni, J., Liang, J., Chen, J., Dong, K., Hu, K., Gao, K., Guan, K., Huang, K., Yu, K., Wang, L., Zhang, L., Zhao, L., Wang, L., Zhang, L., Xu, L., Xia, L., Zhang, M., Zhang, M., Tang, M., Li, M., Wang, M., Li, M., Tian, N., Huang, P., Zhang, P., Wang, Q., Chen, Q., Du, Q., Ge, R., Zhang, R., Pan, R., Wang, R., Chen, R.~J., Jin, R.~L., Chen, R., Lu, S., Zhou, S., Chen, S., Ye, S., Wang, S., Yu, S., Zhou, S., Pan, S., Li, S.~S., Zhou, S., Wu, S., Ye, S., Yun, T., Pei, T., Sun, T., Wang, T., Zeng, W.,
  Zhao, W., Liu, W., Liang, W., Gao, W., Yu, W., Zhang, W., Xiao, W.~L., An, W., Liu, X., Wang, X., Chen, X., Nie, X., Cheng, X., Liu, X., Xie, X., Liu, X., Yang, X., Li, X., Su, X., Lin, X., Li, X.~Q., Jin, X., Shen, X., Chen, X., Sun, X., Wang, X., Song, X., Zhou, X., Wang, X., Shan, X., Li, Y.~K., Wang, Y.~Q., Wei, Y.~X., Zhang, Y., Xu, Y., Li, Y., Zhao, Y., Sun, Y., Wang, Y., Yu, Y., Zhang, Y., Shi, Y., Xiong, Y., He, Y., Piao, Y., Wang, Y., Tan, Y., Ma, Y., Liu, Y., Guo, Y., Ou, Y., Wang, Y., Gong, Y., Zou, Y., He, Y., Xiong, Y., Luo, Y., You, Y., Liu, Y., Zhou, Y., Zhu, Y.~X., Xu, Y., Huang, Y., Li, Y., Zheng, Y., Zhu, Y., Ma, Y., Tang, Y., Zha, Y., Yan, Y., Ren, Z.~Z., Ren, Z., Sha, Z., Fu, Z., Xu, Z., Xie, Z., Zhang, Z., Hao, Z., Ma, Z., Yan, Z., Wu, Z., Gu, Z., Zhu, Z., Liu, Z., Li, Z., Xie, Z., Song, Z., Pan, Z., Huang, Z., Xu, Z., Zhang, Z., and Zhang, Z.
\newblock Deepseek-r1: Incentivizing reasoning capability in llms via reinforcement learning, 2025.
\newblock URL \url{https://arxiv.org/abs/2501.12948}.

\bibitem[Deng et~al.(2023)Deng, Gu, Zheng, Chen, Stevens, Wang, Sun, and Su]{NEURIPS2023_5950bf29}
Deng, X., Gu, Y., Zheng, B., Chen, S., Stevens, S., Wang, B., Sun, H., and Su, Y.
\newblock Mind2web: Towards a generalist agent for the web.
\newblock In Oh, A., Naumann, T., Globerson, A., Saenko, K., Hardt, M., and Levine, S. (eds.), \emph{Advances in Neural Information Processing Systems}, volume~36, pp.\  28091--28114. Curran Associates, Inc., 2023.
\newblock URL \url{https://proceedings.neurips.cc/paper_files/paper/2023/file/5950bf290a1570ea401bf98882128160-Paper-Datasets_and_Benchmarks.pdf}.

\bibitem[Fei et~al.(2024)Fei, Wu, Zhang, Chua, and Yan]{fei2024vitron}
Fei, H., Wu, S., Zhang, H., Chua, T.-S., and Yan, S.
\newblock Vitron: A unified pixel-level vision llm for understanding, generating, segmenting, editing.
\newblock 2024.

\bibitem[Gao et~al.(2024)Gao, Bu, Miao, Wu, Li, Li, Tang, Wu, Zhuang, and Wang]{gao2024generalistvirtualagentssurvey}
Gao, M., Bu, W., Miao, B., Wu, Y., Li, Y., Li, J., Tang, S., Wu, Q., Zhuang, Y., and Wang, M.
\newblock Generalist virtual agents: A survey on autonomous agents across digital platforms, 2024.
\newblock URL \url{https://arxiv.org/abs/2411.10943}.

\bibitem[Gao et~al.(2025)Gao, Liu, Yue, Wu, Chen, Li, Tang, Wu, Chua, and Zhuang]{gao2025benchmarkingmultimodalcotreward}
Gao, M., Liu, X., Yue, Z., Wu, Y., Chen, S., Li, J., Tang, S., Wu, F., Chua, T.-S., and Zhuang, Y.
\newblock Benchmarking multimodal cot reward model stepwise by visual program, 2025.
\newblock URL \url{https://arxiv.org/abs/2504.06606}.

\bibitem[Gou et~al.(2024)Gou, Wang, Zheng, Xie, Chang, Shu, Sun, and Su]{gou2024uground}
Gou, B., Wang, R., Zheng, B., Xie, Y., Chang, C., Shu, Y., Sun, H., and Su, Y.
\newblock Navigating the digital world as humans do: Universal visual grounding for gui agents.
\newblock \emph{arXiv preprint arXiv:2410.05243}, 2024.
\newblock URL \url{https://arxiv.org/abs/2410.05243}.

\bibitem[He et~al.(2024)He, Yao, Ma, Yu, Dai, Zhang, Lan, and Yu]{he2024webvoyager}
He, H., Yao, W., Ma, K., Yu, W., Dai, Y., Zhang, H., Lan, Z., and Yu, D.
\newblock Webvoyager: Building an end-to-end web agent with large multimodal models.
\newblock \emph{arXiv preprint arXiv:2401.13919}, 2024.

\bibitem[Humphreys et~al.(2022)Humphreys, Raposo, Pohlen, Thornton, Chhaparia, Muldal, Abramson, Georgiev, Santoro, and Lillicrap]{pmlr-v162-humphreys22a}
Humphreys, P.~C., Raposo, D., Pohlen, T., Thornton, G., Chhaparia, R., Muldal, A., Abramson, J., Georgiev, P., Santoro, A., and Lillicrap, T.
\newblock A data-driven approach for learning to control computers.
\newblock In Chaudhuri, K., Jegelka, S., Song, L., Szepesvari, C., Niu, G., and Sabato, S. (eds.), \emph{Proceedings of the 39th International Conference on Machine Learning}, volume 162 of \emph{Proceedings of Machine Learning Research}, pp.\  9466--9482. PMLR, 17--23 Jul 2022.
\newblock URL \url{https://proceedings.mlr.press/v162/humphreys22a.html}.

\bibitem[Hurst et~al.(2024)Hurst, Lerer, Goucher, Perelman, Ramesh, Clark, Ostrow, Welihinda, Hayes, Radford, et~al.]{hurst2024gpt}
Hurst, A., Lerer, A., Goucher, A.~P., Perelman, A., Ramesh, A., Clark, A., Ostrow, A., Welihinda, A., Hayes, A., Radford, A., et~al.
\newblock Gpt-4o system card.
\newblock \emph{arXiv preprint arXiv:2410.21276}, 2024.

\bibitem[Koh et~al.(2024)Koh, Lo, Jang, Duvvur, Lim, Huang, Neubig, Zhou, Salakhutdinov, and Fried]{koh2024visualwebarena}
Koh, J.~Y., Lo, R., Jang, L., Duvvur, V., Lim, M.~C., Huang, P.-Y., Neubig, G., Zhou, S., Salakhutdinov, R., and Fried, D.
\newblock Visualwebarena: Evaluating multimodal agents on realistic visual web tasks.
\newblock \emph{arXiv preprint arXiv:2401.13649}, 2024.

\bibitem[Lai et~al.(2024)Lai, Tian, Chen, Yang, Peng, and Jia]{lai2024stepdpo}
Lai, X., Tian, Z., Chen, Y., Yang, S., Peng, X., and Jia, J.
\newblock Step-dpo: Step-wise preference optimization for long-chain reasoning of llms.
\newblock \emph{arXiv:2406.18629}, 2024.

\bibitem[Li et~al.(2020)Li, Wang, Tang, Shi, Wu, Zhuang, and Wang]{li2020unsupervised}
Li, J., Wang, X., Tang, S., Shi, H., Wu, F., Zhuang, Y., and Wang, W.~Y.
\newblock Unsupervised reinforcement learning of transferable meta-skills for embodied navigation.
\newblock In \emph{Proceedings of the IEEE/CVF Conference on Computer Vision and Pattern Recognition}, pp.\  12123--12132, 2020.

\bibitem[Li et~al.(2022)Li, He, Wei, Qian, Zhu, Xie, Zhuang, Tian, and Tang]{li2022fine}
Li, J., He, X., Wei, L., Qian, L., Zhu, L., Xie, L., Zhuang, Y., Tian, Q., and Tang, S.
\newblock Fine-grained semantically aligned vision-language pre-training.
\newblock \emph{Advances in neural information processing systems}, 35:\penalty0 7290--7303, 2022.

\bibitem[Li et~al.(2023{\natexlab{a}})Li, Pan, Ge, Gao, Ji, Zhang, Chua, Tang, Zhang, and Zhuang]{li2023fine}
Li, J., Pan, K., Ge, Z., Gao, M., Ji, W., Zhang, W., Chua, T.-S., Tang, S., Zhang, H., and Zhuang, Y.
\newblock Fine-tuning multimodal llms to follow zero-shot demonstrative instructions.
\newblock In \emph{The Twelfth International Conference on Learning Representations}, 2023{\natexlab{a}}.

\bibitem[Li et~al.(2023{\natexlab{b}})Li, Tang, Zhu, Zhang, Yang, Chua, Wu, and Zhuang]{10121664}
Li, J., Tang, S., Zhu, L., Zhang, W., Yang, Y., Chua, T.-S., Wu, F., and Zhuang, Y.
\newblock Variational cross-graph reasoning and adaptive structured semantics learning for compositional temporal grounding.
\newblock \emph{IEEE Transactions on Pattern Analysis and Machine Intelligence}, 45\penalty0 (10):\penalty0 12601--12617, 2023{\natexlab{b}}.
\newblock \doi{10.1109/TPAMI.2023.3274139}.

\bibitem[Li et~al.(2023{\natexlab{c}})Li, Lin, Zhang, Fu, Chen, Lou, and Chen]{li-etal-2023-making}
Li, Y., Lin, Z., Zhang, S., Fu, Q., Chen, B., Lou, J.-G., and Chen, W.
\newblock Making language models better reasoners with step-aware verifier.
\newblock In Rogers, A., Boyd-Graber, J., and Okazaki, N. (eds.), \emph{Proceedings of the 61st Annual Meeting of the Association for Computational Linguistics (Volume 1: Long Papers)}, pp.\  5315--5333, Toronto, Canada, July 2023{\natexlab{c}}. Association for Computational Linguistics.
\newblock \doi{10.18653/v1/2023.acl-long.291}.
\newblock URL \url{https://aclanthology.org/2023.acl-long.291/}.

\bibitem[Lightman et~al.(2023)Lightman, Kosaraju, Burda, Edwards, Baker, Lee, Leike, Schulman, Sutskever, and Cobbe]{lightman2023lets}
Lightman, H., Kosaraju, V., Burda, Y., Edwards, H., Baker, B., Lee, T., Leike, J., Schulman, J., Sutskever, I., and Cobbe, K.
\newblock Let's verify step by step.
\newblock \emph{arXiv preprint arXiv:2305.20050}, 2023.

\bibitem[Luo et~al.(2024)Luo, Liu, Liu, Phatale, Guo, Lara, Li, Shu, Zhu, Meng, Sun, and Rastogi]{luo2024improvemathematicalreasoninglanguage}
Luo, L., Liu, Y., Liu, R., Phatale, S., Guo, M., Lara, H., Li, Y., Shu, L., Zhu, Y., Meng, L., Sun, J., and Rastogi, A.
\newblock Improve mathematical reasoning in language models by automated process supervision, 2024.
\newblock URL \url{https://arxiv.org/abs/2406.06592}.

\bibitem[Miao et~al.(2024)Miao, Zhang, Li, Tang, Li, Shi, Xiao, and Zhuang]{miao2024radarrobusttwostagemodalityincomplete}
Miao, B., Zhang, W., Li, J., Tang, S., Li, Z., Shi, H., Xiao, J., and Zhuang, Y.
\newblock Radar: Robust two-stage modality-incomplete industrial anomaly detection, 2024.
\newblock URL \url{https://arxiv.org/abs/2410.01737}.

\bibitem[Pan et~al.(2024{\natexlab{a}})Pan, Fan, Li, Yu, Fei, Tang, Hong, Zhang, and Sun]{pan2024towards}
Pan, K., Fan, Z., Li, J., Yu, Q., Fei, H., Tang, S., Hong, R., Zhang, H., and Sun, Q.
\newblock Towards unified multimodal editing with enhanced knowledge collaboration.
\newblock \emph{Advances in Neural Information Processing Systems}, 37:\penalty0 110290--110314, 2024{\natexlab{a}}.

\bibitem[Pan et~al.(2024{\natexlab{b}})Pan, Tang, Li, Fan, Chow, Yan, Chua, Zhuang, and Zhang]{pan2024auto}
Pan, K., Tang, S., Li, J., Fan, Z., Chow, W., Yan, S., Chua, T.-S., Zhuang, Y., and Zhang, H.
\newblock Auto-encoding morph-tokens for multimodal llm.
\newblock \emph{arXiv preprint arXiv:2405.01926}, 2024{\natexlab{b}}.

\bibitem[Pan et~al.(2025)Pan, Lin, Yue, Ao, Jia, Zhao, Li, Tang, and Zhang]{pan2025generative}
Pan, K., Lin, W., Yue, Z., Ao, T., Jia, L., Zhao, W., Li, J., Tang, S., and Zhang, H.
\newblock Generative multimodal pretraining with discrete diffusion timestep tokens.
\newblock \emph{arXiv preprint arXiv:2504.14666}, 2025.

\bibitem[Rafailov et~al.(2023)Rafailov, Sharma, Mitchell, Manning, Ermon, and Finn]{NEURIPS2023_a85b405e}
Rafailov, R., Sharma, A., Mitchell, E., Manning, C.~D., Ermon, S., and Finn, C.
\newblock Direct preference optimization: Your language model is secretly a reward model.
\newblock In Oh, A., Naumann, T., Globerson, A., Saenko, K., Hardt, M., and Levine, S. (eds.), \emph{Advances in Neural Information Processing Systems}, volume~36, pp.\  53728--53741. Curran Associates, Inc., 2023.
\newblock URL \url{https://proceedings.neurips.cc/paper_files/paper/2023/file/a85b405ed65c6477a4fe8302b5e06ce7-Paper-Conference.pdf}.

\bibitem[Rawles et~al.(2024{\natexlab{a}})Rawles, Clinckemaillie, Chang, Waltz, Lau, Fair, Li, Bishop, Li, Campbell-Ajala, Toyama, Berry, Tyamagundlu, Lillicrap, and Riva]{rawles2024androidworlddynamicbenchmarkingenvironment}
Rawles, C., Clinckemaillie, S., Chang, Y., Waltz, J., Lau, G., Fair, M., Li, A., Bishop, W., Li, W., Campbell-Ajala, F., Toyama, D., Berry, R., Tyamagundlu, D., Lillicrap, T., and Riva, O.
\newblock Androidworld: A dynamic benchmarking environment for autonomous agents, 2024{\natexlab{a}}.
\newblock URL \url{https://arxiv.org/abs/2405.14573}.

\bibitem[Rawles et~al.(2024{\natexlab{b}})Rawles, Li, Rodriguez, Riva, and Lillicrap]{10.5555/3666122.3668731}
Rawles, C., Li, A., Rodriguez, D., Riva, O., and Lillicrap, T.
\newblock Android in the wild: a large-scale dataset for android device control.
\newblock In \emph{Proceedings of the 37th International Conference on Neural Information Processing Systems}, NIPS '23, Red Hook, NY, USA, 2024{\natexlab{b}}. Curran Associates Inc.

\bibitem[Shen et~al.(2023)Shen, Song, Tan, Li, Lu, and Zhuang]{DBLP:conf/nips/0001ST00Z23}
Shen, Y., Song, K., Tan, X., Li, D., Lu, W., and Zhuang, Y.
\newblock Hugginggpt: Solving {AI} tasks with chatgpt and its friends in hugging face.
\newblock In Oh, A., Naumann, T., Globerson, A., Saenko, K., Hardt, M., and Levine, S. (eds.), \emph{Advances in Neural Information Processing Systems 36: Annual Conference on Neural Information Processing Systems 2023, NeurIPS 2023, New Orleans, LA, USA, December 10 - 16, 2023}, 2023.
\newblock URL \url{http://papers.nips.cc/paper\_files/paper/2023/hash/77c33e6a367922d003ff102ffb92b658-Abstract-Conference.html}.

\bibitem[Snell et~al.(2024)Snell, Lee, Xu, and Kumar]{snell2024scalingllmtesttimecompute}
Snell, C., Lee, J., Xu, K., and Kumar, A.
\newblock Scaling llm test-time compute optimally can be more effective than scaling model parameters, 2024.
\newblock URL \url{https://arxiv.org/abs/2408.03314}.

\bibitem[Song et~al.(2024)Song, Yin, Yue, Huang, Li, and Lin]{song-etal-2024-trial}
Song, Y., Yin, D., Yue, X., Huang, J., Li, S., and Lin, B.~Y.
\newblock Trial and error: Exploration-based trajectory optimization of {LLM} agents.
\newblock In Ku, L.-W., Martins, A., and Srikumar, V. (eds.), \emph{Proceedings of the 62nd Annual Meeting of the Association for Computational Linguistics (Volume 1: Long Papers)}, pp.\  7584--7600, Bangkok, Thailand, August 2024. Association for Computational Linguistics.
\newblock \doi{10.18653/v1/2024.acl-long.409}.
\newblock URL \url{https://aclanthology.org/2024.acl-long.409/}.

\bibitem[Uesato et~al.(2022)Uesato, Kushman, Kumar, Song, Siegel, Wang, Creswell, Irving, and Higgins]{uesato2022solvingmathwordproblems}
Uesato, J., Kushman, N., Kumar, R., Song, F., Siegel, N., Wang, L., Creswell, A., Irving, G., and Higgins, I.
\newblock Solving math word problems with process- and outcome-based feedback, 2022.
\newblock URL \url{https://arxiv.org/abs/2211.14275}.

\bibitem[Wang et~al.(2024{\natexlab{a}})Wang, Xiong, Xie, Zhao, and Zhang]{ArmoRM}
Wang, H., Xiong, W., Xie, T., Zhao, H., and Zhang, T.
\newblock Interpretable preferences via multi-objective reward modeling and mixture-of-experts.
\newblock In \emph{The 2024 Conference on Empirical Methods in Natural Language Processing}, 2024{\natexlab{a}}.

\bibitem[Wang et~al.(2023)Wang, Ma, Feng, Zhang, Yang, Zhang, Chen, Tang, Chen, Lin, Zhao, Wei, and Wen]{wang2023survey}
Wang, L., Ma, C., Feng, X., Zhang, Z., Yang, H., Zhang, J., Chen, Z., Tang, J., Chen, X., Lin, Y., Zhao, W.~X., Wei, Z., and Wen, J.-R.
\newblock A survey on large language model based autonomous agents, 2023.

\bibitem[Wang et~al.(2024{\natexlab{b}})Wang, Bai, Tan, Wang, Fan, Bai, Chen, Liu, Wang, Ge, Fan, Dang, Du, Ren, Men, Liu, Zhou, Zhou, and Lin]{Qwen2VL}
Wang, P., Bai, S., Tan, S., Wang, S., Fan, Z., Bai, J., Chen, K., Liu, X., Wang, J., Ge, W., Fan, Y., Dang, K., Du, M., Ren, X., Men, R., Liu, D., Zhou, C., Zhou, J., and Lin, J.
\newblock Qwen2-vl: Enhancing vision-language model's perception of the world at any resolution.
\newblock \emph{arXiv preprint arXiv:2409.12191}, 2024{\natexlab{b}}.

\bibitem[Wang et~al.(2024{\natexlab{c}})Wang, Li, Wu, Luo, Hou, Yu, and Shang]{wang-etal-2024-multi-step}
Wang, Z., Li, Y., Wu, Y., Luo, L., Hou, L., Yu, H., and Shang, J.
\newblock Multi-step problem solving through a verifier: An empirical analysis on model-induced process supervision.
\newblock In Al-Onaizan, Y., Bansal, M., and Chen, Y.-N. (eds.), \emph{Findings of the Association for Computational Linguistics: EMNLP 2024}, pp.\  7309--7319, Miami, Florida, USA, November 2024{\natexlab{c}}. Association for Computational Linguistics.
\newblock \doi{10.18653/v1/2024.findings-emnlp.429}.
\newblock URL \url{https://aclanthology.org/2024.findings-emnlp.429/}.

\bibitem[Wu et~al.(2024{\natexlab{a}})Wu, Fei, Qu, Ji, and Chua]{wu24next}
Wu, S., Fei, H., Qu, L., Ji, W., and Chua, T.-S.
\newblock {NE}x{T}-{GPT}: Any-to-any multimodal {LLM}.
\newblock In \emph{Proceedings of the International Conference on Machine Learning}, pp.\  53366--53397, 2024{\natexlab{a}}.

\bibitem[Wu et~al.(2024{\natexlab{b}})Wu, Peng, Du, Zheng, Liu, Wu, Ma, Li, Yang, Zhou, et~al.]{wu2024comparative}
Wu, S., Peng, Z., Du, X., Zheng, T., Liu, M., Wu, J., Ma, J., Li, Y., Yang, J., Zhou, W., et~al.
\newblock A comparative study on reasoning patterns of openai's o1 model.
\newblock \emph{arXiv preprint arXiv:2410.13639}, 2024{\natexlab{b}}.

\bibitem[Wu et~al.(2024{\natexlab{c}})Wu, Wu, Xu, Wang, Sun, Jia, Cheng, Ding, Chen, Liang, et~al.]{wu2024atlas}
Wu, Z., Wu, Z., Xu, F., Wang, Y., Sun, Q., Jia, C., Cheng, K., Ding, Z., Chen, L., Liang, P.~P., et~al.
\newblock Os-atlas: A foundation action model for generalist gui agents.
\newblock \emph{arXiv preprint arXiv:2410.23218}, 2024{\natexlab{c}}.

\bibitem[Xi et~al.(2024)Xi, Ding, Chen, Hong, Guo, Wang, Yang, Liao, Guo, He, Gao, Chen, Zheng, Zou, Gui, Zhang, Qiu, Huang, Wu, and Jiang]{xi2024agentgym}
Xi, Z., Ding, Y., Chen, W., Hong, B., Guo, H., Wang, J., Yang, D., Liao, C., Guo, X., He, W., Gao, S., Chen, L., Zheng, R., Zou, Y., Gui, T., Zhang, Q., Qiu, X., Huang, X., Wu, Z., and Jiang, Y.-G.
\newblock Agentgym: Evolving large language model-based agents across diverse environments, 2024.

\bibitem[Xie et~al.(2024)Xie, Zhang, Chen, Li, Zhao, Cao, Hua, Cheng, Shin, Lei, Liu, Xu, Zhou, Savarese, Xiong, Zhong, and Yu]{OSWorld}
Xie, T., Zhang, D., Chen, J., Li, X., Zhao, S., Cao, R., Hua, T.~J., Cheng, Z., Shin, D., Lei, F., Liu, Y., Xu, Y., Zhou, S., Savarese, S., Xiong, C., Zhong, V., and Yu, T.
\newblock Osworld: Benchmarking multimodal agents for open-ended tasks in real computer environments, 2024.

\bibitem[Xu et~al.(2021)Xu, Masling, Du, Campagna, Heck, Landay, and Lam]{xu-etal-2021-grounding}
Xu, N., Masling, S., Du, M., Campagna, G., Heck, L., Landay, J., and Lam, M.
\newblock Grounding open-domain instructions to automate web support tasks.
\newblock In Toutanova, K., Rumshisky, A., Zettlemoyer, L., Hakkani-Tur, D., Beltagy, I., Bethard, S., Cotterell, R., Chakraborty, T., and Zhou, Y. (eds.), \emph{Proceedings of the 2021 Conference of the North American Chapter of the Association for Computational Linguistics: Human Language Technologies}, pp.\  1022--1032, Online, June 2021. Association for Computational Linguistics.
\newblock \doi{10.18653/v1/2021.naacl-main.80}.
\newblock URL \url{https://aclanthology.org/2021.naacl-main.80}.

\bibitem[Yan et~al.(2023)Yan, Yang, Zhu, Lin, Li, Wang, Yang, Zhong, McAuley, Gao, et~al.]{yan2023gpt}
Yan, A., Yang, Z., Zhu, W., Lin, K., Li, L., Wang, J., Yang, J., Zhong, Y., McAuley, J., Gao, J., et~al.
\newblock Gpt-4v in wonderland: Large multimodal models for zero-shot smartphone gui navigation.
\newblock \emph{arXiv preprint arXiv:2311.07562}, 2023.

\bibitem[Yu et~al.(2024{\natexlab{a}})Yu, Gao, and Wang]{yu-etal-2024-ovm}
Yu, F., Gao, A., and Wang, B.
\newblock {OVM}, outcome-supervised value models for planning in mathematical reasoning.
\newblock In Duh, K., Gomez, H., and Bethard, S. (eds.), \emph{Findings of the Association for Computational Linguistics: NAACL 2024}, pp.\  858--875, Mexico City, Mexico, June 2024{\natexlab{a}}. Association for Computational Linguistics.
\newblock \doi{10.18653/v1/2024.findings-naacl.55}.
\newblock URL \url{https://aclanthology.org/2024.findings-naacl.55/}.

\bibitem[Yu et~al.(2023{\natexlab{a}})Yu, Jiang, Shi, Yu, Liu, Zhang, Kwok, Li, Weller, and Liu]{yu2023metamath}
Yu, L., Jiang, W., Shi, H., Yu, J., Liu, Z., Zhang, Y., Kwok, J.~T., Li, Z., Weller, A., and Liu, W.
\newblock Metamath: Bootstrap your own mathematical questions for large language models.
\newblock \emph{arXiv preprint arXiv:2309.12284}, 2023{\natexlab{a}}.

\bibitem[Yu et~al.(2023{\natexlab{b}})Yu, Li, Wu, Tang, Ji, and Zhuang]{Yu_2023_ICCV}
Yu, Q., Li, J., Wu, Y., Tang, S., Ji, W., and Zhuang, Y.
\newblock Visually-prompted language model for fine-grained scene graph generation in an open world.
\newblock In \emph{Proceedings of the IEEE/CVF International Conference on Computer Vision (ICCV)}, pp.\  21560--21571, October 2023{\natexlab{b}}.

\bibitem[Yu et~al.(2024{\natexlab{b}})Yu, Li, Wei, Pang, Ye, Qin, Tang, Tian, and Zhuang]{Yu_2024_CVPR}
Yu, Q., Li, J., Wei, L., Pang, L., Ye, W., Qin, B., Tang, S., Tian, Q., and Zhuang, Y.
\newblock Hallucidoctor: Mitigating hallucinatory toxicity in visual instruction data.
\newblock In \emph{Proceedings of the IEEE/CVF Conference on Computer Vision and Pattern Recognition (CVPR)}, pp.\  12944--12953, June 2024{\natexlab{b}}.

\bibitem[Yuan et~al.(2023)Yuan, Yuan, Li, Dong, Lu, Tan, Zhou, and Zhou]{yuan2023scaling}
Yuan, Z., Yuan, H., Li, C., Dong, G., Lu, K., Tan, C., Zhou, C., and Zhou, J.
\newblock Scaling relationship on learning mathematical reasoning with large language models, 2023.

\bibitem[Zang et~al.(2025)Zang, Dong, Zhang, Cao, Liu, Ding, Wu, Ma, Duan, Zhang, Chen, Lin, and Wang]{zang2025internlmxcomposer25rewardsimpleeffectivemultimodal}
Zang, Y., Dong, X., Zhang, P., Cao, Y., Liu, Z., Ding, S., Wu, S., Ma, Y., Duan, H., Zhang, W., Chen, K., Lin, D., and Wang, J.
\newblock Internlm-xcomposer2.5-reward: A simple yet effective multi-modal reward model, 2025.
\newblock URL \url{https://arxiv.org/abs/2501.12368}.

\bibitem[Zelikman et~al.(2022)Zelikman, Wu, Mu, and Goodman]{NEURIPS2022_639a9a17}
Zelikman, E., Wu, Y., Mu, J., and Goodman, N.
\newblock Star: Bootstrapping reasoning with reasoning.
\newblock In Koyejo, S., Mohamed, S., Agarwal, A., Belgrave, D., Cho, K., and Oh, A. (eds.), \emph{Advances in Neural Information Processing Systems}, volume~35, pp.\  15476--15488, 2022.
\newblock URL \url{https://proceedings.neurips.cc/paper_files/paper/2022/file/639a9a172c044fbb64175b5fad42e9a5-Paper-Conference.pdf}.

\bibitem[Zhai et~al.(2024)Zhai, Yang, Xu, Dawei, Yang, Ding, and Wang]{zhai2024enhancingdecisionmakingllmagents}
Zhai, Y., Yang, T., Xu, K., Dawei, F., Yang, C., Ding, B., and Wang, H.
\newblock Enhancing decision-making for llm agents via step-level q-value models, 2024.
\newblock URL \url{https://arxiv.org/abs/2409.09345}.

\bibitem[Zhang et~al.(2024{\natexlab{a}})Zhang, Li, He, Zhang, Qiao, Qin, Ma, Kang, Lin, Rajmohan, Zhang, and Zhang]{zhang2024ufouifocusedagentwindows}
Zhang, C., Li, L., He, S., Zhang, X., Qiao, B., Qin, S., Ma, M., Kang, Y., Lin, Q., Rajmohan, S., Zhang, D., and Zhang, Q.
\newblock Ufo: A ui-focused agent for windows os interaction, 2024{\natexlab{a}}.
\newblock URL \url{https://arxiv.org/abs/2402.07939}.

\bibitem[Zhang et~al.(2024{\natexlab{b}})Zhang, Zhoubian, Hu, Yue, Dong, and Tang]{zhang2024rest}
Zhang, D., Zhoubian, S., Hu, Z., Yue, Y., Dong, Y., and Tang, J.
\newblock Rest-mcts*: Llm self-training via process reward guided tree search.
\newblock \emph{arXiv preprint arXiv:2406.03816}, 2024{\natexlab{b}}.

\bibitem[Zhang et~al.(2024{\natexlab{c}})Zhang, Du, Pang, Liu, Gao, and Lin]{zhang2024chain}
Zhang, X., Du, C., Pang, T., Liu, Q., Gao, W., and Lin, M.
\newblock Chain of preference optimization: Improving chain-of-thought reasoning in llms.
\newblock \emph{arXiv preprint arXiv:2406.09136}, 2024{\natexlab{c}}.

\bibitem[Zhou et~al.(2024{\natexlab{a}})Zhou, Yang, Wen, Wen, Wang, Xi, Xu, Yu, and Zhang]{zhou2024trad}
Zhou, R., Yang, Y., Wen, M., Wen, Y., Wang, W., Xi, C., Xu, G., Yu, Y., and Zhang, W.
\newblock {TRAD}: Enhancing llm agents with step-wise thought retrieval and aligned decision.
\newblock In \emph{Proceedings of the 47th International {ACM} {SIGIR} Conference on Research and Development in Information Retrieval ({SIGIR})}, 2024{\natexlab{a}}.

\bibitem[Zhou et~al.(2024{\natexlab{b}})Zhou, Xu, Zhu, Zhou, Lo, Sridhar, Cheng, Bisk, Fried, Alon, et~al.]{zhou2024webarena}
Zhou, S., Xu, F.~F., Zhu, H., Zhou, X., Lo, R., Sridhar, A., Cheng, X., Bisk, Y., Fried, D., Alon, U., et~al.
\newblock Webarena: A realistic web environment for building autonomous agents.
\newblock \emph{ICLR}, 2024{\natexlab{b}}.

\end{thebibliography}
\bibliographystyle{icml2025}

\clearpage
\newpage
\appendix

\lstset{
    literate={“}{{\textquotedblleft}}1
             {”}{{\textquotedblright}}1
             {‘}{{\textquoteleft}}1
             {’}{{\textquoteright}}1,
    breaklines=true, 
}

\title{Appendix}

\author{}
\date{}

\maketitle

This is the Appendix for the paper ``Boosting Virtual Agent Learning and Reasoning: A Step-wise, Multi-dimensional, and Generalist Reward Model with Benchmark''.

\section*{Overview}
In this supplementary material we present:
\begin{itemize}[leftmargin=5pt]
    \vspace{-5pt}
    \item Detailed prompts for testing general MLLMs on \textbf{\textit{SRMEval}} and for building \textbf{\textit{SRM}} are provided in Section~\ref{sec:prompt}.
    \item Pseudocode for the \textbf{\texttt{Similar}} pipeline, including key algorithmic steps, is described in Section~\ref{sec:pseudocode_of_method}.
    \item Human evaluation details and human acceptance of 5-dimension data from \textbf{\textit{SRM}} are illustrated in Section~\ref{sec:human_acceptance_of_SRM}.
    \item Case studies demonstrating the applications of \textbf{\texttt{Similar}} are presented in Section~\ref{sec:case_study_of_similar}.
    \item Comprehensive ablation experiments are conducted and analyzed in Section~\ref{sec:ablation_more}.
    \item Additional visualizations of \textbf{\textit{SRMEval}}, showcasing task performance, are provided in Section~\ref{sec:visualization_of_srmeval}.
\end{itemize}

\section{Detailed Prompt Design}
\label{sec:prompt}

This section provides a comprehensive overview of the prompt designs used in our experiments. We detail the prompts for testing general MLLMs on \textbf{\textit{SRMEval}} and the prompts for building the \textbf{\textit{SRM}} model, highlighting their structure and purpose.

\vspace{-10pt}
\subsection{Prompt for testing general MLLMs on \textbf{\textit{SRMEval}}}

This subsection describes the prompts used to evaluate general MLLMs on the \textbf{\textit{SRMEval}} benchmark.

\begin{tcolorbox}[
    colback=gray!10!white,
    colframe=gray!50!black,
    title=Prompt for \textbf{\textit{SRMEval}} (main part),
    fonttitle=\bfseries,
    boxrule=0.5mm,
    arc=2mm,
    width=\columnwidth,
    breakable,
    before skip=2mm,
    after skip=2mm,
    left=3pt,
    right=3pt,
    top=3pt,
    bottom=3pt
]
{\ttfamily\small
You are an expert in evaluating the performance of a Virtual Agent. 

\vspace{2mm}
The Virtual Agent is designed to help a human user complete specified tasks (such as app usage, web navigation, web content Q\&A, etc.) on various platform applications (such as websites, mobile devices, operation systems, etc.) based on given instructions. Given the user's INSTRUCTION, the OBSERVATION of current platforms, the action TRAJECTORY of the agent, the two ACTION\_X and ACTION\_Y predicted by the agent, and the current action step number STEP\_IDX. Your \textbf{GOAL} is to help me complete step-wise evaluation, that is, evaluate the quality of the Agent's \textbf{ACTION} in a specific dimension. Choose the better action (ACTION\_X or ACTION\_Y) based on the given $\langle$\textbf{EVALUATION DIMENSION}$\rangle$. Output ``Y'' and the reason if ACTION\_X is better, or ``X'' and the reason if ACTION\_Y is better. Do not output responses like ``two actions are similar''.


\vspace{2mm}
\textbf{$\langle$Word Meaning$\rangle$}\\
1.\textbf{INSTRUCTION}: refers to the command of human users to the Agent, which is the specific content that the Agent needs to complete the task on a specific platform, that is, the ultimate \textbf{GOAL} of the Agent.\\
2.\textbf{OBSERVATION}: refers to the specific information of the current platform that an agent can observe on the platform where the task needs to be completed, which is the environment in which the agent is currently located. In our task, observations are presented in the form of images, known as screenshots.\\
3.\textbf{TRAJECTORY}: refers to the action prediction made by an agent in the past to complete the \textbf{INSTRUCTION}, which records all actions taken by the agent from the first step to the current step. If this is the first step, then the trajectory is empty.\\
4.\textbf{ACTION}: refers to the predicted operation of the Agent in the current state to complete the \textbf{INSTRUCTION} in the current step. This operation generally refers to a simple action command, such as ``\textbf{CLICK}'', ``\textbf{TYPE}'', etc. Note that \textbf{ACTION} is the result predicted by the agent after observing the current \textbf{OBSERVATION}, and the Agent often cannot complete the task in one step.\\
5.\textbf{STEP\_IDX}: refers to the sequence number of the Agent executing the current \textbf{ACTION} to complete the \textbf{INSTRUCTION}.
}
\end{tcolorbox}

Here is the evaluation dimension part of the prompts.

\textit{Helpfulness}.

\begin{tcolorbox}[
    colback=gray!10!white,
    colframe=gray!50!black,
    title=Prompt for \textbf{\textit{SRMEval}} (Helpfulness),
    fonttitle=\bfseries,
    boxrule=0.5mm,
    arc=2mm,
    width=\columnwidth,
    breakable,
    before skip=2mm,
    after skip=2mm,
    left=3pt,
    right=3pt,
    top=3pt,
    bottom=3pt
]
{\ttfamily\small
1.\textbf{[HELPFULNESS]}\\
1.1 Meaning: It indicates the degree to which this step contributes to the completion of the final task. There are good and bad contributions, the correct steps will give a positive contribution, and the wrong steps will give a negative contribution.\\
1.2 Design motivation: Different steps contribute differently to the completion of the final task, with good steps helping to accomplish the task and bad steps hindering it. Good steps should be rewarded positively, while bad steps should be punished negatively. If each step is correct and the total number of steps is 5, then the contribution of each step can be considered as 1/5, meaning that each step completes 1/5 of the final task. If 4 more steps are needed from the current step and the current step is incorrect, then the contribution of the current step is -1/4, indicating that it hinders 1/4 of the final task progress.
}
\end{tcolorbox}

\textit{Odds of Success}.

\begin{tcolorbox}[
    colback=gray!10!white,
    colframe=gray!50!black,
    title=Prompt for \textbf{\textit{SRMEval}} (Odds of Success),
    fonttitle=\bfseries,
    boxrule=0.5mm,
    arc=2mm,
    width=\columnwidth,
    breakable,
    before skip=2mm,
    after skip=2mm,
    left=3pt,
    right=3pt,
    top=3pt,
    bottom=3pt
]
{\ttfamily\small
2.\textbf{[ODDS OF SUCCESS]}\\
2.1 Meaning: It indicates the potential of the step to complete the task, which is the probability of a step reaching the completion of the task.\
2.2 Design motivation: The more correct steps lead to a higher probability of success in the final task, and the more incorrect steps lead to a higher probability of failure in the final task. Different steps have different potential to complete the task. If one step of the agent is to follow the Instructions to complete the task, then this step generally has high potential. We can derive the probability of a step leading to success from the N paths generated by that step, which serves as the potential for that step to complete the task which is crucial for evaluating.
}
\end{tcolorbox}

\textit{Efficiency}.

\begin{tcolorbox}[
    colback=gray!10!white,
    colframe=gray!50!black,
    title=Prompt for \textbf{\textit{SRMEval}} (Efficiency),
    fonttitle=\bfseries,
    boxrule=0.5mm,
    arc=2mm,
    width=\columnwidth,
    breakable,
    before skip=2mm,
    after skip=2mm,
    left=3pt,
    right=3pt,
    top=3pt,
    bottom=3pt
]
{\ttfamily\small
3.\textbf{[EFFICIENCY]}\\
3.1 Meaning: It indicates whether this step is efficient in completing the task. We calculate this metric as the difference between `the number of steps required to complete the final task after the current step' and `the number of steps required to complete the final task after the previous step', divided by `the total number of steps required to complete the task'. This indicates the degree of efficiency improvement in completing tasks after the current step is executed.\\
3.2 Design motivation: A basic assumption is that the fewer steps the Agent operates, the more efficient it is, because the consumption of these paths (time consumption, hardware consumption) can be considered to be the least and the efficiency is the highest. Therefore, if the operation of a step can reduce the number of steps required to complete the task as a whole, then it can be considered that the operation of this step is very efficient. For example, after the previous step, it takes 7 steps to complete the task, but after the current step, it only takes 4 steps to complete the task. The difference of 7-4=3 is the efficiency improvement of the current step in completing the final task.
}
\end{tcolorbox}


\textit{Task Relevance}.

\begin{tcolorbox}[
    colback=gray!10!white,
    colframe=gray!50!black,
    title=Prompt for \textbf{\textit{SRMEval}} (Task Relevance),
    fonttitle=\bfseries,
    boxrule=0.5mm,
    arc=2mm,
    width=\columnwidth,
    breakable,
    before skip=2mm,
    after skip=2mm,
    left=3pt,
    right=3pt,
    top=3pt,
    bottom=3pt
]
{\ttfamily\small
4.\textbf{[TASK RELEVANCE]}\\
4.1 Meaning: It indicates is whether the operation of the Agent is related to achieving the \textbf{INSTRUCTION}.\\
4.2 Design motivation: Some operational steps may prevent the task from being completed, but they are related to the task (for example, we need to ask the agent to take notes, and the agent takes notes, which is related to the task, but the recorded note content is incorrect, indicating that this is an incorrect step). Some operational steps may be meaningless, but they can still lead to task completion (such as clicking on a blank screen without generating any response, which is unrelated to the task, but the agent's subsequent actions can still result in task success). Therefore, an indicator is needed to identify whether the current step of operation is related to the task.\\
4.3 Range of values after mapping: \{0, 1\}. The larger the value, the greater the correlation between the step and the task.
}
\end{tcolorbox}


\textit{Coherence}.

\begin{tcolorbox}[
    colback=gray!10!white,
    colframe=gray!50!black,
    title=Prompt for \textbf{\textit{SRMEval}} (Coherence),
    fonttitle=\bfseries,
    boxrule=0.5mm,
    arc=2mm,
    width=\columnwidth,
    breakable,
    before skip=2mm,
    after skip=2mm,
    left=3pt,
    right=3pt,
    top=3pt,
    bottom=3pt
]
{\ttfamily\small
5.\textbf{[COHERENCE]}\\
5.1 Meaning: It represents the compactness and coherence between the current step and the previous step.\\
5.2 Design motivation: Some operations, although task-related, not inefficient, and highly likely to lead to success, lack coherence with the previous step. For example, the task is to ``query the Lakers' game results and record them in the Note''. The Agent operations are as follows: a Open the browser; b. Open Note; c. Create new notes; d. Search for Lakers games; e. Query the results of the competition; f. Record the results of the competition in your notes. It can be found that the operations of a and b lack coherence, and it is more in line with human preferences to directly search for competition results after opening the browser instead of simultaneously opening Note.\\
5.3 Range of values after mapping: \{0, 1\}. The larger the value, the greater the coherence of the step.
}
\end{tcolorbox}


Total dimension and Trajectory-level dimension.

\begin{tcolorbox}[
    colback=gray!10!white,
    colframe=gray!50!black,
    title=Prompt for \textbf{\textit{SRMEval}} (Total and Trajectory-level),
    fonttitle=\bfseries,
    boxrule=0.5mm,
    arc=2mm,
    width=\columnwidth,
    breakable,
    before skip=2mm,
    after skip=2mm,
    left=3pt,
    right=3pt,
    top=3pt,
    bottom=3pt
]
{\ttfamily\small
6.\textbf{[TOTAL]}\\
Meaning: Integrated decision-making based on the 5 dimensions mentioned earlier.\\

7.\textbf{[TRAJECTORY]}\\
Meaning: Represents the quality of the entire trajectory, which can be expressed as the average total score of all steps in the trajectory.
}
\end{tcolorbox}

\subsection{Prompt for building \textbf{\textit{SRM}}}

This subsection outlines the prompts designed for constructing the \textbf{\textit{SRM}} model.

\begin{tcolorbox}[
    colback=gray!10!white,
    colframe=gray!50!black,
    title=Prompt for building \textbf{\textit{SRM}},
    fonttitle=\bfseries,
    boxrule=0.5mm,
    arc=2mm,
    width=\columnwidth,
    breakable,
    before skip=2mm,
    after skip=2mm,
    left=3pt,
    right=3pt,
    top=3pt,
    bottom=3pt
]
{\ttfamily\small
You are a virtual agent.\\

The Virtual Agent is designed to help a human user complete specified tasks (such as app usage, web navigation, web content Q\&A, etc.) on various platform applications (such as websites, mobile devices, operation systems, etc.) based on given instructions.

\vspace{2mm}
You will predict the next action based on the following content [\textbf{INSTRUCTION}], [\textbf{OBSERVATION}], [\textbf{REASON\_STEPS}]:

\vspace{2mm}
1.\textbf{[INSTRUCTION]}: It is your ultimate \textbf{GOAL}, and all your actions are aimed at completing this task.

2.\textbf{[OBSERVATION]}: It is an observation of an image, which is the screenshot of the platform (such as a computer screen).

3.\textbf{[REASON\_STEPS]}: They are the trajectory of the actions you performed in the past to complete the instruction, from which you can understand how you thought in order to complete the instruction. If it is empty, it means it is currently the first step.
}
\end{tcolorbox}

\section{\textbf{\texttt{Similar}} Pipeline Pseudocode}
\label{sec:pseudocode_of_method}

The rapid advancement of MLLMs~\cite{li2020unsupervised, li2022fine, Yu_2023_ICCV, Yu_2024_CVPR, pan2024towards, wu24next, fei2024vitron, miao2024radarrobusttwostagemodalityincomplete}, which excel at integrating text, vision, and other modalities, has enabled the development of GVAs~\cite{gao2024generalistvirtualagentssurvey, zhang2024ufouifocusedagentwindows, DBLP:conf/nips/0001ST00Z23}, which also inspires us to study Reward Models for GVAs with its Benchmark.

In this section, we present the pipeline pseudocode for training our proposed \textbf{\texttt{Similar}} model. The pipeline consists of three main components: 1) a five-dimensional process supervision framework to evaluate agent steps, 2) an automatic generalist dataset collecting process, and 3) a Triple-M strategy for reward model training.

\vspace{-3pt}

\subsection{Five-Dimensional Process Supervision Framework}

The five-dimensional process supervision framework systematically evaluates the quality of an agent's steps using five distinct dimensions: \textit{Helpfulness (H)}, \textit{Odds of Success (OS)}, \textit{Efficiency (E)}, \textit{Task Relevance (TR)}, and \textit{Coherence (C)}. The following pseudocode outlines the computation of these dimensions for a given step \( S_i \), providing a comprehensive assessment of step quality.

\vspace{-8pt}

\subsection{Automatic Generalist Dataset Collecting}

\vspace{-3pt}
The automatic generalist dataset collecting process leverages the MCTS-P algorithm to collect annotated step-wise data across multiple platforms, including Web, Android, Linux, and Windows.

\vspace{-8pt}

\subsection{Triple-M Strategy for Reward Model Training}

\vspace{-3pt}
The Triple-M strategy integrates multi-step, multi-dimensional, and multi-modal data for reward model training, ensuring high flexibility, robustness, and adaptability. The following pseudocode outlines the two-stage training process, which includes regression layer optimization and dynamic gating network adjustment.

\setlength{\textfloatsep}{10pt}
\begin{algorithm}[!t]
   \caption{Five-Dimensional Process Supervision}
   \label{alg:five-dim}
   \scriptsize
\begin{algorithmic}[1]
   \STATE \textbf{Input:} Step \( S_i \), ground truth \( a^* \), number of simulations \( N \)
   \STATE \textbf{Output:} Five-dimensional scores \( (H_i, OS_i, E_i, TR_i, C_i) \)

   \STATE \textbf{Compute Helpfulness (H):}
   \STATE \( H_i = \frac{1 - AC_{i-1}}{M - i + 1} (1 - 2r_i) \)
   \STATE \( AC_i = \max(AC_{i-1} + H_i, 0) \)

   \STATE \textbf{Compute Odds of Success (OS):}
   \STATE \( OS_i = \frac{\sum^N_{j=1} \mathbb{I}(a_{i,j} = a^*)}{N} \)

   \STATE \textbf{Compute Efficiency (E):}
   \STATE \( E_i = \frac{Len_{i-1} - Len_i}{len_0} \)
   \STATE \( Len_i = \text{avg}(Len(S_{i,j})) \)

   \STATE \textbf{Compute Task Relevance (TR) and Coherence (C):}
   \STATE \( TR_i = \text{MLLM\_Evaluate}(S_i, \text{instruction}) \)
   \STATE \( C_i = \text{MLLM\_Evaluate}(S_i, S_{i-1}) \)

   \STATE \textbf{return} \( (H_i, OS_i, E_i, TR_i, C_i) \)
\end{algorithmic}
\end{algorithm}

\vspace{-5pt}
\begin{algorithm}[!t]
   \caption{Automatic Generalist Dataset Collecting}
   \label{alg:data-collect}
   \scriptsize
\begin{algorithmic}[1]
   \STATE \textbf{Input:} Task instruction \( q \), platforms \( \mathcal{P} = \{\text{Web, Android, Linux, Windows}\} \)
   \STATE \textbf{Output:} Annotated dataset \( \mathcal{D} \)

   \STATE Initialize empty dataset \( \mathcal{D} \)
   \FOR{each platform \( p \in \mathcal{P} \)}
      \STATE Initialize MCTS-P tree \( T_q \) for task \( q \) on platform \( p \)
      \FOR{each node \( S_{i,j} \) in \( T_q \)}
         \STATE Calculate minimum steps \( M \) to reach a correct answer
         \STATE Simulate \( N \) trajectories to compute basic reward \( r_i \)
         \STATE Compute \( H_i, OS_i, E_i \) using formulas from Algorithm~\ref{alg:five-dim}
         \STATE Evaluate \( TR_i \) and \( C_i \) using MLLM (e.g., GPT-4)
         \IF{node \( S_{i,j} \) leads to a complete trajectory}
            \STATE Verify correctness using platform-specific evaluation methods
            \STATE Add annotated step \( (S_{i,j}, H_i, OS_i, E_i, TR_i, C_i) \) to \( \mathcal{D} \)
         \ENDIF
      \ENDFOR
   \ENDFOR
   \STATE Prune incomplete branches from \( T_q \)
   \STATE \textbf{return} Annotated dataset \( \mathcal{D} \)
\end{algorithmic}
\end{algorithm}

\begin{algorithm}[!t]
   \caption{Triple-M Strategy for Reward Model Training}
   \label{alg:triple-m}
   \scriptsize
\begin{algorithmic}[1]
   \STATE \textbf{Input:} Training dataset \( \mathcal{D} \), pre-trained MLLM \( f_\theta \), gating network \( g_\phi \)
   \STATE \textbf{Output:} Trained reward model \( R \)

   \STATE \textbf{Stage 1: Regression Layer Training}
   \FOR{each batch \( (x, y, r) \in \mathcal{D} \)}
      \STATE Extract hidden state \( h = f_\theta(x \oplus y) \)
      \STATE Compute predicted scores \( \hat{r} = W^\top h \)
      \STATE Update \( \theta, W \) using \( L_{RG} = \| \hat{r} - r \|_2^2 \)
   \ENDFOR

   \STATE \textbf{Stage 2: Gating Network Training}
   \FOR{each batch \( (x, y_{\text{chosen}}, y_{\text{rejected}}) \in \mathcal{D} \)}
      \STATE Compute coefficients \( w = g_\phi(f_\theta(x)) \)
      \STATE Compute preference scores \( R_{\text{chosen}} = w^\top r_{\text{chosen}} \)
      \STATE Compute preference scores \( R_{\text{rejected}} = w^\top r_{\text{rejected}} \)
      \STATE Update \( \phi \) using \( L_{BT} = -\log \frac{\exp(R_{\text{chosen}})}{\exp(R_{\text{chosen}}) + \exp(R_{\text{rejected}})} \)
   \ENDFOR

   \STATE \textbf{return} Trained reward model \( R = g_\phi(f_\theta(x))^\top r \)
\end{algorithmic}
\end{algorithm}

\begin{table}[b]
\caption{Sample size and human acceptance rate for each dimension in \textbf{\textit{SRM}}.}
\label{tab:human_acceptance}
\vspace{-5pt}
\renewcommand{\arraystretch}{1.2}
\begin{center}
\begin{tabular}{@{}ccc@{}}
\toprule
\textbf{Dimension} & \textbf{Sample Size} & \textbf{Human Acceptance} \\
\midrule
Helpfulness & 6000 & 87.9\% \\
Odds of Success & 2000 & 78.8\% \\
Efficiency & 6000 & 82.6\% \\
Task Relevance & 1000 & 84.7\% \\
Coherence & 2000 & 93.5\% \\
\bottomrule
\end{tabular}
\end{center}
\vspace{-15pt}
\end{table}

\section{Human Evaluation Details and Human Acceptance of  \textbf{\textit{SRM}}}
\label{sec:human_acceptance_of_SRM}

To ensure high-quality annotations, we collaborated with a professional commercial data labeling team. The process included: \textbf{1) Training Phase:} Annotators underwent three rounds of iterative ``label-review-feedback'' cycles to clarify ambiguities of annotation (e.g., the complexity of UI interaction tasks). Only after achieving $>95\%$ accuracy on validation samples did formal annotation begin. \textbf{2) Formal Annotation:} Each test sample in \textbf{\textit{SRMEval}} was independently labeled by three annotators and three checkers. The final data in test set required $>99\%$ accuracy.

To validate the quality of our five-dimensional assessment data and ensure alignment with human preferences, we randomly sampled a batch of data from the \textbf{\textit{SRM}} Benchmark. The sample size varied across dimensions due to differences in score distributions. This stems from their fundamental design - for instance, \textit{Task Relevance} and \textit{Coherence} are binary values, which naturally yield fewer possible preference pairs. Human annotators were then asked to select the better action from candidate action pairs in the sampled data, based on specific evaluation types. If the annotator considers a sample correct, mark it as $1$; otherwise, mark it as $0$, and calculate Accuracy as Human Acceptance.

The results, as shown in Table~\ref{tab:human_acceptance}, demonstrate that the human acceptance rate for all five dimensions exceeds $78.8\%$, strongly indicating the superiority of our designed annotation dimensions and the high quality of the collected data.

The evaluation process was further enhanced by incorporating a rigorous double-blind annotation protocol, where neither the annotators nor the analysts were aware of the origin or automated scores of the candidate actions.

\section{Case Studies on \textbf{\texttt{Similar}} Applications}
\label{sec:case_study_of_similar}

\subsection{\textit{Helpfulness}}

\begin{tcolorbox}[
    colback=gray!10!white,
    colframe=gray!50!black,
    title=Case of \textit{Helpfulness},
    fonttitle=\bfseries,
    boxrule=0.5mm,
    arc=2mm,
    width=\columnwidth,
    breakable,
    before skip=2mm,
    after skip=2mm,
    left=3pt,
    right=3pt,
    top=3pt,
    bottom=3pt
]

\textbf{$\blacktriangleright$ Input:} 

{\ttfamily\small
\begin{lstlisting}[
    basicstyle=\ttfamily\small,
    breaklines=true,
    breakatwhitespace=true,
    xleftmargin=0pt,
    xrightmargin=0pt,
    columns=flexible,
    keepspaces=true,
    frame=none,
    resetmargins=true,
    breakindent=0pt
]
[INST]
Create a timer with 0 hours, 16 minutes, and 35 seconds. Do not start the timer.
[/INST]
\end{lstlisting}
\begin{lstlisting}[
    basicstyle=\ttfamily\small,
    breaklines=true,
    breakatwhitespace=true,
    xleftmargin=0pt,
    xrightmargin=0pt,
    columns=flexible,
    keepspaces=true,
    frame=none,
    resetmargins=true,
    breakindent=0pt
]
[OBS]
\end{lstlisting}
\begin{center}
\includegraphics[width=0.2\columnwidth]{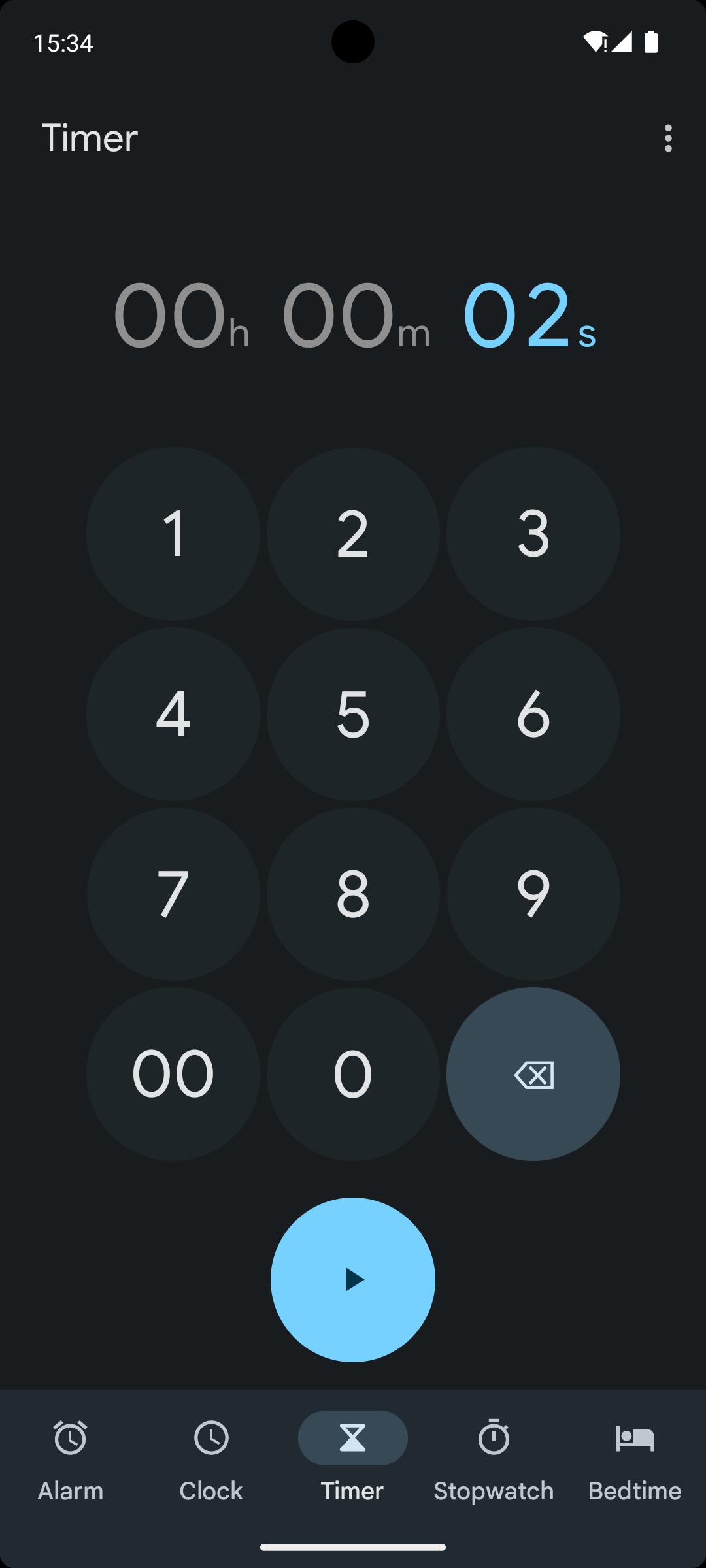}
\end{center}
\begin{lstlisting}[
    basicstyle=\ttfamily\small,
    breaklines=true,
    breakatwhitespace=true,
    xleftmargin=0pt,
    xrightmargin=0pt,
    columns=flexible,
    keepspaces=true,
    frame=none,
    resetmargins=true,
    breakindent=0pt
]
[/OBS]
\end{lstlisting}

\begin{lstlisting}[
    basicstyle=\ttfamily\small,
    breaklines=true,
    breakatwhitespace=true,
    xleftmargin=0pt,
    xrightmargin=0pt,
    columns=flexible,
    keepspaces=true,
    frame=none,
    resetmargins=true,
    breakindent=0pt
]
[TRAJ]
Step 1: Click Search.

Step 2: Click Clock.

Step 3: Click 2.
[/TRAJ]
\end{lstlisting}

\textbf{$\blacktriangleright$ Output:}

\begin{lstlisting}[
    basicstyle=\ttfamily\small,
    breaklines=true,
    breakatwhitespace=true,
    xleftmargin=0pt,
    xrightmargin=0pt,
    columns=flexible,
    keepspaces=true,
    frame=none,
    resetmargins=true,
    breakindent=0pt
]
[ACTION_X]
Click Backspace 2.
[/ACTION_X]
[SCORE_X]
H: 0.72
[/SCORE_X]

[ACTION_Y]
Click 1.
[/ACTION_Y]
[SCORE_Y]
H: -0.37
[/SCORE_Y]
\end{lstlisting}
}
\end{tcolorbox}

In this case, the task is to set a timer for $16$ minutes and $35$ seconds. According to the reasoning steps, the previous action was clicking ``2", which does not match the required time. The current step should involve deleting the incorrect input immediately. ACTION\_X (Backspace) is correct and more helpful, while ACTION\_Y (Click $1$) further hinders task completion and is clearly harmful. Therefore, ACTION\_X receives a higher \textit{Helpfulness} score.

\subsection{\textit{Odds of Success}}

\begin{tcolorbox}[
    colback=gray!10!white,
    colframe=gray!50!black,
    title=Case of \textit{Odds of Success},
    fonttitle=\bfseries,
    boxrule=0.5mm,
    arc=2mm,
    width=\columnwidth,
    breakable,
    before skip=2mm,
    after skip=2mm,
    left=3pt,
    right=3pt,
    top=3pt,
    bottom=3pt
]

\textbf{$\blacktriangleright$ Input:} 

{\ttfamily\small
\begin{lstlisting}[
    basicstyle=\ttfamily\small,
    breaklines=true,
    breakatwhitespace=true,
    xleftmargin=0pt,
    xrightmargin=0pt,
    columns=flexible,
    keepspaces=true,
    frame=none,
    resetmargins=true,
    breakindent=0pt
]
[INST]
In Simple Calendar Pro, delete all the calendar events on 2023-10-27.
[/INST]
\end{lstlisting}
\begin{lstlisting}[
    basicstyle=\ttfamily\small,
    breaklines=true,
    breakatwhitespace=true,
    xleftmargin=0pt,
    xrightmargin=0pt,
    columns=flexible,
    keepspaces=true,
    frame=none,
    resetmargins=true,
    breakindent=0pt
]
[OBS]
\end{lstlisting}
\begin{center}
\includegraphics[width=0.2\columnwidth]{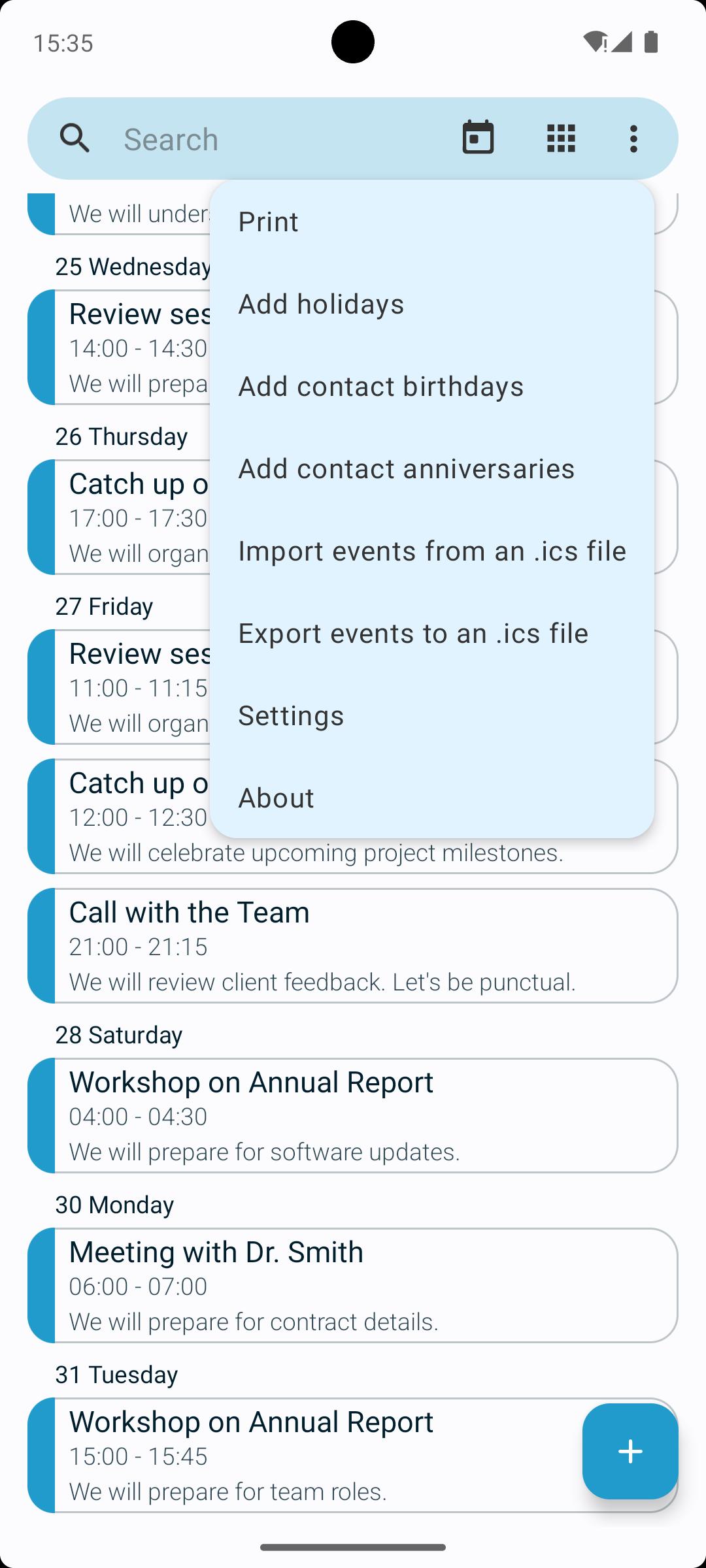}
\end{center}
\begin{lstlisting}[
    basicstyle=\ttfamily\small,
    breaklines=true,
    breakatwhitespace=true,
    xleftmargin=0pt,
    xrightmargin=0pt,
    columns=flexible,
    keepspaces=true,
    frame=none,
    resetmargins=true,
    breakindent=0pt
]
[/OBS]
\end{lstlisting}

\begin{lstlisting}[
    basicstyle=\ttfamily\small,
    breaklines=true,
    breakatwhitespace=true,
    xleftmargin=0pt,
    xrightmargin=0pt,
    columns=flexible,
    keepspaces=true,
    frame=none,
    resetmargins=true,
    breakindent=0pt
]
[TRAJ]
Step 1: Click Search.

Step 2: Click Calendar.

Step 3: Scroll down.

Step 4: Long press 27 Friday.

Step 5: Click More options.
[/TRAJ]
\end{lstlisting}

\textbf{$\blacktriangleright$ Output:}

\begin{lstlisting}[
    basicstyle=\ttfamily\small,
    breaklines=true,
    breakatwhitespace=true,
    xleftmargin=0pt,
    xrightmargin=0pt,
    columns=flexible,
    keepspaces=true,
    frame=none,
    resetmargins=true,
    breakindent=0pt
]
[ACTION_X]
Navigate back.
[/ACTION_X]
[SCORE_X]
OS: 0.75
[/SCORE_X]

[ACTION_Y]
Click Import events from a .ics file.
[/ACTION_Y]
[SCORE_Y]
OS: 0.13
[/SCORE_Y]
\end{lstlisting}
}
\end{tcolorbox}

In this case, the task is to delete events on a specific day in the Calendar. The previous step, clicking ``More Options", aimed to locate the delete button, but the current observation shows no delete option is available. Therefore, the correct action is to navigate back and search for the delete button elsewhere. ACTION\_X (Navigate back) is the appropriate choice, while ACTION\_Y (Import a file) is clearly incorrect. Thus, ACTION\_X receives a higher \textit{Odds of Success} score.

\subsection{\textit{Efficiency}}

\begin{tcolorbox}[
    colback=gray!10!white,
    colframe=gray!50!black,
    title=Case of \textit{Efficiency},
    fonttitle=\bfseries,
    boxrule=0.5mm,
    arc=2mm,
    width=\columnwidth,
    breakable,
    before skip=2mm,
    after skip=2mm,
    left=3pt,
    right=3pt,
    top=3pt,
    bottom=3pt
]

\textbf{$\blacktriangleright$ Input:} 

{\ttfamily\small
\begin{lstlisting}[
    basicstyle=\ttfamily\small,
    breaklines=true,
    breakatwhitespace=true,
    xleftmargin=0pt,
    xrightmargin=0pt,
    columns=flexible,
    keepspaces=true,
    frame=none,
    resetmargins=true,
    breakindent=0pt
]
[INST]
In Simple Calendar Pro, delete all the events.
[/INST]
\end{lstlisting}
\begin{lstlisting}[
    basicstyle=\ttfamily\small,
    breaklines=true,
    breakatwhitespace=true,
    xleftmargin=0pt,
    xrightmargin=0pt,
    columns=flexible,
    keepspaces=true,
    frame=none,
    resetmargins=true,
    breakindent=0pt
]
[OBS]
\end{lstlisting}
\begin{center}
\includegraphics[width=0.2\columnwidth]{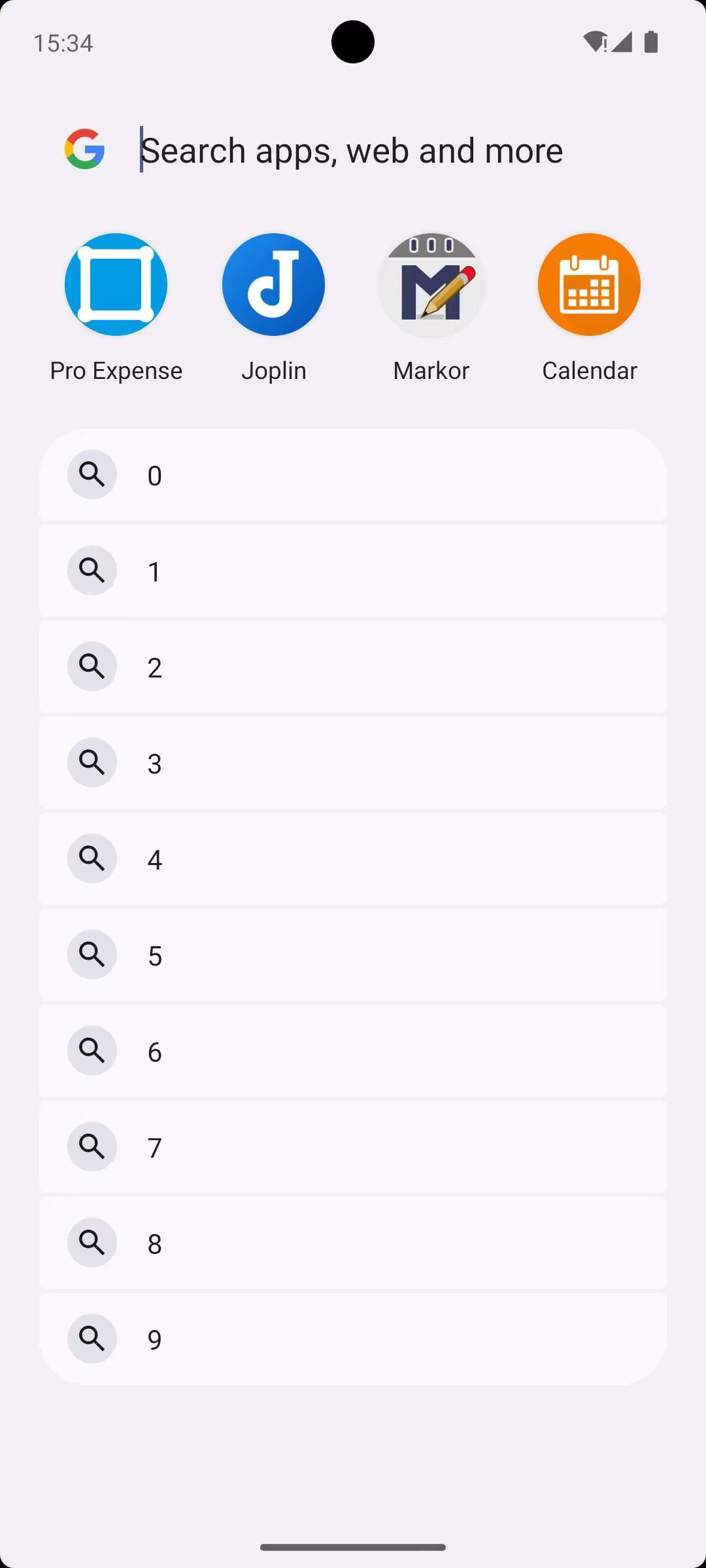}
\end{center}
\begin{lstlisting}[
    basicstyle=\ttfamily\small,
    breaklines=true,
    breakatwhitespace=true,
    xleftmargin=0pt,
    xrightmargin=0pt,
    columns=flexible,
    keepspaces=true,
    frame=none,
    resetmargins=true,
    breakindent=0pt
]
[/OBS]
\end{lstlisting}

\begin{lstlisting}[
    basicstyle=\ttfamily\small,
    breaklines=true,
    breakatwhitespace=true,
    xleftmargin=0pt,
    xrightmargin=0pt,
    columns=flexible,
    keepspaces=true,
    frame=none,
    resetmargins=true,
    breakindent=0pt
]
[TRAJ]
Step 1: Click Search.
[/TRAJ]
\end{lstlisting}

\textbf{$\blacktriangleright$ Output:}

\begin{lstlisting}[
    basicstyle=\ttfamily\small,
    breaklines=true,
    breakatwhitespace=true,
    xleftmargin=0pt,
    xrightmargin=0pt,
    columns=flexible,
    keepspaces=true,
    frame=none,
    resetmargins=true,
    breakindent=0pt
]
[ACTION_X]
Click Calender.
[/ACTION_X]
[SCORE_X]
E: 0.88
[/SCORE_X]

[ACTION_Y]
Input text ``Simple Calendar Pro''.
[/ACTION_Y]
[SCORE_Y]
E: 0.65
[/SCORE_Y]
\end{lstlisting}
}
\end{tcolorbox}

In this case, since the search interface displays past search history, including ``Calendar", ACTION\_X (Click ``Calendar") is more efficient in the \textit{Efficiency} dimension. However, ACTION\_Y is also a correct approach, so its \textit{Efficiency} score remains relatively high.

\subsection{\textit{Task Relevance}}

\begin{tcolorbox}[
    colback=gray!10!white,
    colframe=gray!50!black,
    title=Case of \textit{Task Relevance},
    fonttitle=\bfseries,
    boxrule=0.5mm,
    arc=2mm,
    width=\columnwidth,
    breakable,
    before skip=2mm,
    after skip=2mm,
    left=3pt,
    right=3pt,
    top=3pt,
    bottom=3pt
]

\textbf{$\blacktriangleright$ Input:} 

{\ttfamily\small
\begin{lstlisting}[
    basicstyle=\ttfamily\small,
    breaklines=true,
    breakatwhitespace=true,
    xleftmargin=0pt,
    xrightmargin=0pt,
    columns=flexible,
    keepspaces=true,
    frame=none,
    resetmargins=true,
    breakindent=0pt
]
[INST]
Add these recipes to the Broccoli app:

1. Chicken Alfredo Pasta  
   - Description: A healthy, delicious meal.  
   - Servings: 2  
   - Prep Time: 10 mins  
   - Ingredients: As desired  
   - Directions: Cook pasta, toss with Alfredo sauce and grilled chicken. Top with Parmesan and spices.

2. Quinoa Salad with Vegetables  
   - Description: Quick and easy for busy days.  
   - Servings: 8  
   - Prep Time: 30 mins  
   - Ingredients: To your liking  
   - Directions: Mix quinoa, diced veggies, feta, and lemon olive oil dressing. Add spices for flavor.

3. Butternut Squash Soup  
   - Description: A healthy, delicious choice.  
   - Servings: 1  
   - Prep Time: 45 mins  
   - Ingredients: Per taste  
   - Directions: Saute onions and garlic, add squash and broth. Puree and season with nutmeg, salt, and pepper. Substitute as needed.
[/INST]
\end{lstlisting}
\begin{lstlisting}[
    basicstyle=\ttfamily\small,
    breaklines=true,
    breakatwhitespace=true,
    xleftmargin=0pt,
    xrightmargin=0pt,
    columns=flexible,
    keepspaces=true,
    frame=none,
    resetmargins=true,
    breakindent=0pt
]
[OBS]
\end{lstlisting}
\begin{center}
\includegraphics[width=0.2\columnwidth]{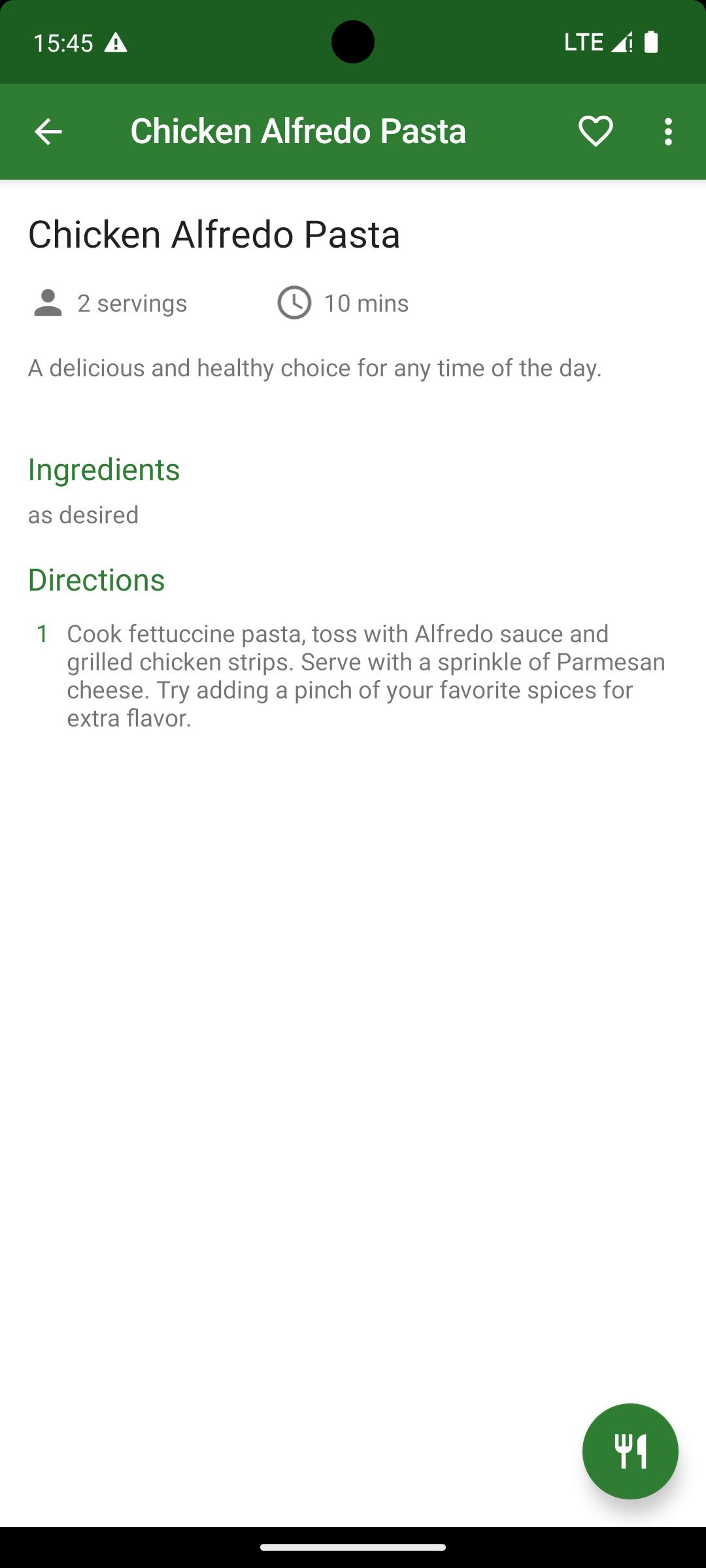}
\end{center}
\begin{lstlisting}[
    basicstyle=\ttfamily\small,
    breaklines=true,
    breakatwhitespace=true,
    xleftmargin=0pt,
    xrightmargin=0pt,
    columns=flexible,
    keepspaces=true,
    frame=none,
    resetmargins=true,
    breakindent=0pt
]
[/OBS]
\end{lstlisting}

\begin{lstlisting}[
    basicstyle=\ttfamily\small,
    breaklines=true,
    breakatwhitespace=true,
    xleftmargin=0pt,
    xrightmargin=0pt,
    columns=flexible,
    keepspaces=true,
    frame=none,
    resetmargins=true,
    breakindent=0pt
]
[TRAJ]
Step 1: Click Search.

Step 2: Input text Broccoli.

Step 3: Click Broccoli.

Step 4: Click New Recipe.

Step 5: Input text Chicken Alfredo Pasta.

Step 6: Input text A delicious and healthy choice for any time of the day.

Step 7: Input text 2 servings.

Step 8: Input text 10 mins.

Step 9: Input text as desired.

Step 10: Scroll down.

Step 11: Input text Cook fettuccine pasta, toss with Alfredo sauce and grilled chicken strips. Serve with a sprinkle of Parmesan cheese. Try adding a pinch of your favorite spices for extra flavor.

Step 12: Click SAVE.
[/TRAJ]
\end{lstlisting}

\textbf{$\blacktriangleright$ Output:}

\begin{lstlisting}[
    basicstyle=\ttfamily\small,
    breaklines=true,
    breakatwhitespace=true,
    xleftmargin=0pt,
    xrightmargin=0pt,
    columns=flexible,
    keepspaces=true,
    frame=none,
    resetmargins=true,
    breakindent=0pt
]
[ACTION_X]
Click Cook.
[/ACTION_X]
[SCORE_X]
TR: 1
[/SCORE_X]

[ACTION_Y]
Navigate back.
[/ACTION_Y]
[SCORE_Y]
TR: 0
[/SCORE_Y]
\end{lstlisting}
}
\end{tcolorbox}

In this case, the task is to create a recipe. The first 12 steps have completed the recipe creation process. In the Broccoli app, the next action should be to directly click ``Cook" (i.e., ACTION\_X), which is highly relevant to the instruction. In contrast, ``Navigate back" (i.e., ACTION\_Y) is not directly related to the instruction. Therefore, ACTION\_X receives a higher \textit{Task Relevance} score.

\subsection{\textit{Coherence}}

\begin{tcolorbox}[
    colback=gray!10!white,
    colframe=gray!50!black,
    title=Case of \textit{Coherence},
    fonttitle=\bfseries,
    boxrule=0.5mm,
    arc=2mm,
    width=\columnwidth,
    breakable,
    before skip=2mm,
    after skip=2mm,
    left=3pt,
    right=3pt,
    top=3pt,
    bottom=3pt
]

\textbf{$\blacktriangleright$ Input:} 

{\ttfamily\small
\begin{lstlisting}[
    basicstyle=\ttfamily\small,
    breaklines=true,
    breakatwhitespace=true,
    xleftmargin=0pt,
    xrightmargin=0pt,
    columns=flexible,
    keepspaces=true,
    frame=none,
    resetmargins=true,
    breakindent=0pt
]
[INST]
Add this exact product to my shopping cart. I think it is in the "Herbs, Spices \& Seasonings" category.
[/INST]
\end{lstlisting}
\begin{lstlisting}[
    basicstyle=\ttfamily\small,
    breaklines=true,
    breakatwhitespace=true,
    xleftmargin=0pt,
    xrightmargin=0pt,
    columns=flexible,
    keepspaces=true,
    frame=none,
    resetmargins=true,
    breakindent=0pt
]
[OBS]
\end{lstlisting}
\begin{center}
\includegraphics[width=0.7\columnwidth]{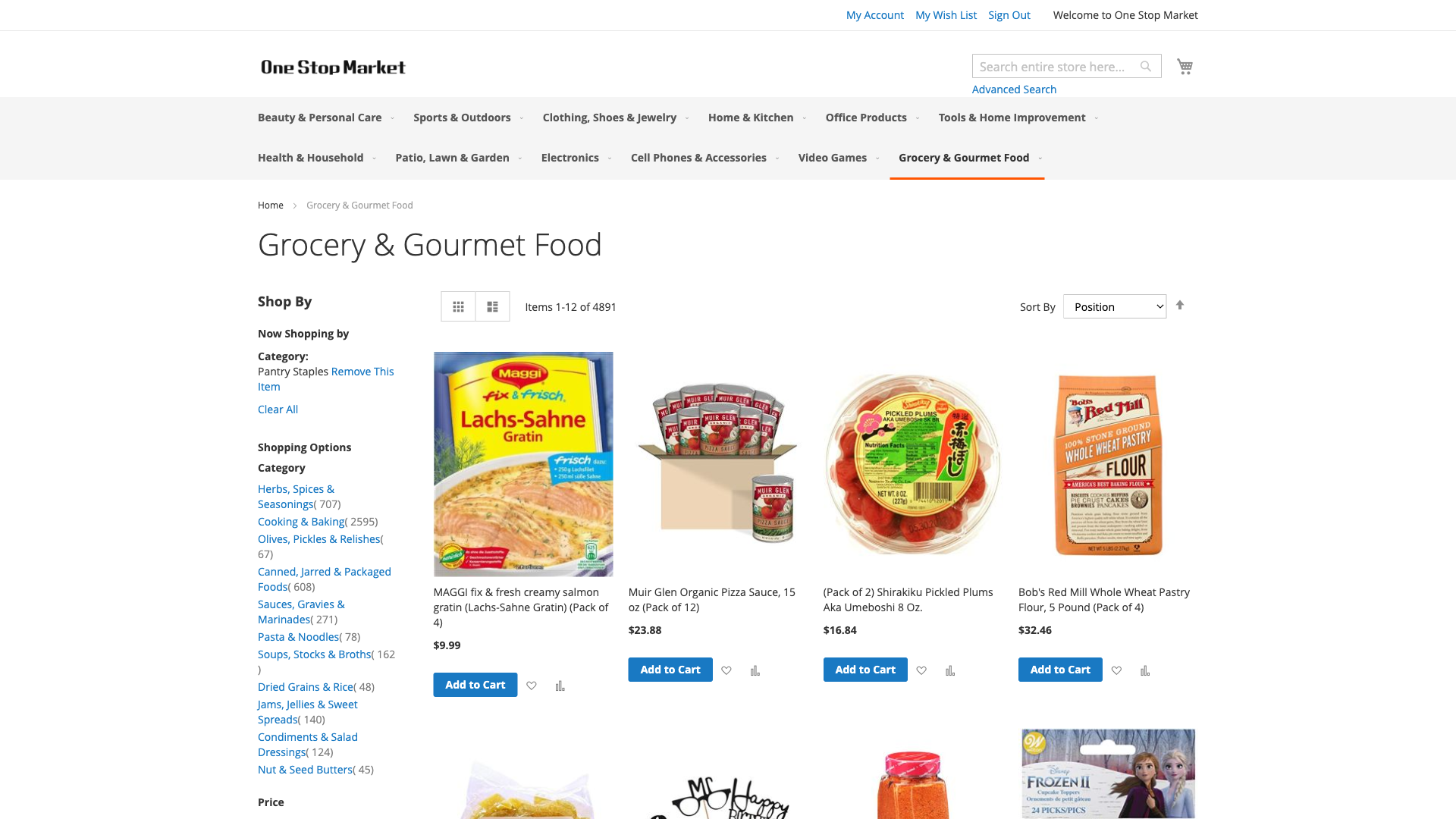}
\end{center}
\begin{lstlisting}[
    basicstyle=\ttfamily\small,
    breaklines=true,
    breakatwhitespace=true,
    xleftmargin=0pt,
    xrightmargin=0pt,
    columns=flexible,
    keepspaces=true,
    frame=none,
    resetmargins=true,
    breakindent=0pt
]
[/OBS]
\end{lstlisting}

\begin{lstlisting}[
    basicstyle=\ttfamily\small,
    breaklines=true,
    breakatwhitespace=true,
    xleftmargin=0pt,
    xrightmargin=0pt,
    columns=flexible,
    keepspaces=true,
    frame=none,
    resetmargins=true,
    breakindent=0pt
]
[TRAJ]
Step 1: Click menuitem '\ue622 Grocery & Gourmet Food' hasPopup: menu.

Step 2: Click link 'Pantry Staples( 4891 item )'.
[/TRAJ]
\end{lstlisting}

\textbf{$\blacktriangleright$ Output:}

\begin{lstlisting}[
    basicstyle=\ttfamily\small,
    breaklines=true,
    breakatwhitespace=true,
    xleftmargin=0pt,
    xrightmargin=0pt,
    columns=flexible,
    keepspaces=true,
    frame=none,
    resetmargins=true,
    breakindent=0pt
]
[ACTION_X]
Click link 'Herbs, Spices \& Seasonings( 707 item )'.
[/ACTION_X]
[SCORE_X]
C: 1
[/SCORE_X]

[ACTION_Y]
Click menuitem '\ue622 Grocery & Gourmet Food' hasPopup: menu.
[/ACTION_Y]
[SCORE_Y]
C: 0
[/SCORE_Y]
\end{lstlisting}
}
\end{tcolorbox}

In this case, the task is to find relevant products in the ``Herbs, Spices \& Seasonings" category. The previous step involved the agent entering a link, and the current step should logically advance the task. Clearly, ACTION\_X (clicking the ``Herbs, Spices \& Seasonings" link) is a coherent and logical continuation, while ACTION\_Y (repeating the action from step 1) is unintelligible and incoherent. Therefore, ACTION\_X receives a higher \textit{Coherence} score.

\begin{table*}[t]
\caption{Ablation study (inference experiments). \textbf{\texttt{Similar}} in table represents \textbf{\texttt{Similar}}-TM-Llama. }
\label{tab:ablation_study_more}
\renewcommand{\arraystretch}{1.1}
\vskip 0.15in
\vspace{-0.7cm}
\begin{center}
\small
\begin{sc}
\setlength{\tabcolsep}{6pt} 
\begin{tabular}{@{}lccccc|ccc@{}}
\toprule
& \multicolumn{5}{c}{\textbf{Dimension}} & \multicolumn{3}{c}{\textbf{Success Rate}} \\
\cmidrule(lr){2-6} \cmidrule(lr){7-9}
Model & H & OS & E & TR & C & \makecell{Android World} & \makecell{WebArena} & \makecell{OSWorld} \\
\midrule
Backbone & & & & & & 30.4 & 20.6 & 14.3 \\
\midrule
+H & $\checkmark$ & & & & & 32.5 & 26.1 & 15.8 \\
+OS & & $\checkmark$ & & & & 31.9 & 24.7 & 15.4 \\
+E & & & $\checkmark$ & & & 31.6 & 23.3 & 15.2 \\
+TR & & & & $\checkmark$ & & 31.1 & 21.6 & 14.9 \\
+C & & & & & $\checkmark$ & 30.9 & 21.0 & 14.8 \\
\midrule
+H,OS & $\checkmark$ & $\checkmark$ & & & & 33.4 & 31.4 & 16.7 \\
+H,E & $\checkmark$ & & $\checkmark$ & & & 33.1 & 29.8 & 16.5 \\
+H,TR & $\checkmark$ & & & $\checkmark$ & & 32.8 & 28.5 & 16.3 \\
+H,C & $\checkmark$ & & & & $\checkmark$ & 32.6 & 28.2 & 16.2 \\
+OS,E & & $\checkmark$ & $\checkmark$ & & & 32.7 & 27.5 & 16.3 \\
+OS,TR & & $\checkmark$ & & $\checkmark$ & & 32.3 & 26.8 & 16.0 \\
+OS,C & & $\checkmark$ & & & $\checkmark$ & 32.1 & 26.5 & 15.9 \\
+E,TR & & & $\checkmark$ & $\checkmark$ & & 31.8 & 25.9 & 15.7 \\
+E,C & & & $\checkmark$ & & $\checkmark$ & 31.7 & 25.7 & 15.6 \\
+TR,C & & & & $\checkmark$ & $\checkmark$ & 31.5 & 22.5 & 15.1 \\
\midrule
+H,OS,E & $\checkmark$ & $\checkmark$ & $\checkmark$ & & & 34.3 & 35.9 & 17.2 \\
+H,OS,TR & $\checkmark$ & $\checkmark$ & & $\checkmark$ & & 34.0 & 34.5 & 17.0 \\
+H,OS,C & $\checkmark$ & $\checkmark$ & & & $\checkmark$ & 33.8 & 34.2 & 16.9 \\
+H,E,TR & $\checkmark$ & & $\checkmark$ & $\checkmark$ & & 33.5 & 33.8 & 16.7 \\
+H,E,C & $\checkmark$ & & $\checkmark$ & & $\checkmark$ & 33.4 & 33.6 & 16.6 \\
+H,TR,C & $\checkmark$ & & & $\checkmark$ & $\checkmark$ & 33.2 & 33.1 & 16.5 \\
+OS,E,TR & & $\checkmark$ & $\checkmark$ & $\checkmark$ & & 32.9 & 32.8 & 16.4 \\
+OS,E,C & & $\checkmark$ & $\checkmark$ & & $\checkmark$ & 32.8 & 32.7 & 16.3 \\
+OS,TR,C & & $\checkmark$ & & $\checkmark$ & $\checkmark$ & 32.6 & 32.5 & 16.2 \\
+E,TR,C & & & $\checkmark$ & $\checkmark$ & $\checkmark$ & 32.4 & 32.3 & 16.1 \\
\midrule
+H,OS,E,TR & $\checkmark$ & $\checkmark$ & $\checkmark$ & $\checkmark$ & & 35.1 & 37.7 & 17.8 \\
+H,OS,E,C & $\checkmark$ & $\checkmark$ & $\checkmark$ & & $\checkmark$ & 34.7 & 37.2 & 17.6 \\
+H,OS,TR,C & $\checkmark$ & $\checkmark$ & & $\checkmark$ & $\checkmark$ & 34.2 & 36.5 & 17.3 \\
+H,E,TR,C & $\checkmark$ & & $\checkmark$ & $\checkmark$ & $\checkmark$ & 33.9 & 35.7 & 17.1 \\
+OS,E,TR,C & & $\checkmark$ & $\checkmark$ & $\checkmark$ & $\checkmark$ & 33.1 & 33.9 & 16.9 \\
\midrule
\textbf{\texttt{Similar}} & $\checkmark$ & $\checkmark$ & $\checkmark$ & $\checkmark$ & $\checkmark$ & \textbf{35.4} & \textbf{38.2} & \textbf{17.8} \\
\bottomrule
\end{tabular}
\end{sc}
\end{center}
\vskip -0.1in
\vspace{3pt}
\end{table*}

\section{Comprehensive Ablation Experiments}
\label{sec:ablation_more}

The extended ablation study, presented in Table~\ref{tab:ablation_study_more}, provides a more comprehensive analysis of the impact of each dimension in the \textbf{\texttt{Similar}} model. The results confirm the trends observed in the main experiments and offer additional insights into the contributions of the five dimensions—\textit{Helpfulness (H)}, \textit{Odds of Success (OS)}, \textit{Efficiency (E)}, \textit{Task Relevance (TR)}, and \textit{Coherence (C)}—across three benchmarks: Android World, WebArena, and OSWorld.

\subsection{Impact of Individual Dimensions}
The results demonstrate that the \textit{Helpfulness (H)} dimension has the most significant impact on performance, consistent with the findings in the main experiments. For example, adding \textit{H} alone improves the success rate on Android World from 30.7\% to 32.5\%, on WebArena from 20.6\% to 26.1\%, and on OSWorld from 14.6\% to 15.8\%. This aligns with our hypothesis that the quality of a step is primarily reflected in its contribution to task completion, which \textit{H} effectively captures. The \textit{Odds of Success (OS)} dimension follows, with improvements of 31.9\%, 24.7\%, and 15.4\% on the respective benchmarks, indicating its importance in guiding the agent. The \textit{Efficiency (E)} dimension also contributes positively, though to a lesser extent, while \textit{Task Relevance (TR)} and \textit{Coherence (C)} show more modest improvements, consistent with their secondary roles.

\subsection{Combined Impact of Multiple Dimensions}
The extended results further highlight the synergistic effects of combining multiple dimensions. For instance, the combination of \textit{H} and \textit{OS} achieves success rates of 33.4\%, 31.4\%, and 16.7\% on Android World, WebArena, and OSWorld, respectively, outperforming models with only one of these dimensions. Similarly, the combination of \textit{H}, \textit{OS}, and \textit{E} yields even higher success rates (34.3\%, 35.9\%, and 17.2\%), demonstrating the cumulative benefits of integrating complementary dimensions. These results reinforce the importance of fine-grained rewards over coarse-grained ones, as models with partial-dimensional rewards consistently underperform compared to the full \textbf{\texttt{Similar}} model.



\vspace{-10pt}

\section{More Visualizations of \textbf{\textit{SRMEval}}}
\label{sec:visualization_of_srmeval}

\begin{table*}[t]
\caption{Cases of \textbf{\textit{SRMEval}}.}
\label{tab:visualization_of_srmeval}
\renewcommand{\arraystretch}{1.5}
\begin{center}
\small
\begin{tabular}{@{}>{\centering\arraybackslash}m{2.5cm}>{\centering\arraybackslash}m{3cm}>{\centering\arraybackslash}m{1.3cm}>{\raggedright\arraybackslash}m{1.8cm}>{\centering\arraybackslash}m{0.6cm}|>{\centering\arraybackslash}m{1.8cm}>{\centering\arraybackslash}m{1.8cm}@{}}
\toprule
\textbf{Instruction} & \textbf{Observation} & \textbf{Step Idx} & \textbf{Trajectory} & \textbf{Type} & \multicolumn{2}{|c}{\textbf{Candidate Action Pair}} \\
\midrule
In Simple Calendar Pro, delete all the events. & 
\raisebox{-0.1\height}{\includegraphics[width=0.04\textwidth]{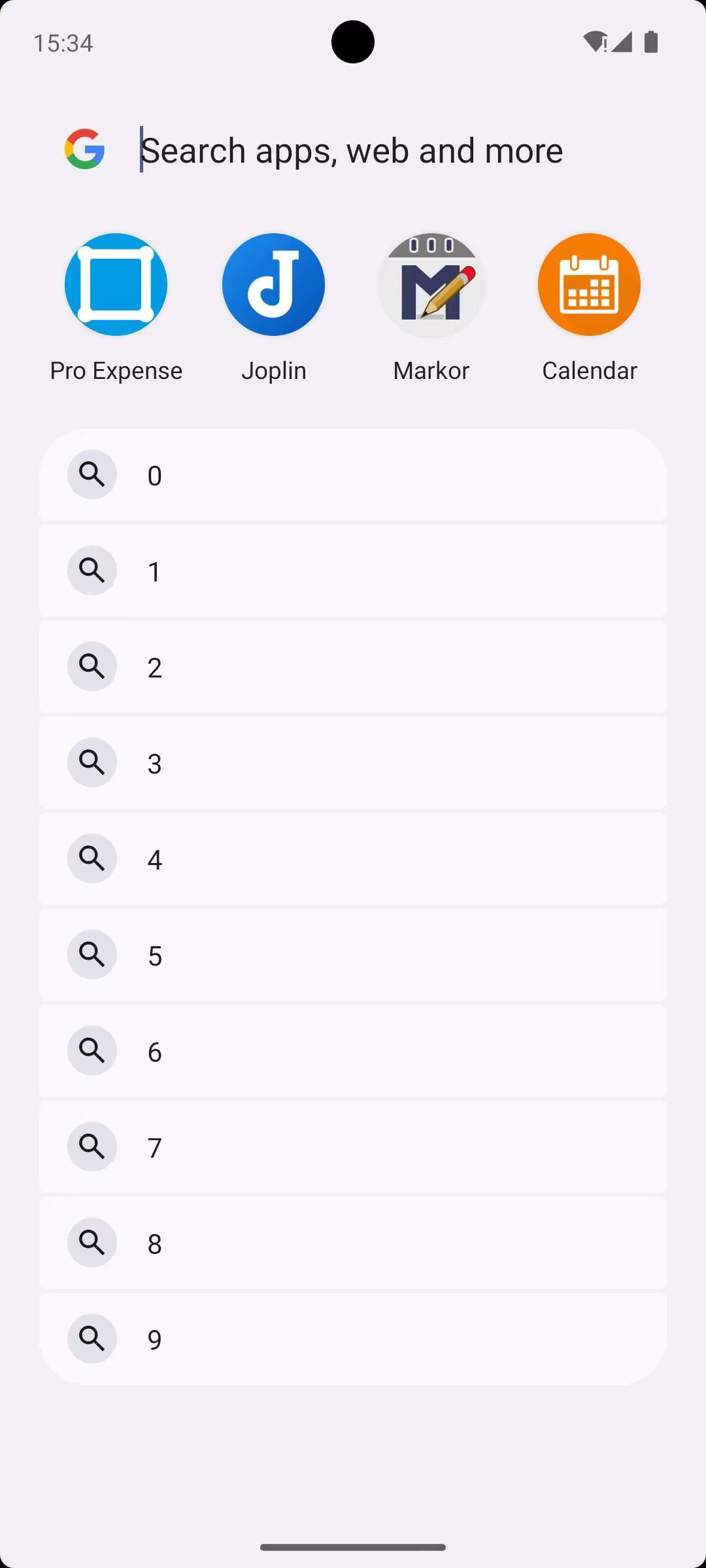}} & 
2 & 
Step 1: Click Search. & 
E & 
Click Calender. & 
Input text ``Simple Calendar Pro''. \\
\midrule

In Simple Calendar Pro, delete all the calendar events on 2023-10-27. & 
\raisebox{-0.1\height}{\includegraphics[width=0.045\textwidth]{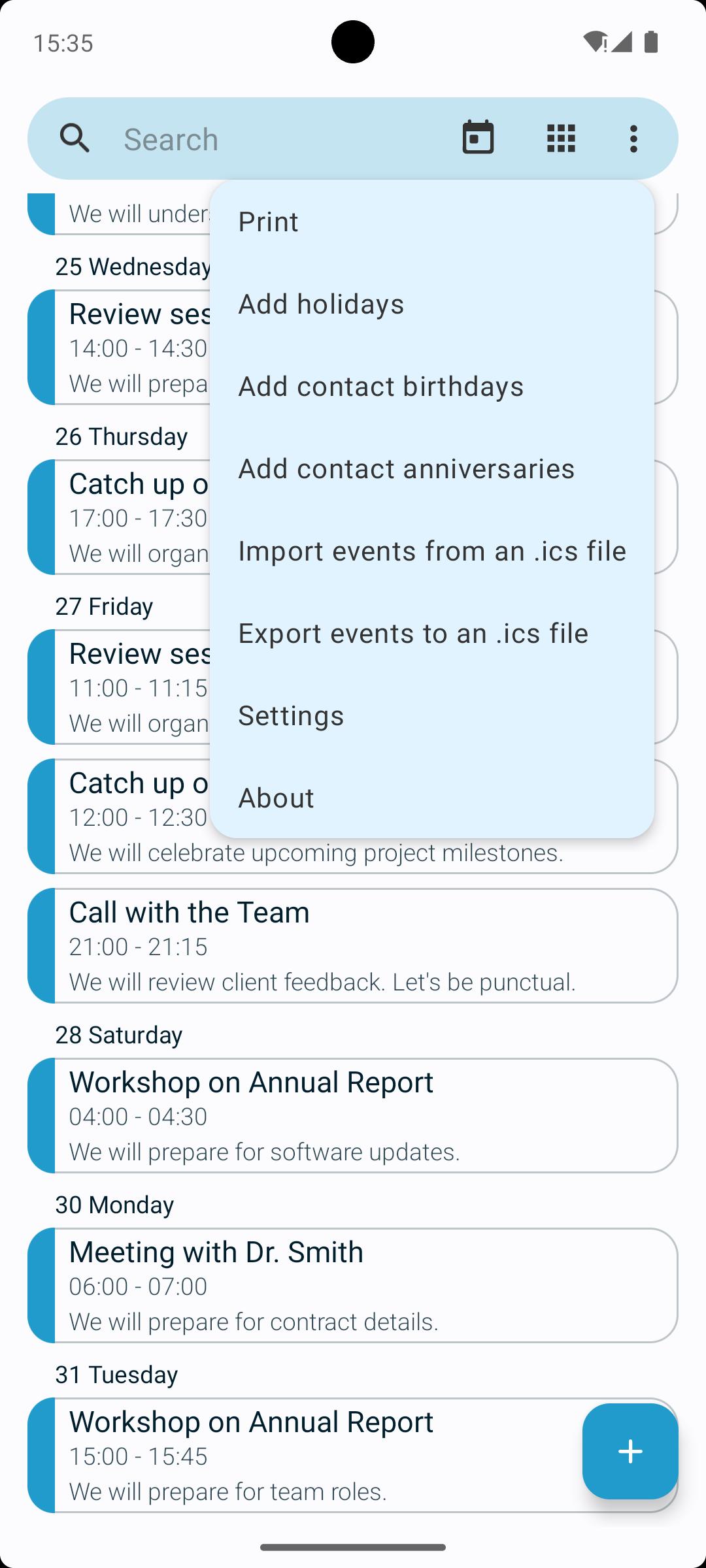}} & 
6 & 
Step 1: Click Search.

Step 2: Click Calendar.

Step 3: Scroll down.

Step 4: Long press 27 Friday.

Step 5: Click More options. & 
OS & 
Navigate back. & 
Click Import events from an.ics file. \\
\midrule

Add this exact product to my shopping cart. I think it is in the ``Herbs, Spices \& Seasonings" category.& 
\raisebox{-0.1\height}{\includegraphics[width=0.2\textwidth]{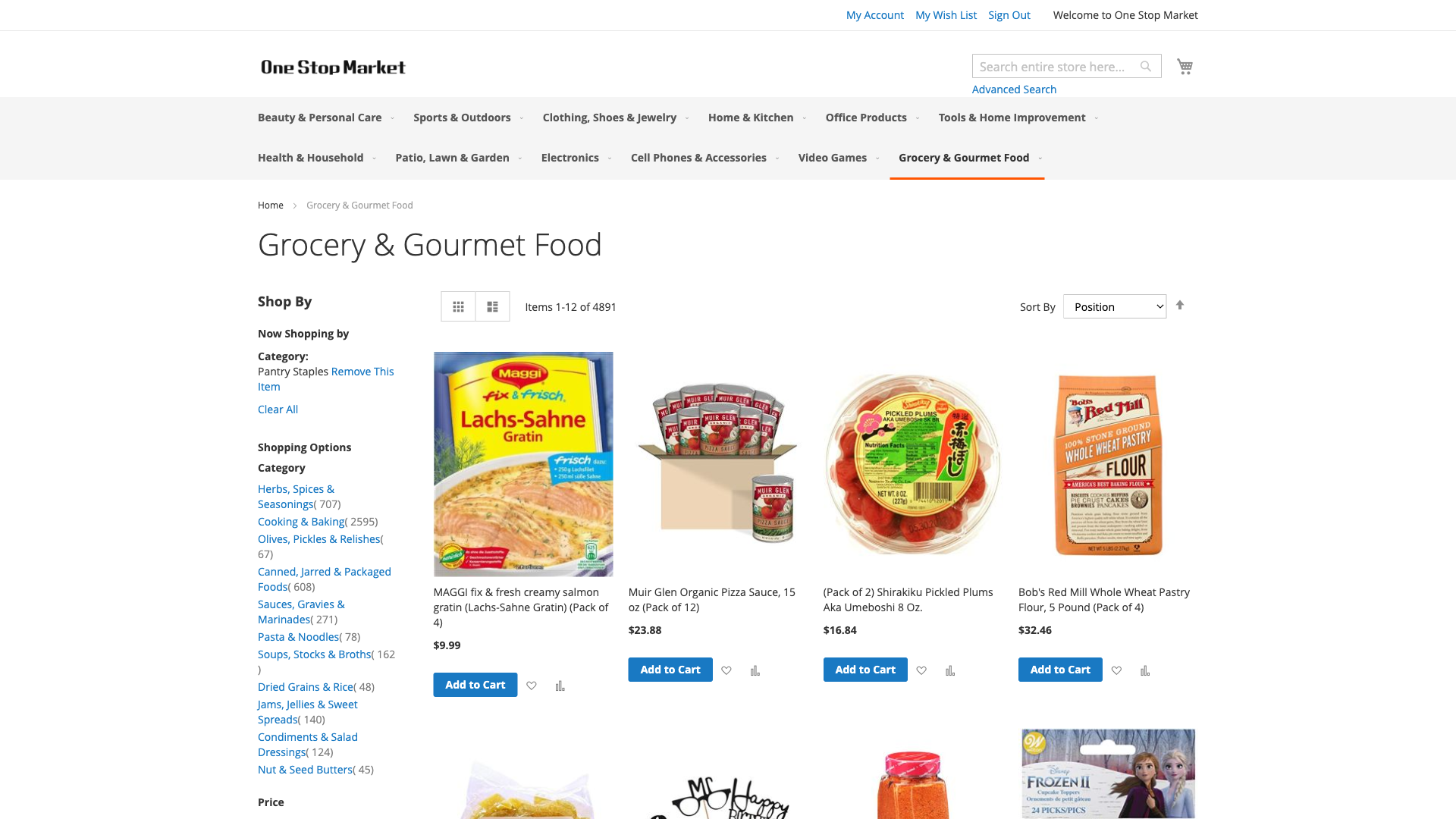}} & 
3 & 
Step 1: Click menuitem ``Grocery \& Gourmet Food'' hasPopup: menu.

Step 2: Click link ``Pantry Staples( 4891 item )''. & 
C & 
Click link ``Herbs, Spices \& Seasonings( 707 item )''. & 
Click menuitem ``Grocery \& Gourmet Food'' hasPopup: menu. \\
\midrule

Add a exact product to my shopping cart. & 
\raisebox{-0.1\height}{\includegraphics[width=0.2\textwidth]{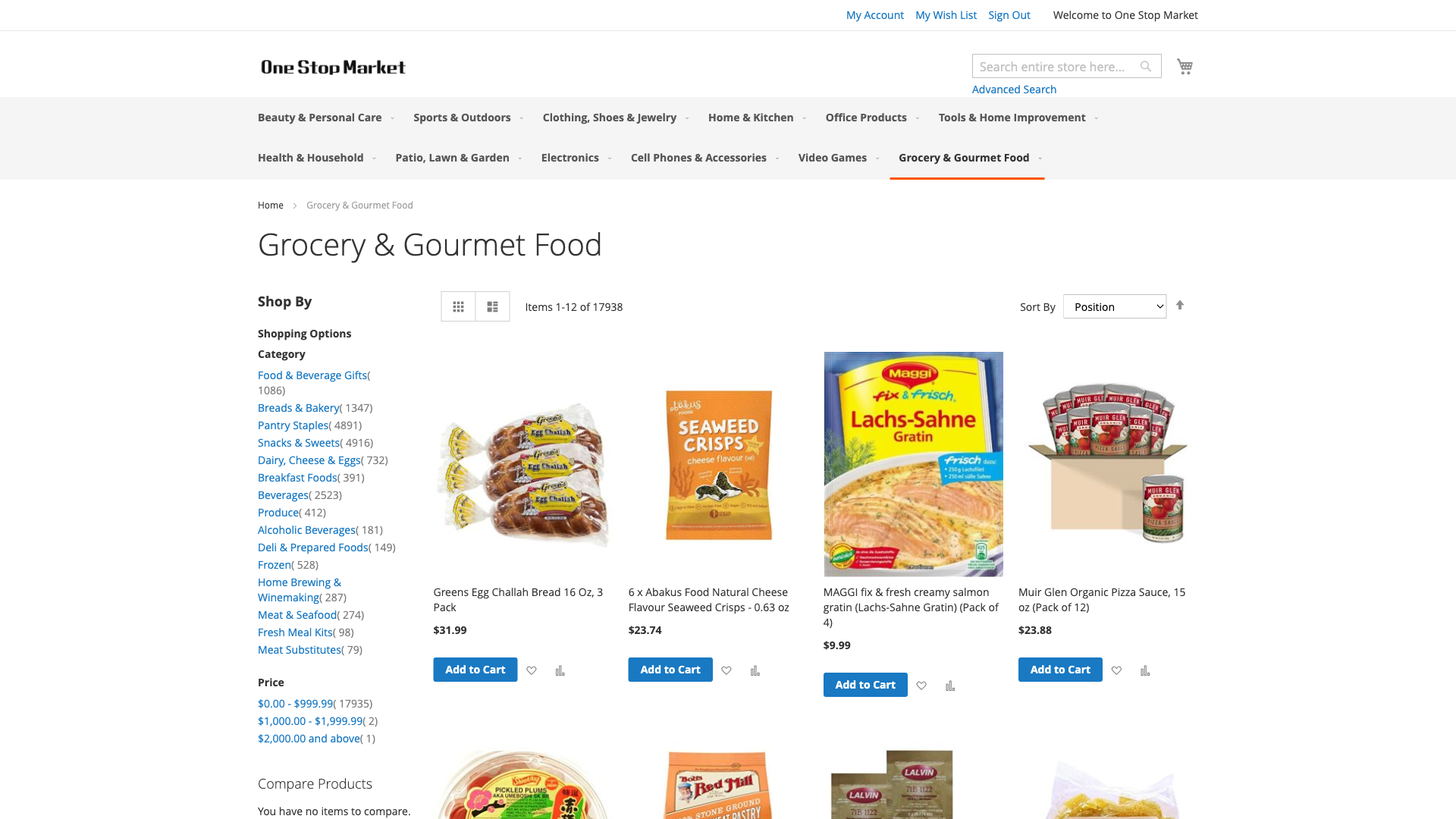}} & 
2 & 
Step 1: Click menuitem ``Grocery \& Gourmet Food'' hasPopup: menu. & 
H & 
Scroll down. & 
Click menuitem ``Grocery \& Gourmet Food'' hasPopup: menu. \\
\midrule

Can you add the red flower seeds with around 4 stars to my cart? & 
\raisebox{-0.1\height}{\includegraphics[width=0.2\textwidth]{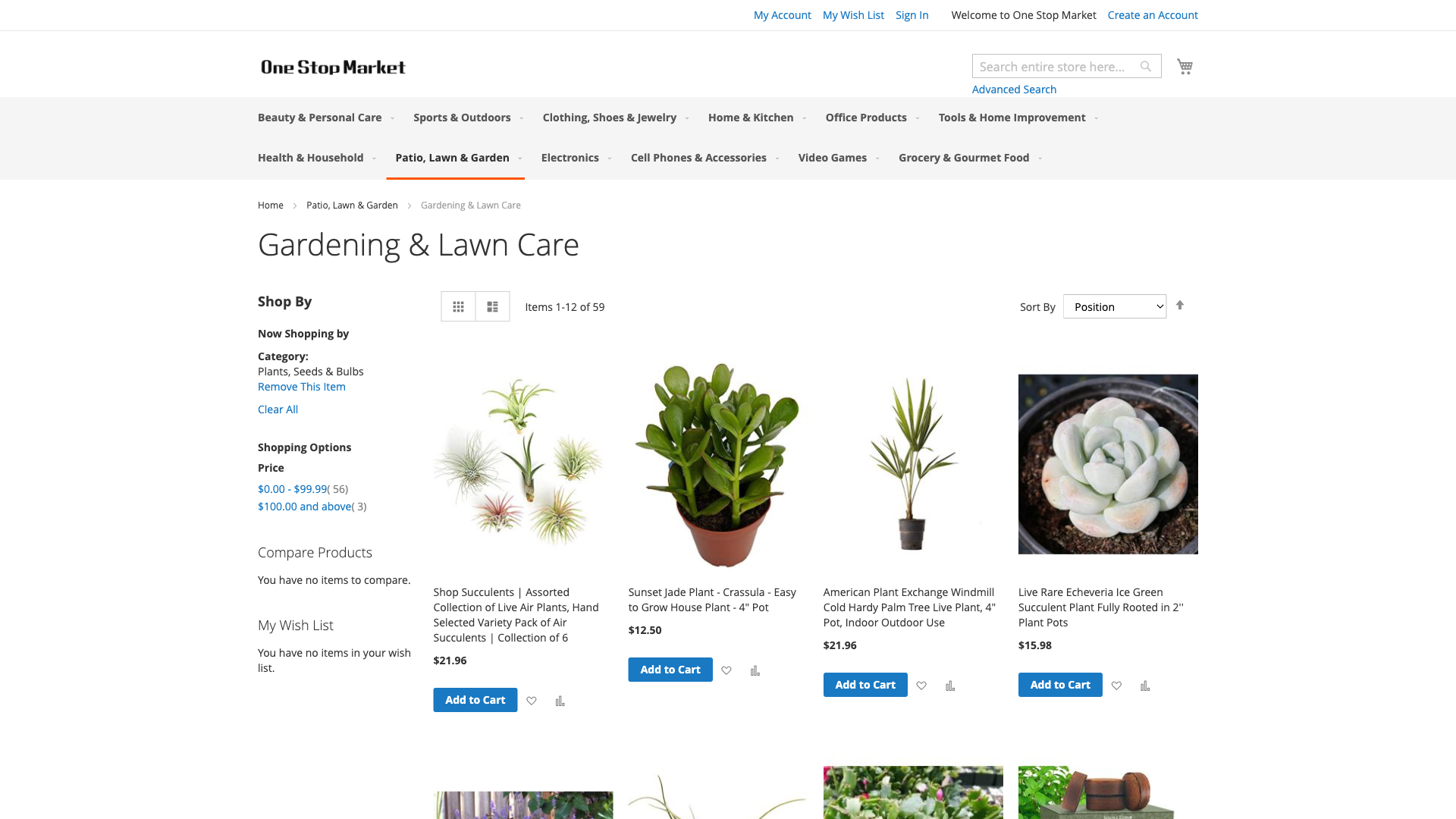}} & 
4 & 
Step 1: Scroll down.

Step 2: Click link ``Page 2''.

Step 3: Click link ``Plants, Seeds \& Bulbs( 59 item )'' & 
TR & 
Click link ``Page Next''.  & 
Hover menuitem ``Patio, Lawn \& Garden'' hasPopup: menu. \\
\midrule

Can you make Bing the main search thingy when I look stuff up on the internet? & 
\raisebox{-0.1\height}{\includegraphics[width=0.2\textwidth]{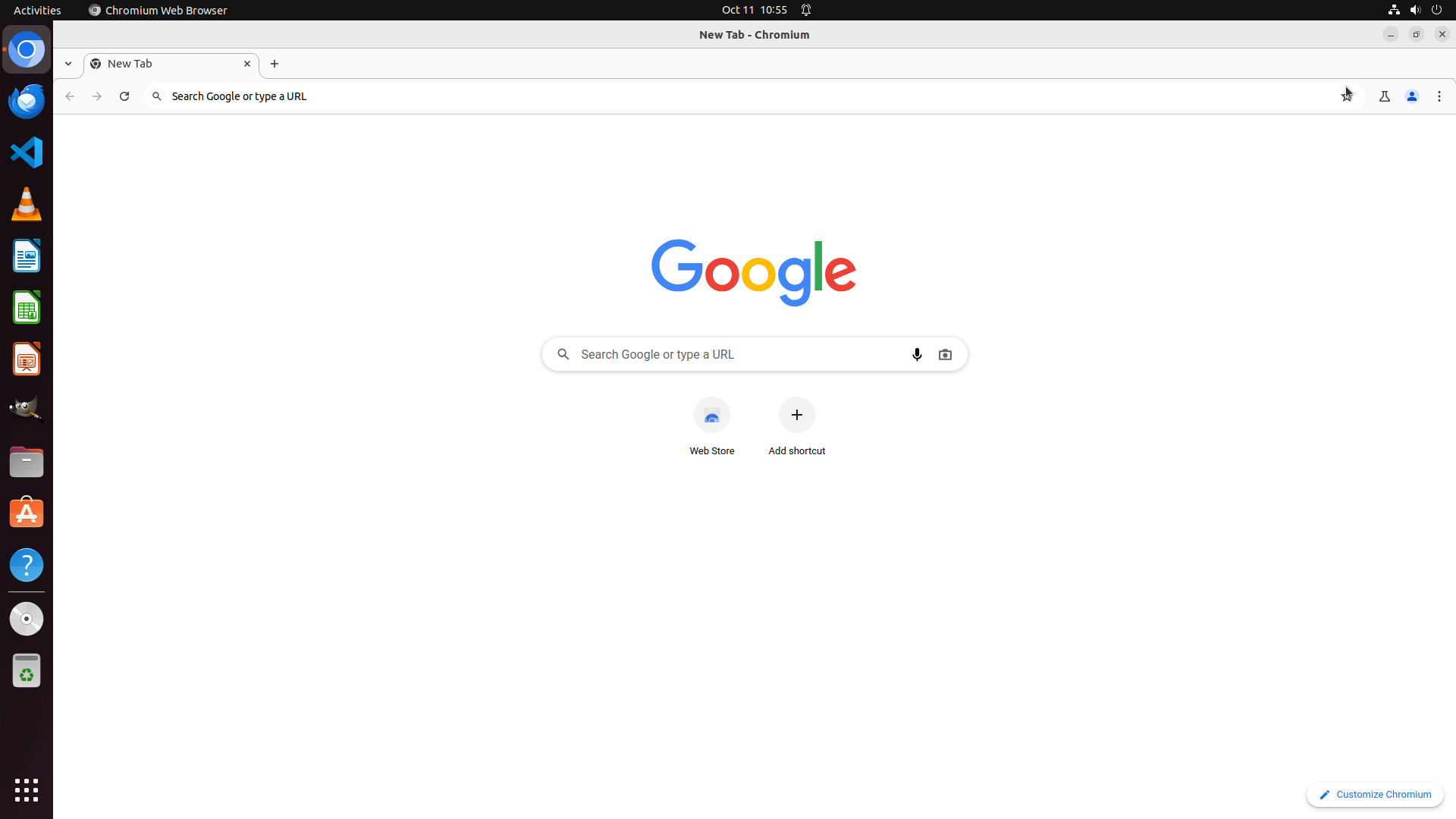}} & 
3 & 
Step 1: Coordinates for ``Customize Chromium".

Step 2: Click on ``Customize Chromium". & 
Total & 
Click on the address and search bar Type the URL to access search engine settings.  & 
Click on the ``Account Settings" button coordinates for ``Account Settings".
 \\

\bottomrule
\end{tabular}
\end{center}
\vspace{-0.2cm}
\end{table*}

As depicted in Table~\ref{tab:visualization_of_srmeval}, we present additional visualizations of \textit{SRMEval}. From the data in the table, we can more clearly understand the content of \textit{SRMEval}, which is the first benchmark in the virtual agent domain designed for step-wise, multi-dimensional, and multi-platform evaluation of reward models. It comprehensively tests the ability of reward models to assess the quality of agent actions, as well as the degree of preference alignment.

\end{document}